\documentclass[10pt,twocolumn,letterpaper]{article}

\usepackage[pagenumbers]{cvpr}

\usepackage{cuted}      %
\usepackage{currfile}
\usepackage{caption}
\usepackage{amssymb,amsthm,mathtools,mathrsfs}
\usepackage{enumitem}
\usepackage{algorithm}
\usepackage{algpseudocode}
\usepackage{multirow}
\usepackage{placeins}
\usepackage{url}
\usepackage{tabularx}   %
\usepackage{colortbl}   %
\definecolor{abiblock}{RGB}{217,241,213}
\colorlet{baseblock}{gray!10}

\usepackage{amsmath,amsfonts,bm}

\def\eqref#1{equation~\ref{#1}}

\def\1{\bm{1}}

\def\va{{\bm{a}}}
\def\vb{{\bm{b}}}
\def\vc{{\bm{c}}}

\DeclareMathAlphabet{\mathsfit}{\encodingdefault}{\sfdefault}{m}{sl}
\SetMathAlphabet{\mathsfit}{bold}{\encodingdefault}{\sfdefault}{bx}{n}

\newcommand{\E}{\mathbb{E}}

\newcommand{\R}{\mathbb{R}}

\newcommand{\Var}{\mathrm{Var}}

\DeclareMathOperator*{\argmax}{arg\,max}

\newcommand{\method}{ABI}

\providecommand{\tbd}{\textcolor{red}{TBD}}

\newcommand{\abilogo}{\smash{\raisebox{-0.3em}{\includegraphics[height=1.25em]{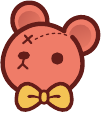}}}\hspace{0.3em}}

\newif\ifcfloss
\cflossfalse  %

\newcommand{\T}{\mathcal{T}}
\newcommand{\Ind}{\mathbf{1}}

\DeclareMathOperator{\Law}{Law}
\DeclareMathOperator{\spn}{span}
\providecommand{\TV}{\operatorname{TV}}
\newcommand{\norm}[1]{\lVert #1\rVert}
\newcommand{\supnorm}[1]{\lVert #1\rVert_{\infty}}
\newcommand{\opnorm}[1]{\lVert #1\rVert_{\mathrm{op}}}

\newcommand{\hF}{\widehat F}

\newcommand{\hJ}{\widehat J}
\newcommand{\hg}{\widehat g}

\theoremstyle{plain}
\newtheorem{proposition}{Proposition}
\newtheorem{lemma}{Lemma}[section]
\newtheorem{corollary}{Corollary}
\theoremstyle{definition}

\newtheorem{example}{Example}[section]
\theoremstyle{remark}
\newtheorem{remark}{Remark}[section]

\renewcommand{\paragraph}[1]{\vspace{.35em}\noindent\textbf{#1}\hspace{.4em}}

\definecolor{cvprblue}{rgb}{0.21,0.49,0.74}
\usepackage[pagebackref,breaklinks,colorlinks,allcolors=cvprblue]{hyperref}
\hypersetup{pdftitle={ABI: Actor-Free Behavior-Induced Representations for Zero-Shot Test-Time Guidance},
  pdfauthor={Yifei Yao, Yude Zou, Guanbo Jiao, Jianming Ma, Xing Gao, Chunhua Shen}}

\def\confName{CVPR}
\def\confYear{2027}

\newif\ifsupp
\supptrue

\title{\abilogo ABI: Actor-Free Behavior-Induced\\Representations for Zero-Shot Test-Time Guidance}

\author{%
Yifei Yao\textsuperscript{1,*}\quad
Yude Zou\textsuperscript{2,*}\quad
Guanbo Jiao\textsuperscript{3,*}\quad
Jianming Ma\textsuperscript{2}\quad
Xing Gao\textsuperscript{1}\quad
Chunhua Shen\textsuperscript{4}\\[0.35em]
{\normalsize \textsuperscript{1}Shanghai Artificial Intelligence Laboratory\hspace{0.4em}
\textsuperscript{2}Shanghai Jiao Tong University\hspace{0.4em}
\textsuperscript{3}Fudan University\hspace{0.4em}
\textsuperscript{4}Zhejiang University}%
}

\makeatletter
\def\@maketitle{%
   \newpage
   \null
   \vskip .12in
   \begin{center}
      {\Large \bf \@title \par}
      \vspace*{10pt}%
      {\large
        \lineskip .5em
        \begin{tabular}[t]{c}
          \@author
        \end{tabular}
        \par
      }
      \vspace*{8pt}
   \end{center}
}
\newcommand\blfootnote[1]{%
  \begingroup
  \renewcommand\thefootnote{}%
  \def\@makefntext##1{\noindent ##1}%
  \footnotetext{#1}%
  \addtocounter{footnote}{-1}%
  \endgroup
}
\makeatother

\newcommand{\abiprojectpage}{\url{https://securitycfs.github.io/ABI-demo/}}

\begin{document}
\maketitle
\begin{abstract}
Behavior-cloning policies show strong ability in sequential control by generating coherent multi-step action chunks, yet their purely imitative objective limits zero-shot generalization to unseen tasks. While offline reinforcement learning offers principled policy improvement, existing methods either require task-specific reward supervision or remain tied to a specific actor during critic training or policy deployment, inflating training cost and restricting cross-policy generalization. We observe that evaluating actions under the data-collecting behavior largely preserves their relative value ordering across target policies, offering a way around actor coupling. Combining this property with reward-free successor prediction, we propose Actor-Free Behavior-Induced Representations (\method{}) to learn multi-horizon action-chunk representations from unlabeled trajectories without reward supervision or actor training. At test time, pairing the frozen representation with lightweight task readouts forms a chunk critic that guides target policies without parameter updates and transfers directly across distinct unseen policies. Our analysis establishes when the value ordering is preserved across policies and when selecting among proposed chunks improves closed-loop return. Across 65 locomotion, navigation, and manipulation tasks, \method{}'s chunk critic guides weak imitation policies to match or surpass zero-shot RL baselines with high sample efficiency, outperforms existing zero-shot critics on identical proposals, and improves externally trained RL actors without retraining. From raw pixels, \method{} delivers substantial guidance gains and surpasses prior zero-shot RL baselines on visual manipulation, highlighting its potential for visuomotor control.\ifdefined\abiprojectpage\par\vspace{0.25em}\noindent{\normalfont\footnotesize\textbf{Project page:} \abiprojectpage}\vspace{-6pt}\fi
\end{abstract}

\blfootnote{\textsuperscript{*}Equal contribution.\\Contact at \href{mailto:yao.yifei.ro@gmail.com}{\texttt{yao.yifei.ro@gmail.com}}.}
\section{Introduction}
\label{sec:intro}
\ifdefined\abiteaserearly\else\begin{figure*}[t]
\centering
\includegraphics[width=\linewidth]{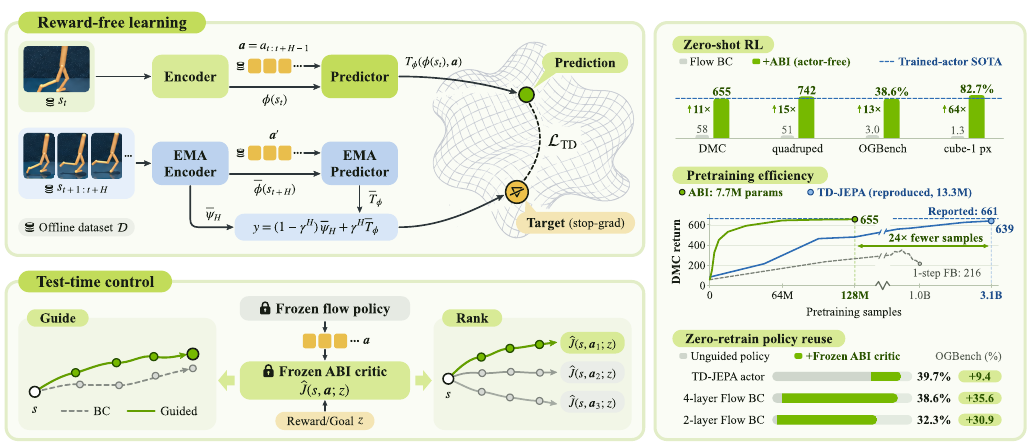}
\caption{\textbf{\method{} pretrains a reward- and actor-free chunk representation to guide frozen policies at test time.} \emph{Left:} From unlabeled trajectories $\mathcal{D}$, the predictor $T_\phi$ regresses discounted chunk arrivals and EMA targets without rewards or an actor, as in \cref{sec:method-latent}. At test time, a task readout on cached states forms a zero-update chunk critic $\hJ(s,\va;z)$ to guide policies via gradient guidance (\emph{Guide}) or candidate selection (\emph{Rank}). \emph{Right:} \method{} guides frozen Flow BC to match or surpass trained-actor baselines on state suites and visual manipulation, reaches reproduced TD-JEPA's DMC return with $24\times$ fewer samples, and reuses one frozen critic across distinct unseen policies.\looseness=-1}
\label{fig:teaser}
\label{fig:pipeline}
\end{figure*}
\fi

Modern generative policies in manipulation and other control domains widely rely on behavior cloning from demonstrations to predict multi-step action chunks \cite{zhao2023learning}. Representative approaches include diffusion \cite{chi2023diffusion} and flow-matching policies \cite{lipman2023flow}, as well as large-scale vision-language-action (VLA) models \cite{black2024pi0,kim2024openvla,intelligence2025pi_} and world-action models (WAMs) \cite{ye2026world,kim2026cosmos,yuan2026fast,wang2026openwam}. Imitating demonstrations at the chunk level captures multimodal action patterns while maintaining temporal consistency across long horizons. However, collected demonstrations often cover only specific tasks or local state regions. Purely imitative policies can reproduce demonstrated behaviors, but still lack evaluative signals to recombine them for unseen objectives.

Several paradigms leverage offline trajectory datasets to improve policies beyond pure imitation. Reward-supervised offline reinforcement learning, such as IQL \cite{kostrikov2022offline} and its extensions to generative or chunked policies \cite{hansen2023idql,nakamoto2024steering,park2025flow,li2025qchunking}, learns value critics without online interaction, but still depends on task-specific reward labels and per-task retraining. Zero-shot reinforcement learning reduces this requirement by pretraining task-agnostic representations on unlabeled trajectories to solve unseen tasks at test time \cite{park2024foundation}, as in Forward--Backward (FB) \cite{touati2021learning,touati2022does} and TD-JEPA \cite{bagatella2026td}. These representations underlie frontier control policies such as Meta Motivo \cite{tirinzoni2025zeroshot} and BFM-Zero \cite{li2026bfmzero}. However, existing zero-shot methods mostly rely on jointly training a task-conditioned actor for bootstrapping, which induces out-of-distribution overestimation \cite{jeen2024zero} and unstable convergence \cite{zheng2026can}. Coupling critic learning to an actor also escalates training cost with policy complexity and weakens generalization across policies. To avoid this coupling, one-step FB \cite{zheng2026can} removes the actor from critic training by evaluating single actions under the data-collecting behavior \cite{brandfonbrener2021offline}, yet still trains a task-conditioned actor for execution. Whether zero-shot RL can remove actor dependence in both training and execution to guide independently trained policies has not been fully explored.

Our observation, that evaluating actions under the data-collecting behavior largely preserves their relative value ordering across target policies, suggests a way around this coupling. Combining this property with task-agnostic successor representations \cite{dayan1993improving,barreto2017successor} allows value representations to be trained without a target policy or rewards. Since test-time guidance only compares actions proposed by a policy, a critic formed from these representations can guide that policy without updating it. Building on this principle, we introduce \textbf{A}ctor-Free \textbf{B}ehavior-\textbf{I}nduced Representations (\textbf{\method{}}) for chunked control, as shown in \cref{fig:pipeline}. \method{} pretrains a multi-horizon encoder--predictor backbone over multi-step action chunks from unlabeled state or pixel trajectories without rewards, task inputs, or simulators. Paired with lightweight task readouts on cached states, \method{} forms a test-time chunk critic without gradient updates \cite{touati2022does,bagatella2026td,anonymous2026density}. This critic guides both chunked and single-step policies via candidate selection or Q-guided flow (QGF) \cite{zhou2026test}. We further establish when it preserves cross-policy ranking, improves closed-loop return under selection, and yields local gains under gradient guidance.

We evaluate \method{} on 65 locomotion, navigation, and manipulation tasks across 13 state and pixel datasets \cite{bagatella2026td}, as summarized in \cref{fig:teaser}. Pretrained once per dataset without an actor, \method{} guides a weak Flow BC policy from 58.4 to 655.2 return on DMC and from 3.0\% to 38.6\% success on OGBench, matching or surpassing multiple zero-shot RL baselines while using $24\times$ fewer DMC pretraining samples than TD-JEPA reproduced from its official code and reported settings. To verify that these gains stem from the critic itself, we compare zero-shot critics on identical action proposals: \method{} leads on both suites, more than doubles OGBench success over TD-JEPA and one-step FB critics, and avoids the sharp drop of single-step critics when executing full action chunks. Moreover, the same frozen critic transfers across policies without retraining, improving external TD-JEPA and one-step FB actors trained with their own critics and achieving higher task success with stronger base proposers. Finally, \method{} scales directly to raw pixels, substantially lifting visual Flow BC and surpassing prior pixel zero-shot RL baselines across all four visual manipulation datasets. We summarize our contributions as follows:\looseness=-1
\begin{enumerate}[leftmargin=*,nosep]
\item We propose \method{}, an efficient actor-free framework that pretrains reward-free multi-horizon chunk representations from states or pixels and forms zero-update critics for test-time policy guidance.
\item We establish theoretical conditions for \method{} guidance, covering cross-policy value-ordering preservation, closed-loop gains under whole-chunk and rolling selection, and data-supported gradient steps.
\item We empirically demonstrate that \method{} guides weak imitation policies to match or surpass zero-shot RL baselines, outperforms prior critics on shared proposals, and improves external actors without retraining.
\end{enumerate}

\section{Related Work}
\label{sec:related}

\paragraph{Zero-shot reinforcement learning.}
Classical successor representations factorize action values into reward-free future dynamics and linear reward weights \cite{dayan1993improving,barreto2017successor}. Forward--Backward (FB) factorizations model future state occupancies \cite{blier2021learning,touati2021learning,touati2022does} and underlie behavioral foundation models \cite{tirinzoni2025zeroshot,li2026bfmzero}, while other methods estimate occupancies via contrastive, generative, or categorical TD objectives \cite{eysenbach2021clearning,eysenbach2022contrastive,zheng2024contrastive,zhu2024distributional,farebrother2025temporal,anonymous2026density} or embed goal distances \cite{park2024foundation,park2026dual}. Building on self-predictive latent architectures \cite{grill2020bootstrap,schwarzer2021data,assran2023ijepa,assran2025vjepa2}, TD-JEPA predicts target-network embeddings directly in latent space \cite{bagatella2026td}. However, most zero-shot RL methods jointly train a task-conditioned actor for bootstrapping and control, whereas one-step FB \cite{zheng2026can} bootstraps single actions on the data-collecting behavior \cite{brandfonbrener2021offline} yet still trains an actor for execution. In contrast, \method{} removes task-conditioned actor training in both pretraining and deployment, extending behavior-bootstrapped latent prediction from single actions to multi-horizon action chunks.

\paragraph{Test-time policy guidance.}
Return-conditioned and classifier-guided trajectory models direct generation toward high-value outcomes \cite{janner2022planning,ajay2023conditional}, while offline RL critics \cite{fujimoto2019off,kostrikov2022offline} avoid unstable policy fine-tuning by re-ranking or reweighting action proposals \cite{ghasemipour2021emaq,chen2023offline,hansen2023idql,nakamoto2024steering} or guiding latent or flow-matching sampling at inference \cite{wagenmaker2025steering,zhou2026test,yang2026guided}. Recent offline RL methods also distill one-step actors and learn chunk-level Q-functions \cite{park2025flow,li2025qchunking,li2025decoupled}, and success verifiers trained on outcome-labeled rollouts select among sampled actions \cite{attarian2026update,park2026diver}. However, these approaches still require task-specific reward or outcome labels during training or couple critic updates to a specific policy. \method{} instead learns one reward-free representation per unlabeled dataset, forming zero-update chunk critics to guide diverse frozen policies via candidate selection or gradient steps.

\section{Method}
\label{sec:method}

\method{} pretrains a task-free representation of discounted multi-step action consequences under the data-collecting behavior $\beta$ from unlabeled trajectories $\mathcal D$, independent of any downstream policy such as a behavior-cloning proposer $\mu$. Paired with lightweight task readouts on cached offline states at test time, this frozen representation directly forms a chunk critic $\hJ$ without gradient updates, as shown in \cref{fig:pipeline}. Because $\hJ$ evaluates continuation under $\beta$ rather than the guided policy, we further establish when it preserves candidate ordering across policies and improves closed-loop control under test-time guidance.

\subsection{Evaluating chunks under a fixed behavior}
\label{sec:prelim}

Unlike actor-coupled successor methods \cite{touati2021learning,bagatella2026td} or single-action evaluation \cite{zheng2026can}, \method{} evaluates multi-step action chunks without an actor by fixing post-chunk continuation to the data-collecting behavior $\beta$. Consider a controlled Markov process $(\mathcal S,\mathcal A,\mathcal P,\gamma)$ with discount $\gamma\in(0,1)$ and bounded reward $r:\mathcal S\to\R$, and let $\beta(\cdot\mid s)$ denote the empirical distribution of $H$-step chunks in $\mathcal D$ from state $s$ across single or mixed collection sources. Executing an $H$-step chunk $\va=(a_0,\ldots,a_{H-1})$ from $s_0=s$ and continuing under $\beta(\cdot\mid s_{kH})$ at boundary states $s_{kH}$ for $k\ge1$ defines chunk value $Q^\beta_r(s,\va):=\E^\beta_{s,\va}\sum_{t\ge0}\gamma^t r(s_{t+1})$, normalized occupancy $m_\beta$, and normalized chunk value $J^\beta_r$:
\begin{equation}
\begin{aligned}
&m_\beta(\cdot\mid s,\va):=(1-\gamma)\E^\beta_{s,\va}
\textstyle\sum_{t\ge0}\gamma^t\delta_{s_{t+1}}(\cdot),\\
&J^\beta_r(s,\va):=\textstyle\int r(x)m_\beta(\mathrm dx\!\mid\! s,\va)
=(1-\gamma)Q^\beta_r(s,\va),
\end{aligned}
\label{eq:readout-occupancy}
\end{equation}
with Dirac measure $\delta_{s_{t+1}}$ at $s_{t+1}$. Because $m_\beta$ does not depend on $r$, a bounded feature map $\psi:\mathcal S\to\R^d$ summarizes discounted outcomes via the chunk successor representation
\begin{equation}
F^\beta_\psi(s,\va):=\E^\beta_{s,\va}\textstyle\sum_{t\ge0}\gamma^t\psi(s_{t+1})
=\tfrac{1}{1-\gamma}\textstyle\int\psi\,\mathrm dm_\beta.
\label{eq:abi-def}
\end{equation}
For any task vector $z\in\R^d$ and linear reward $r_z(x)=\psi(x)^\top z$, \cref{eq:abi-def} gives $Q^\beta_{r_z}(s,\va)=F^\beta_\psi(s,\va)^\top z$ and $J^\beta_{r_z}(s,\va)=(1-\gamma)F^\beta_\psi(s,\va)^\top z$ \cite{dayan1993improving,barreto2017successor}, so the learned representation remains shared across tasks and policies.

\subsection{Reward-free representation pretraining}
\label{sec:method-latent}
\label{sec:method-fixed}
\label{sec:method-arch}

To learn $F^\beta_\psi$ in \cref{eq:abi-def} without rewards or an actor, we split the discounted sum after the $H$ arrivals of the initial chunk, which defines the backup $\T^\beta$:
\begin{equation}
(\T^\beta F)(s,\va)=\E_{s,\va}\big[\textstyle\sum_{j=1}^{H}\gamma^{j-1}\psi(s_j)+\gamma^H F(s_H,\va')\big],
\label{eq:successor}
\end{equation}
where $\va'\sim\beta(\cdot\mid s_H)$. As shown in \cref{prop:contraction}, for fixed bounded features, $\T^\beta$ is a $\gamma^H$-contraction with unique fixed point $F^\beta_\psi$, extending the fixed-behavior contraction result \cite{zheng2026can} to $H$-step chunks.

Building on TD-JEPA's joint-embedding predictive architecture \cite{bagatella2026td}, we parameterize conditioning and arrival encoders $\phi(s_0)$ and $\psi(s_j)$ for proprioceptive states or pixels \cite{yarats2021image}, paired with a task-free predictor $T_\phi(\phi(s_0),\va)$ and stop-gradient exponential moving-average (EMA) target networks $(\overline\phi,\overline\psi,\overline T_\phi)$. Instead of querying a learned actor, we aggregate all $H$ within-chunk arrivals and replace actor-sampled bootstrap actions with the next recorded chunk $\va'$. On offline tuples $(s_0,\va,s_{1:H},\va')\sim\mathcal D$, we minimize
\begin{equation}
\begin{aligned}
\mathcal L_{\mathrm{TD}}&=\E_{\mathcal D}\norm{T_\phi(\phi(s_0),\va)-y}_2^2,\\
y&=(1-\gamma^H)\overline\psi_H
+\gamma^H\overline T_\phi(\overline\phi(s_H),\va'),
\end{aligned}
\label{eq:td-loss}
\end{equation}
where $\overline\psi_H:=\sum_{j=1}^{H}\gamma^{j-1}\overline\psi(s_j)/\sum_{j=1}^{H}\gamma^{j-1}$ is the normalized discounted arrival mean satisfying $\mbox{$(1-\gamma^H)\overline\psi_H$}=(1-\gamma)\sum_{j=1}^{H}\gamma^{j-1}\overline\psi(s_j)$. Observed arrivals anchor within-chunk consequences while $\overline T_\phi$ bootstraps continuation beyond $s_H$, training $T_\phi$ to estimate $(1-\gamma)F^\beta_\psi$.

To prevent representational collapse in \cref{eq:td-loss}, we regularize both encoders with variance-covariance penalties \cite{bardes2022vicreg}, a symmetric prediction loss swapping $\phi$ and $\psi$, and auxiliary state-reconstruction grounding on proprioceptive inputs. Architectural details are given in \cref{app:impl}. Conditioning $T_\phi$ on $H$ across a horizon set $\mathcal H$ with length-matched bootstrap chunks $\va'$ trains a single multi-horizon predictor that supports queries at any $H\in\mathcal H$, as formalized in \cref{cor:multi-h}. After pretraining, we freeze the EMA networks, denoted $(\phi,\psi,T_\phi)$ below.

\subsection{Task readouts for chunk scoring}
\label{sec:method-readout}

Following standard zero-shot RL task inference \cite{touati2022does,bagatella2026td}, a downstream task is specified at test time by evaluating its reward or goal indicator on a cached subset of offline states $\{s'_i\}$. Paired with a task readout on the frozen network weights, pretrained \method{} directly forms a chunk critic $\hJ(s,\va)$ without gradient updates. \method{} pairs the successor backbone of \cref{eq:td-loss} with a closed-form linear readout $\hJ_\psi$ for dense rewards, and supports a categorical density readout $\hJ_D$ \cite{anonymous2026density} for sparse state-space goals.

\paragraph{Feature-moment readout.}
The linear readout computes $\hJ_\psi(s,\va)=T_\phi(\phi(s),\va)^\top z$, fitting $z$ in closed form via ridge regression on $\psi(s'_i)$ for dense rewards or reward-weighted centered features for sparse goals. Requiring no auxiliary heads, this inner product estimates $J^\beta_{r_z}(s,\va)$ for the induced linear reward $r_z=\psi^\top z$. \Cref{app:score-budgets} decomposes the resulting score error against $J^\beta_r(s,\va)$.

\paragraph{Density readout.}
When low-rank linear projections blur localized goal indicators on proprioceptive states, $\hJ_D$ trains lightweight heads \cite{anonymous2026density} on frozen chunk and arrival features in a second reward-free stage via distributional TD at discount $\gamma^H$, with backbone target $y=\overline\psi_H$ in \cref{eq:td-loss} on sparse-goal state suites, as detailed in \cref{app:density-theory,app:within-chunk}. At test time, evaluating these heads over $\{s'_i\}$ yields occupancy weights $p(i\mid s,\va)$ to compute the score $\hJ_D(s,\va)=\sum_i p(i\mid s,\va)r(s'_i)$ without gradient updates.

\subsection{Ranking-preserving reuse}
\label{sec:method-limits}

Reusing one frozen critic $\hJ$ across policies requires ranking candidate chunks from a proposer by their value under a target continuation policy $\pi$, even though $\hJ$ evaluates continuation under $\beta$. Let $V^\pi_r(s):=\E^\pi_s\sum_{t\ge0}\gamma^t r(s_{t+1})$ denote the unnormalized state value under $\pi$. With $(s,r)$ fixed and suppressed, write $J^\pi(\va)$ for the normalized value of executing $\va$ in full before following $\pi$, $e(\va):=\hJ(\va)-J^\beta(\va)$ for the behavior-score error, and let $u(\va):=\E_{s,\va}\Delta V(s_H)$ with $\Delta V:=V^\beta_r-V^\pi_r$. Since $J^\beta$ and $J^\pi$ share the initial chunk $\va$, within-chunk rewards cancel:
\begin{equation}
\hJ(\va)-J^\pi(\va)
=e(\va)+(1-\gamma)\gamma^H u(\va).
\label{eq:reuse-split}
\end{equation}
Only variation across a candidate set $\mathcal A_s$ affects ordering: with critic-error spread $\eta(\mathcal A_s):=\max_{\mathcal A_s}e-\min_{\mathcal A_s}e$, define the continuation spread
\begin{equation}
\delta_\pi(\mathcal A_s):=(1-\gamma)\gamma^H\big(\max_{\mathcal A_s}u-\min_{\mathcal A_s}u\big),
\label{eq:gap-terms}
\end{equation}
and gap-error budget $\varepsilon_\Delta:=\eta(\mathcal A_s)+\delta_\pi(\mathcal A_s)$.

\begin{proposition}[Ranking preservation]
\label{prop:ranking}
For any bounded reward $r$, critic score $\hJ$, stationary continuation policy $\pi$ from $s_H$, and nonempty finite candidate set $\mathcal A_s$ at state $s$, every pair $\va,\vb\in\mathcal A_s$ satisfies $\big|[\hJ(\va)-\hJ(\vb)]-[J^\pi(\va)-J^\pi(\vb)]\big|\le\varepsilon_\Delta$. Hence any gap exceeding $\varepsilon_\Delta$ keeps its sign under both scores. In particular, $\hJ(\va)-\hJ(\vb)>\varepsilon_\Delta$ implies $J^\pi(\va)>J^\pi(\vb)$.
\end{proposition}

\Cref{prop:ranking}, proved in \cref{app:proof-ranking}, refines approximate-greedy bounds \cite{singh1994upper} from uniform error to candidate-set spreads and clarifies cross-policy critic reuse. First, for a fixed candidate set $\mathcal A_s$, $\eta(\mathcal A_s)$ cancels state-only score offsets and does not depend on $\pi$, so changing $\pi$ alters only $\delta_\pi(\mathcal A_s)$. Second, $\delta_\pi(\mathcal A_s)$ carries a factor $\gamma^H$, cancels any continuation gap shared across candidates, and depends only on how $\Delta V(s_H)$ varies across candidate endpoints, as \cref{rem:eps-q-compare,ex:two-branch} illustrate. We evaluate chunk ranking and cross-policy reuse in \cref{sec:exp-swap,sec:exp-reuse}.

\subsection{Test-time guidance and policy improvement}
\label{sec:method-guidance}
\label{sec:method-improvement}
\label{sec:method-gradient}

At test time, the frozen critic $\hJ$ guides an independent proposer $\mu$ without updating policy parameters, converting the ordering condition of \cref{prop:ranking} into policy improvement under candidate selection, rolling replanning, and data-supported gradient steps. Best-of-$N$ (\textbf{BFN}) picks the top chunk among $N$ draws from $\mu(\cdot\mid s)$, a variant of Q-guided flow (\textbf{QGF}) \cite{zhou2026test} projects endpoint critic gradients onto low temporal frequencies and normalizes them as detailed in \cref{app:qgf-impl}, and \textbf{QGF+BFN4} selects from four guided draws and one unguided draw.

\begin{table*}[t]
\centering
\caption{\textbf{From states, \method{} guides a weak Flow BC policy to match or surpass actor-based zero-shot RL without training an actor.} Return on ExORL (DMC) and success rate (\%) on OGBench. \emph{With task-conditioned actor training:} eight joint actor--critic methods from \citet{bagatella2026td}, Laplacian (Lap.)~\cite{wu2019laplacian}, ICVF~\cite{ghosh2023reinforcement}, HILP~\cite{park2024foundation}, FB~\cite{touati2021learning}, RLDP~\cite{jajoo2026regularized}, BYOL~\cite{grill2020bootstrap}, BYOL-$\gamma$~\cite{lawson2025self}, and TD-JEPA, plus one-step FB~\cite{zheng2026can}, which trains an actor from a behavior critic. $^\star$Adapted from self-supervised learning. $^\dagger$Reproduced with released code, detailed in \cref{app:zs-baselines}. \emph{Frozen Flow BC (no actor training):} unguided Flow BC (gray) and \method{} guidance (green) via BFN16, QGF, or QGF+BFN4 under rolling execution. Our runs use 3 seeds, with s.e.\ in \cref{tab:zeroshot-full}. \textbf{Bold} marks the highest mean per group.}
\label{tab:zeroshot}
\footnotesize
\setlength{\tabcolsep}{3.55pt}
\setlength{\aboverulesep}{0pt}\setlength{\belowrulesep}{0pt}\renewcommand{\arraystretch}{1.15}%
\begin{tabular}{@{\hspace{2pt}}l ccccccccc >{\columncolor{baseblock}}c *{2}{>{\columncolor{abiblock}}c}>{\columncolor{abiblock}[\tabcolsep][3pt]}c@{\hspace{3pt}}}
\toprule
 & \multicolumn{9}{c}{With task-conditioned actor training} & \multicolumn{4}{c}{Frozen Flow BC (no actor training)} \\
\cmidrule(lr){2-10}\cmidrule{11-14}
 & \multicolumn{8}{c}{\emph{Joint actor--critic}} & \emph{Behavior critic} & \multicolumn{1}{>{\columncolor{baseblock}}c}{\emph{Unguided}} & \multicolumn{3}{>{\columncolor{abiblock}[\tabcolsep][3pt]}c@{\hspace{3pt}}}{\textbf{$+$ \method{} critic (ours)}} \\
\cmidrule(lr){2-9}\cmidrule(lr){10-10}\cmidrule{11-14}
Dataset & Lap. & ICVF$^\star$ & HILP & FB & RLDP & BYOL$^\star$ & BYOL-$\gamma^\star$ & TD-JEPA & One-step FB\rlap{$^\dagger$} & Flow BC & BFN16 & QGF & QGF+BFN4 \\
\midrule
walker & 769.7 & 727.0 & 796.4 & \textbf{811.5} & 723.9 & 746.8 & 786.1 & 785.2 & 128.9 & 158.2 & 628.8 & 717.6 & \textbf{781.3} \\
cheetah & 614.5 & 606.3 & 618.3 & 672.7 & 575.6 & 622.8 & 647.2 & \textbf{688.7} & 364.2 & 103.5 & 566.0 & \textbf{671.5} & 670.7 \\
quadruped & 635.0 & \textbf{708.5} & 694.8 & 595.6 & 665.0 & 611.8 & 683.1 & 691.4 & 313.6 & 50.6 & 690.5 & 728.6 & \textbf{742.4} \\
pointmass & 345.1 & 435.5 & 371.0 & \textbf{513.0} & 476.3 & 493.0 & 465.1 & 479.3 & 57.7 & $-$78.6 & 404.8 & 400.5 & \textbf{426.5} \\
\midrule[0.4pt]
\emph{DMC (avg)} & 591.1 & 619.3 & 620.1 & 648.2 & 610.2 & 618.6 & 645.4 & \textbf{661.2} & 216.1 & 58.4 & 572.5 & 629.5 & \textbf{655.2} \\
\midrule
antmaze-mn & 50.0 & 79.8 & \textbf{83.6} & 73.0 & 74.6 & 58.4 & 51.4 & 70.4 & 47.0 & 5.0 & \textbf{72.7} & 65.7 & 68.7 \\
antmaze-ln & 21.6 & \textbf{58.4} & 52.6 & 36.8 & 36.4 & 26.6 & 21.8 & 57.2 & 19.0 & 2.3 & \textbf{40.3} & 0.7 & 33.0 \\
antmaze-ms & 21.4 & 39.0 & 50.6 & \textbf{70.4} & 58.4 & 60.6 & 45.6 & 61.6 & 31.0 & 0.3 & \textbf{72.7} & 45.3 & 67.3 \\
antmaze-ls & 11.8 & 13.2 & 12.2 & \textbf{49.8} & 19.6 & 25.8 & 20.2 & 40.6 & 4.7 & 0.0 & \textbf{9.0} & 0.0 & 5.7 \\
antmaze-me & 0.8 & 0.0 & 2.0 & \textbf{51.6} & 4.8 & 11.4 & 19.6 & 20.2 & 5.0 & 0.3 & \textbf{46.3} & 33.0 & 38.3 \\
cube-single & 15.1 & 20.4 & 74.2 & 49.6 & 19.8 & 22.0 & \textbf{79.4} & 34.2 & 4.0 & 11.0 & \textbf{77.3} & 49.0 & 56.3 \\
cube-double & 2.0 & 5.0 & \textbf{20.0} & 2.6 & 3.8 & 4.4 & 2.6 & 3.6 & 0.0 & 0.0 & 0.0 & 0.0 & 0.0 \\
scene & 7.8 & \textbf{45.4} & 43.8 & 12.8 & 11.6 & 15.4 & 14.4 & 38.4 & 8.7 & 7.0 & \textbf{16.3} & 6.3 & 5.0 \\
puzzle-3x3 & 2.8 & 16.6 & 2.8 & 4.8 & 14.6 & 13.2 & \textbf{18.8} & 15.6 & 8.0 & 0.7 & \textbf{13.0} & 9.3 & 10.3 \\
\midrule[0.4pt]
\emph{OGBench (avg)} & 14.8 & 30.9 & 38.0 & \textbf{39.0} & 27.1 & 26.4 & 30.4 & 38.0 & 14.1 & 3.0 & \textbf{38.6} & 23.3 & 31.6 \\
\bottomrule
\end{tabular}

\end{table*}

\paragraph{Candidate selection and whole-chunk execution.}
At state $s$, let $\mathcal A_s$ be a sampled candidate set containing an unguided draw $\vb\sim\mu(\cdot\mid s)$, as in both BFN and QGF+BFN4. With maximum candidate gain $G:=\max_{\mathcal A_s}J^\pi-J^\pi(\vb)\ge0$ and $(G,\varepsilon_\Delta)$ evaluated at $\pi=\mu$, selecting $\hat\va\in\argmax_{\mathcal A_s}\hJ$ yields
\begin{equation}
J^\mu(\hat\va)-J^\mu(\vb)
\ge G-\varepsilon_\Delta,
\label{eq:selection-gain}
\end{equation}
so selection improves chunk value over $\vb$ whenever $G>\varepsilon_\Delta$. Under whole-chunk execution ($H_{\mathrm{exec}}{=}H$), repeating this selection at every boundary $s_{kH}$ from initial state $s_0$ induces a stationary chunk policy $\nu$. With selection loss $L:=\max_{\mathcal A_s}J^\mu-J^\mu(\hat\va)\le\varepsilon_\Delta$ and normalized boundary-state occupancy $d^\nu_{s_0}(\cdot):=(1-\gamma^H)\E^\nu_{s_0}\sum_{k\ge0}\gamma^{Hk}\delta_{s_{kH}}(\cdot)$ of $\nu$, applying the performance-difference identity \cite{kakade2002approximately} in \cref{prop:headroom} gives
\begin{equation}
(1-\gamma)\big(V^\nu_r-V^\mu_r\big)(s_0)
=\frac{\E_{s\sim d^\nu_{s_0}}\E\big[G-L\mid s\big]}{1-\gamma^H},
\label{eq:closed-loop-gain}
\end{equation}
where the inner expectation averages over draws at $s$, so discounted closed-loop return improves whenever $\E[G-L\mid s]>0$ across visited boundaries.

\paragraph{Finite-chunk replanning.}
Under rolling execution ($H_{\mathrm{exec}}{=}1$), the agent draws $\mathcal A_s$ containing $\vb\sim\mu(\cdot\mid s)$ at each step, selects $\hat\va\in\argmax_{\mathcal A_s}\hJ$, and applies only $\hat a_0$. Let $\mu_1$ be the first-action marginal of $\mu$. Because \cref{eq:reuse-split} depends on $\pi$ only through $V^\pi_r(s_H)$, \cref{app:rolling} shows that it applies equally to the primitive-step continuation $\pi=\mu_1$. Suppressing $s$, let $J^{\mu_1}_H(\va)$ and $J^{\mu_1}_1(a_0)$ denote the normalized values of executing $\va$ versus only $a_0$ before following $\mu_1$. Their difference $\gamma\tau(\va)$ captures the discounted value gained by executing the tail $a_{1:H-1}$ over continuing with $\mu_1$ from $s_1$. Evaluating $(G,\varepsilon_\Delta)$ at $\pi=\mu_1$ in \cref{prop:r7-rolling} gives
\begin{equation}
\begin{aligned}
J^{\mu_1}_1(a_0)&=J^{\mu_1}_H(\va)-\gamma\tau(\va),\\
J^{\mu_1}_1(\hat a_0)-J^{\mu_1}_1(b_0)
&\ge G-\varepsilon_\Delta-\gamma[\tau(\hat\va)-\tau(\vb)],
\end{aligned}
\label{eq:rolling-bound}
\end{equation}
so discounted closed-loop return improves over $\mu_1$ whenever the right-hand side has positive conditional expectation along visited states. Default rolling guidance scores proposed chunks directly, whereas single-step actor guidance averages $\hJ$ over $K$ shared-noise Flow BC completions.

\paragraph{Gradient guidance and data support.}
When high-value chunks are rare under $\mu$, ascending $\nabla_{\va}\hJ$ moves proposals beyond finite candidate samples, provided offline data cover local action neighborhoods. Formally, \cref{prop:nonid} shows that the TD objective constrains scores and action gradients only around covered pairs $(s,\va)$, leaving off-support gradients unconstrained. Conversely, on covered neighborhoods with smooth score-error and value gradients, \cref{cor:ranking-gradient,cor:fixed-step,prop:flow} establish that a small neighborhood gap budget $\varepsilon_\Delta$ bounds action-gradient error, so idealized short subspace gradient steps increase the value $J^\pi(\va)$ once the projected gradient exceeds the action-gradient~and~flow~\mbox{errors}.

\providecommand{\tbd}{\textcolor{red}{TBD}}
\newcommand{\zsDmcOneStep}{216.1$\,\pm\,$17.9}
\newcommand{\zsDmcOneStepBfn}{137.4$\,\pm\,$3.6}
\newcommand{\zsDmcOneStepQgf}{152.8$\,\pm\,$13.2}
\newcommand{\zsDmcBc}{58.4$\,\pm\,$3.5}
\newcommand{\zsDmcAbiBfn}{572.5$\,\pm\,$0.3}
\newcommand{\zsDmcAbiQgf}{629.5$\,\pm\,$2.8}
\newcommand{\zsDmcAbiQgfBfn}{655.2$\,\pm\,$4.2}
\newcommand{\zsDmcAbiActor}{\tbd}
\newcommand{\zsDmcAbiActorChunk}{\tbd}
\newcommand{\zsDmcTdjepa}{661.2$\,\pm\,$6.3}
\newcommand{\zsDmcFb}{648.2$\,\pm\,$4.1}
\newcommand{\zsOgOneStep}{14.1$\,\pm\,$0.8}
\newcommand{\zsOgOneStepBfn}{4.1$\,\pm\,$0.7}
\newcommand{\zsOgOneStepQgf}{1.6$\,\pm\,$0.2}
\newcommand{\zsOgBc}{3.0$\,\pm\,$0.4}
\newcommand{\zsOgAbiBfn}{38.6$\,\pm\,$0.6}
\newcommand{\zsOgAbiQgf}{23.3$\,\pm\,$0.3}
\newcommand{\zsOgAbiQgfBfn}{31.6$\,\pm\,$0.9}
\newcommand{\zsOgAbiActor}{\tbd}
\newcommand{\zsOgAbiActorChunk}{\tbd}
\newcommand{\zsOgTdjepa}{38.0$\,\pm\,$0.8}
\newcommand{\zsOgFb}{39.0$\,\pm\,$0.7}
\newcommand{\zsReproCheetah}{562.0}
\newcommand{\zsReproCheetahSeeds}{1}
\newcommand{\zsReproScene}{10.0$\,\pm\,$1.2}
\newcommand{\zsReproSceneSeeds}{3}
\newcommand{\zsDmcWinsBfn}{3}
\newcommand{\zsDmcWinsQgf}{1}
\newcommand{\zsActorCells}{0}
\newcommand{\zsDmcAbiProtBfn}{\tbd}
\newcommand{\zsDmcAbiProtQgf}{\tbd}
\newcommand{\zsDmcAbiProtQgfBfn}{\tbd}
\newcommand{\zsOgAbiProtBfn}{\tbd}
\newcommand{\zsOgAbiProtQgf}{\tbd}
\newcommand{\zsOgAbiProtQgfBfn}{\tbd}
\newcommand{\zsDmcWinsBfnProt}{0}
\newcommand{\zsDmcWinsQgfProt}{1}
\newcommand{\zsDmcAbiFinBfn}{257.6$\,\pm\,$13.6}
\newcommand{\zsDmcAbiFinQgf}{300.5$\,\pm\,$25.0}
\newcommand{\zsDmcAbiFinQgfBfn}{309.5$\,\pm\,$22.6}
\newcommand{\zsDmcAbiFinActor}{111.1$\,\pm\,$7.8}
\newcommand{\zsDmcAbiFinActorChunk}{96.0$\,\pm\,$6.3}
\newcommand{\zsOgAbiFinBfn}{2.7$\,\pm\,$0.4}
\newcommand{\zsOgAbiFinQgf}{0.1$\,\pm\,$0.1}
\newcommand{\zsOgAbiFinQgfBfn}{0.1$\,\pm\,$0.1}
\newcommand{\zsOgAbiFinActor}{0.2$\,\pm\,$0.2}
\newcommand{\zsOgAbiFinActorChunk}{0.1$\,\pm\,$0.1}
\newcommand{\zsDmcWinsBfnFin}{4}
\newcommand{\zsDmcWinsQgfFin}{4}
\newcommand{\zsAbiMinSeeds}{3}
\newcommand{\zsAbiProtMinSeeds}{0}

\renewcommand{\zsDmcAbiBfn}{572.5$\,\pm\,$0.3}
\renewcommand{\zsDmcAbiQgf}{629.5$\,\pm\,$2.8}
\renewcommand{\zsDmcAbiQgfBfn}{655.2$\,\pm\,$4.2}
\renewcommand{\zsOgAbiBfn}{38.6$\,\pm\,$0.6}
\renewcommand{\zsOgAbiQgf}{23.3$\,\pm\,$0.3}
\renewcommand{\zsOgAbiQgfBfn}{31.6$\,\pm\,$0.9}
\renewcommand{\zsOgBc}{3.0$\,\pm\,$0.4}

\section{Experiments}
\label{sec:exp}

\begin{figure*}[t]
\centering
\includegraphics[width=\linewidth]{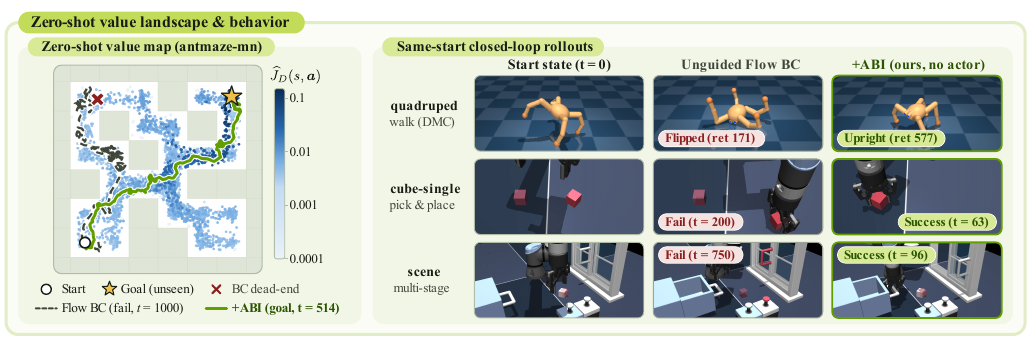}
\caption{\textbf{\method{} shapes goal-directed value landscapes and guides closed-loop rollouts.} \emph{Left:} On \texttt{antmaze-mn}, the zero-shot score $\hJ_D(s,\va)$ over recorded state--chunk pairs rises along the corridor toward an unseen goal. Unguided Flow BC stalls in a dead end, whereas \method{} guides it to the goal. \emph{Right:} From identical initial states, \method{} guidance keeps \texttt{quadruped} upright, places the block at the goal on \texttt{cube-single}, and opens both the drawer and the window on \texttt{scene}. Rollouts use BFN16 with $H_{\mathrm{exec}}{=}8$.}
\label{fig:valuemap}
\end{figure*}

We evaluate \method{} across 65 locomotion, navigation, and manipulation tasks spanning four DeepMind Control (DMC) domains from ExORL \cite{tassa2018deepmind,todorov2012mujoco,yarats2022don} and nine goal-conditioned datasets from OGBench \cite{park2025ogbench}. ExORL tests dense-reward control from broad exploratory trajectories \cite{burda2019exploration}, whereas OGBench tests sparse goal-reaching from narrow trajectories that do not demonstrate the evaluation tasks, including \texttt{play} manipulation and \texttt{antmaze} navigation and stitching suites. Following \citet{bagatella2026td}, we test proprioceptive state and $64\times64$ RGB pixel variants of all 13 datasets across three seeds, examining zero-shot policy guidance, head-to-head critic comparisons on identical proposals, cross-policy critic reuse, guidance from raw pixels, and design ablations.

On each unlabeled dataset, we independently pretrain \method{} across horizons $\mathcal H{=}\{4,8,16,32\}$ without rewards, task inputs, or policy training, and separately train an 8-step flow-matching behavior-cloning proposer (\textbf{Flow BC}). At deployment, both models remain frozen: we query \method{} at $H{=}8$, infer each task from up to 10k cached offline states without gradient updates, and evaluate closed-loop rolling execution ($H_{\mathrm{exec}}{=}1$) on a held-out test split unless noted otherwise. On state suites, the task readout follows reward structure, using linear $\hJ_\psi$ for dense DMC rewards and density $\hJ_D$ \cite{anonymous2026density} for sparse OGBench goals, while the guidance mode follows dataset coverage as detailed in \cref{app:zs-guidance}. Pixel experiments use the linear readout $\hJ_\psi$ throughout, reusing each suite's top guidance mode from states. Full protocol and baseline details appear in \cref{app:zeroshot}.

\subsection{Zero-shot guidance of an unseen policy}
\label{sec:exp-zeroshot}

In \cref{tab:zeroshot}, \method{} guides weak Flow BC from 58.4 to 655.2 return on DMC and from 3.0\% to 38.6\% success on OGBench without training an actor, matching or surpassing multiple actor-based zero-shot RL baselines. Against one-step FB, \method{} leads on every DMC domain (655.2 vs.\ 216.1) and more than doubles OGBench success (38.6\% vs.\ 14.1\%). Guiding Flow BC also achieves the highest return among all methods on \texttt{quadruped} (742.4) and recombines non-demonstration \texttt{play} fragments on \texttt{cube-single} to surpass TD-JEPA's trained actor (77.3\% vs.\ 34.2\%).

Qualitatively, \cref{fig:valuemap} visualizes zero-shot value scoring and same-start closed-loop rollouts under whole-chunk execution ($H_{\mathrm{exec}}{=}8$, BFN16), as detailed in \cref{app:zs-abi}. On \texttt{antmaze-mn}, $\hJ_D(s,\va)$ over recorded state--chunk pairs rises along the winding corridor toward an unseen goal, guiding Flow BC out of the dead end where unguided rollouts stall. Across control suites, \method{} guidance from identical initial states prevents \texttt{quadruped} from flipping over (577 vs.\ 171 return), completes pick-and-place on \texttt{cube-single}, and opens both the drawer and the window in sequence on multi-stage \texttt{scene}.

Whether gradient guidance improves over candidate selection tracks dataset coverage. On broad exploratory ExORL trajectories, adding gradient steps raises DMC return from 572.5 under BFN16 to 629.5 under QGF and 655.2 under QGF+BFN4. Conversely, on narrow OGBench trajectories where gradient steps leave the data support, hyperparameter-free in-support selection with BFN16 performs best at 38.6\%.

\paragraph{Pretraining efficiency.}
In \cref{fig:teaser}, removing the joint actor-optimization loop lets \method{} reach reproduced TD-JEPA's full-budget DMC return with $24\times$ fewer representation pretraining samples (128M vs.\ 3072M). In \cref{app:zs-convergence}, it also outperforms reproduced TD-JEPA and one-step FB at equal or larger sample budgets on DMC and OGBench.

\subsection{Comparing critics on a fixed policy}
\label{sec:exp-swap}

\begin{table}[t]
\centering
\caption{\textbf{On identical proposals, the \method{} critic achieves the highest suite averages under both execution modes.} Under BFN16, each critic scores the same frozen Flow BC proposals with its native task inference and action input (8-step chunks for \method{}, single actions for the others). $H_{\mathrm{exec}}{=}1$ replans after every action, and $H_{\mathrm{exec}}{=}8$ executes the whole chunk. Mean $\pm$ s.e.\ over 3 seeds. \textbf{Bold} marks the highest mean per column.}
\label{tab:zeroshot-swap}
\begingroup\scriptsize\setlength{\tabcolsep}{0pt}\setlength{\aboverulesep}{0pt}\setlength{\belowrulesep}{0pt}\renewcommand{\arraystretch}{1.18}%
\begin{tabularx}{\columnwidth}{@{}>{\hspace{2pt}}l *{4}{>{\centering\arraybackslash}X}@{}}
\toprule
 & \multicolumn{2}{c}{DMC (return)} & \multicolumn{2}{c}{OGBench (success, \%)} \\
\cmidrule(lr){2-3}\cmidrule(lr){4-5}
Critic (BFN16) & $H_{\mathrm{exec}}{=}1$ & $H_{\mathrm{exec}}{=}8$ & $H_{\mathrm{exec}}{=}1$ & $H_{\mathrm{exec}}{=}8$ \\
\midrule
Unguided Flow BC & 58.4{\fontsize{6}{7}\selectfont$\,\pm\,$3.5} & 63.3{\fontsize{6}{7}\selectfont$\,\pm\,$3.7} & 3.0{\fontsize{6}{7}\selectfont$\,\pm\,$0.4} & 2.1{\fontsize{6}{7}\selectfont$\,\pm\,$0.5} \\
\rowcolor{abiblock}\textbf{\method{} critic (ours)} & \textbf{572.5}{\fontsize{6}{7}\selectfont$\,\pm\,$0.3} & \textbf{541.9}{\fontsize{6}{7}\selectfont$\,\pm\,$3.6} & \textbf{38.6}{\fontsize{6}{7}\selectfont$\,\pm\,$0.6} & \textbf{29.2}{\fontsize{6}{7}\selectfont$\,\pm\,$1.0} \\
TD-JEPA critic & 555.6{\fontsize{6}{7}\selectfont$\,\pm\,$0.8} & 191.8{\fontsize{6}{7}\selectfont$\,\pm\,$9.1} & 16.7{\fontsize{6}{7}\selectfont$\,\pm\,$0.3} & 11.7{\fontsize{6}{7}\selectfont$\,\pm\,$0.1} \\
One-step FB critic & 190.2{\fontsize{6}{7}\selectfont$\,\pm\,$20.4} & 143.5{\fontsize{6}{7}\selectfont$\,\pm\,$1.8} & 12.8{\fontsize{6}{7}\selectfont$\,\pm\,$0.9} & 7.6{\fontsize{6}{7}\selectfont$\,\pm\,$0.4} \\
\bottomrule
\end{tabularx}
\endgroup

\end{table}

\begin{figure*}[t]
\centering
\includegraphics[width=\linewidth,trim=0 5bp 0 4bp,clip]{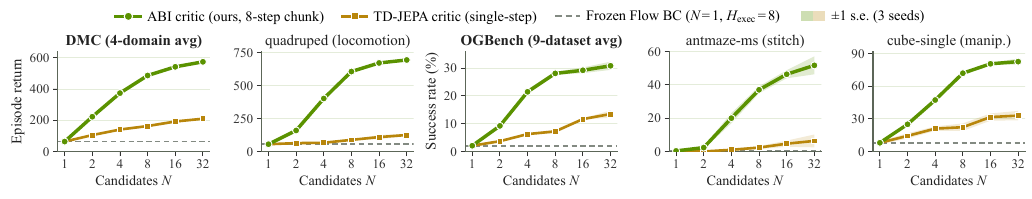}
\caption{\textbf{Under whole-chunk execution ($H_{\mathrm{exec}}{=}8$), candidate scaling requires a chunk-level critic.} Best-of-$N$ selection over 8-step frozen Flow BC proposals executed open loop, with values in \cref{tab:zeroshot-nsweep}. Because TD-JEPA's single-step critic cannot evaluate multi-step action sequences, it gains little from larger $N$: on DMC, \method{} at $N{=}2$ already exceeds TD-JEPA at $N{=}32$.}
\label{fig:n-scaling}
\par\nointerlineskip\vspace{\dblfloatsep}
\begin{minipage}[t]{\columnwidth}
\vspace{0pt}
\centering
\includegraphics[width=\linewidth,trim=0 2bp 0 3bp,clip]{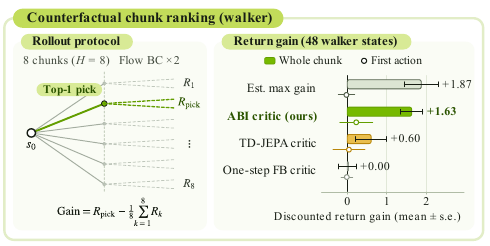}
\caption{\textbf{On \texttt{walker}, the \method{} critic's top-scored chunk approaches the estimated maximum gain.} Discounted return gain over a random pick among eight Flow BC chunks for whole-chunk (bars) and first-action (circles) execution. \emph{Est.\ max gain} picks the highest-return chunk on one rollout and evaluates it on an independent rollout. Mean $\pm$ s.e.\ over 3 seeds, with 95\% bootstrap intervals in \cref{tab:zs-rank}.}
\label{fig:rank}
\end{minipage}\hfill
\begin{minipage}[t]{\columnwidth}
\vspace{0pt}
\centering
\captionsetup{type=table,position=top}
\caption{\textbf{One frozen \method{} critic improves external single-step RL actors at test time.} At every step, the \method{} critic ranks $N{=}16$ action proposals from the external actor via $K{=}4$ shared-noise Flow BC completions with 8-step scores and $H_{\mathrm{exec}}{=}1$, as detailed in \cref{app:zs-steer-actors}. Mean $\pm$ s.e.\ over 3 seeds on the test split. $^\dagger$Reproduced with the authors' released code. \textbf{Bold} marks the highest mean per block.}
\label{tab:zeroshot-actors}
\begingroup\footnotesize\setlength{\tabcolsep}{0pt}\setlength{\aboverulesep}{0pt}\setlength{\belowrulesep}{0pt}\renewcommand{\arraystretch}{1.30}%
\begin{tabularx}{\columnwidth}{@{}l *{2}{>{\centering\arraybackslash}X}@{}}
\toprule
External policy / selector & DMC (return) & OGBench (success, \%) \\
\midrule
\multicolumn{3}{@{}l}{\emph{TD-JEPA single-step actor \cite{bagatella2026td}}$^\dagger$} \\
\hspace{4pt}Unguided actor & 638.9{\fontsize{7}{8}\selectfont$\,\pm\,$9.3} & 30.3{\fontsize{7}{8}\selectfont$\,\pm\,$1.2} \\
\hspace{4pt}Random pick ($N{=}16$) & 630.8{\fontsize{7}{8}\selectfont$\,\pm\,$11.8} & 31.7{\fontsize{7}{8}\selectfont$\,\pm\,$1.3} \\
\rowcolor{abiblock}\hspace{4pt}\textbf{$+$ \method{} critic (ours)} & \textbf{646.8}{\fontsize{7}{8}\selectfont$\,\pm\,$8.8} & \textbf{39.7}{\fontsize{7}{8}\selectfont$\,\pm\,$0.6} \\
\midrule
\multicolumn{3}{@{}l}{\emph{One-step FB single-step actor \cite{zheng2026can}}$^\dagger$} \\
\hspace{4pt}Unguided actor & 216.1{\fontsize{7}{8}\selectfont$\,\pm\,$17.9} & 14.1{\fontsize{7}{8}\selectfont$\,\pm\,$0.8} \\
\hspace{4pt}Random pick ($N{=}16$) & 211.6{\fontsize{7}{8}\selectfont$\,\pm\,$16.4} & 12.9{\fontsize{7}{8}\selectfont$\,\pm\,$0.2} \\
\rowcolor{abiblock}\hspace{4pt}\textbf{$+$ \method{} critic (ours)} & \textbf{243.5}{\fontsize{7}{8}\selectfont$\,\pm\,$23.4} & \textbf{17.0}{\fontsize{7}{8}\selectfont$\,\pm\,$0.4} \\
\bottomrule
\end{tabularx}
\endgroup

\end{minipage}
\end{figure*}

To isolate critic quality from policy training, \cref{tab:zeroshot-swap} evaluates zero-shot critics on identical frozen Flow BC proposals under BFN16. Across both suites, the \method{} critic delivers the largest average improvement: it more than doubles baseline critic success on OGBench (38.6\% vs.\ 16.7\% for TD-JEPA and 12.8\% for one-step FB) and achieves the highest average DMC return (572.5 vs.\ 555.6 for TD-JEPA). This lead also holds under QGF and QGF+BFN4 in \cref{app:zs-swap}.

As shown in \cref{tab:zeroshot-swap}, executing full 8-step chunks open loop ($H_{\mathrm{exec}}{=}8$), as in chunked policy deployment \cite{zhao2023learning,chi2023diffusion,black2024pi0}, separates chunk and single-step critics most sharply. Because single-step critics cannot evaluate multi-step action sequences, TD-JEPA's critic retains just 35\% of its rolling DMC return (191.8 of 555.6), whereas the \method{} critic retains 95\% on DMC (541.9 of 572.5) and more than doubles both baselines on OGBench (29.2\% vs.\ 11.7\% and 7.6\%). \Cref{fig:n-scaling} shows that scaling the candidate budget $N$ widens this gap further: on DMC, \method{} at $N{=}2$ (222.4) already exceeds TD-JEPA's critic at $N{=}32$ (209.6).

\paragraph{Counterfactual chunk ranking.}
\label{sec:exp-rank}
In \cref{fig:rank}, to measure ranking quality on environment rollouts, we execute eight frozen Flow BC chunks per state on 48 \texttt{walker} states where dense rewards resolve single-chunk return gaps. Sparse-reward domains are reported in \cref{app:zs-rank}. Under whole-chunk execution, \method{}'s top-scored chunk gains $+1.63$ return over a random pick, corresponding to $G-L$ in \cref{sec:method-improvement} and capturing 87\% of the estimated maximum gain $G$ of $+1.87$, vs.\ $+0.60$ (32\%) for TD-JEPA's critic.

\subsection{Reusing one frozen critic across policies}
\label{sec:exp-reuse}

\paragraph{Guiding external trained actors.}
As shown in \cref{tab:zeroshot-actors}, a single \method{} critic pretrained without an actor also improves external actors trained with their own critics. Scoring 16 proposals via shared-noise Flow BC completions, the frozen critic of \cref{tab:zeroshot} lifts reproduced TD-JEPA's actor from 30.3\% to 39.7\% on OGBench, against 31.7\% for a random pick. It likewise improves one-step FB's actor over the unguided actor and a random pick, lifting DMC return from 216.1 to 243.5 and OGBench success from 14.1\% to 17.0\%.
\looseness=-1 %

\paragraph{Guiding a stronger base proposer.}
As shown in \cref{fig:teaser}, the same frozen critic converts a stronger proposer directly into higher task success without retraining. On OGBench, we replace the 2-layer Flow BC with the larger 4-layer one used in \cref{tab:zeroshot}, raising unguided success from 1.4\% to 3.0\%. Under the identical frozen \method{} critic, BFN16 success rises from 32.3\% to 38.6\%, as detailed in \cref{app:zs-flowbcplus}. On visual OGBench under whole-chunk execution, the same upgrade lifts success from 21.1\% to 31.6\%.

\begin{table}[t]
\centering
\caption{\textbf{From raw pixels, \method{} substantially improves frozen visual Flow BC on both suites.} Return on DMC-RGB (QGF+BFN4) and success (\%) on visual OGBench (BFN16) with $\hJ_\psi$ and $H_{\mathrm{exec}}{=}1$. Mean over 3 seeds with $\pm$ s.e.\ on suite averages and full sweep in \cref{tab:zeroshot-pixels-full}. \textbf{Bold} marks the highest mean per row.}
\label{tab:zeroshot-pixels}
\begingroup\scriptsize\setlength{\tabcolsep}{0pt}\setlength{\aboverulesep}{0pt}\setlength{\belowrulesep}{0pt}\renewcommand{\arraystretch}{1.12}%
\begin{tabularx}{\columnwidth}{@{}>{\hspace{2pt}}l >{\hsize=.9\hsize\centering\arraybackslash}X >{\columncolor{baseblock}\hsize=.75\hsize\centering\arraybackslash}X >{\columncolor{abiblock}\hsize=1.35\hsize\centering\arraybackslash}X@{}}
\toprule
 & \emph{Trained actor} & \multicolumn{2}{c}{\emph{Frozen Flow BC}} \\
\cmidrule(lr){2-2}\cmidrule{3-4}
Dataset & TD-JEPA~\cite{bagatella2026td} & Unguided & \textbf{$+$ \method{} critic (ours)} \\
\midrule
walker & \textbf{738.9} & 159.2 & 357.1 \\
cheetah & \textbf{706.0} & 115.7 & 663.2 \\
quadruped & \textbf{626.7} & 212.7 & 242.5 \\
pointmass & \textbf{443.7} & $-$71.9 & 130.8 \\
\midrule[0.4pt]
\emph{DMC-RGB (avg)} & \textbf{628.8} & 103.9{\fontsize{6}{7}\selectfont$\,\pm\,$5.7} & 348.4{\fontsize{6}{7}\selectfont$\,\pm\,$8.9} \\
\midrule
antmaze-mn & \textbf{96.7} & 20.7 & 82.0 \\
antmaze-ln & \textbf{74.6} & 8.7 & 44.0 \\
antmaze-ms & \textbf{84.4} & 4.7 & 16.7 \\
antmaze-ls & \textbf{28.8} & 0.0 & 0.7 \\
antmaze-me & \textbf{0.2} & 0.0 & 0.0 \\
cube-single & 67.8 & 1.3 & \textbf{82.7} \\
cube-double & 3.0 & 1.3 & \textbf{15.3} \\
scene & 14.2 & 6.0 & \textbf{38.7} \\
puzzle-3x3 & 2.4 & 4.7 & \textbf{11.3} \\
\midrule[0.4pt]
\emph{Visual OGBench (avg)} & \textbf{41.3} & 5.3{\fontsize{6}{7}\selectfont$\,\pm\,$0.3} & 32.4{\fontsize{6}{7}\selectfont$\,\pm\,$0.8} \\
\bottomrule
\end{tabularx}
\endgroup

\par\nointerlineskip\vspace{\floatsep}
\includegraphics[width=\columnwidth]{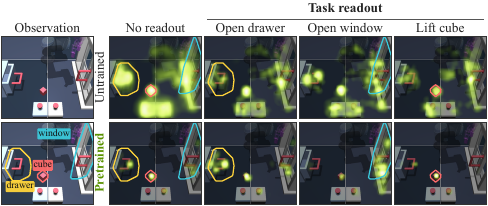}
\captionof{figure}{\textbf{Pretraining grounds the representation in manipulable objects, and each task readout shifts sensitivity to its target object.} Occlusion sensitivity on visual \texttt{scene} for the pixel critic of \cref{tab:zeroshot-pixels}, detailed in \cref{app:zs-pixel-focus}.}
\label{fig:pixel-focus}
\end{table}

\subsection{Visual guidance from raw pixels}
\label{sec:exp-pixels}

On stacked $64\times64$ RGB frames, \method{} trains a convolutional encoder end to end and deploys $\hJ_\psi$ with each suite's top guidance from \cref{tab:zeroshot}. As shown in \cref{tab:zeroshot-pixels}, this lifts visual Flow BC from 103.9 to 348.4 return on DMC-RGB and from 5.3\% to 32.4\% success on visual OGBench. Across all four visual manipulation datasets, \method{} surpasses TD-JEPA's pixel actor (37.0\% vs.\ 21.9\% on average) and every other pixel zero-shot RL baseline in \citet{bagatella2026td} (up to 22.4\% on average), reaching 82.7\% vs.\ 67.8\% on \texttt{cube-single} (93.3\% at $H_{\mathrm{exec}}{=}8$), 38.7\% vs.\ 14.2\% on \texttt{scene}, 15.3\% vs.\ 3.0\% on \texttt{cube-double}, and 11.3\% vs.\ 2.4\% on \texttt{puzzle-3x3}. Pairing multi-step action chunks with chunk-level value selection is thus well suited to contact-rich visuomotor manipulation, while the remaining gaps from state results reflect both the frozen visual proposer's candidate coverage, as the proposer upgrade in \cref{sec:exp-reuse} shows, and end-to-end visual value estimation with~a~linear~\mbox{readout}.

In \cref{fig:pixel-focus}, occlusion-sensitivity probing on visual \texttt{scene} shows that reward-free pretraining alone concentrates the shared representation $T_\phi$ on manipulable objects, whereas an untrained encoder exhibits diffuse sensitivity. Applying different linear readouts to this frozen $T_\phi$ then shifts score sensitivity toward each task's target object across the 64 probed frames in \cref{tab:zs-pixel-focus}.

\subsection{Ablation study}
\label{sec:exp-design}
\label{sec:exp-ablation}

We ablate \method{}'s representation and readout choices on state suites under whole-chunk execution ($H_{\mathrm{exec}}{=}8$) in \cref{tab:zeroshot-variants}.

\paragraph{Multi-horizon pretraining.}
Supervising one shared predictor across $\mathcal H{=}\{4,8,16,32\}$ couples representation learning across short and long chunk horizons: even when queried solely at $H{=}8$ with the density readout, it outperforms a fixed-$H{=}8$ predictor under BFN16 on both DMC (515.8 vs.\ 475.7) and long-horizon mazes (\texttt{antmaze-mn}: 41.0\% vs.\ 17.7\%). Under rolling execution, single-step \method{} ($H{=}1$) lifts DMC Flow BC from 58.4 to 497.9 across all domains, and multi-horizon pretraining raises it to 572.5, as reported in \cref{app:zs-lengths}.

\paragraph{Readout.}
In the controlled 2-layer Flow BC ablations, switching from the density readout to the linear readout $\hJ_\psi$ raises dense-reward DMC return under BFN16 and QGF. Conversely, on sparse-goal OGBench, the density readout beats the linear one across all guidance modes.

\section{Conclusion}
\label{sec:conclusion}

We presented Actor-Free Behavior-Induced Representations (\method{}), a zero-shot reinforcement learning framework that decouples multi-step value learning from both reward supervision and actor training. By learning multi-horizon action-chunk representations under the data-collecting behavior and pairing them with zero-update task readouts at test time, \method{} forms a chunk critic with theoretical conditions for cross-policy ranking preservation, closed-loop selection improvement, and data-supported gradient guidance. Across diverse benchmarks, a single pretrained \method{} representation per dataset guides weak imitation policies to match or surpass zero-shot RL baselines with high sample efficiency, outperforms existing zero-shot critics on shared proposals, and transfers across externally trained RL actors without retraining. These results show that reward-free behavior representations provide test-time guidance across tasks and unseen policies without policy optimization.

\paragraph{Limitations.}
As analyzed in \cref{sec:method-limits,sec:method-guidance,app:limitations}, \method{}'s improvement bounds rely on offline coverage of queried action chunks and bounded continuation discrepancy across candidate endpoints, so gradient guidance remains constrained on narrow datasets. Moreover, while our experiments evaluate single-objective episodes, composing task readouts across multi-stage tasks, strengthening reward-free visual representations for complex scenes, and scaling actor-free chunk guidance to large-scale visuomotor policies remain promising future directions.

{
    \small
    \bibliographystyle{ieeenat_fullname}
    \bibliography{main}
}

\ifsupp
\clearpage
\onecolumn
\makeatletter
\pretocmd{\@ShipoutPicture}{\@inlabelfalse}{}{}
\makeatother
\renewcommand{\topfraction}{.90}
\renewcommand{\bottomfraction}{.90}
\renewcommand{\textfraction}{.07}
\renewcommand{\floatpagefraction}{.85}
\setcounter{topnumber}{3}
\setcounter{bottomnumber}{2}
\setcounter{totalnumber}{4}
\begin{center}{\Large\textbf{Appendix}}\end{center}
\appendix
\counterwithin{proposition}{section}
\renewcommand{\theHproposition}{app.\thesection.\arabic{proposition}}
\counterwithin{corollary}{section}
\renewcommand{\theHcorollary}{app.\thesection.\arabic{corollary}}
\providecommand{\TV}{\operatorname{TV}}

\section{Notation and foundational assumptions}
\label{app:notation}

Following \cref{sec:method}, we consider a controlled Markov process $(\mathcal S,\mathcal A,\mathcal P,\gamma)$ with discount $\gamma\in(0,1)$, integer chunk length $H\ge1$, and bounded reward $r:\mathcal S\to\R$, abbreviated $r_t:=r(s_{t+1})$ along a trajectory. A \emph{boundary pair} $(s,\va)\in\mathcal S\times\mathcal A^H$ consists of a boundary state $s$ together with the $H$-step action chunk $\va=(a_0,\dots,a_{H-1})$ executed from it, producing within-chunk arrivals $(s_1,\dots,s_H)$ and the next boundary pair $(s_H,\va')$. Where compact pair notation is convenient in \cref{app:density-theory}, we abbreviate $x=(s,\va)$, $x'=(s_H,\va')$, and $Y_j:=s_j$ for $j=1,\dots,H$.

We distinguish three expectations: $\E_{s,\va}[\,\cdot\,]=\E_x[\,\cdot\,]:=\E[\,\cdot\mid s_0=s,\ \mathrm{do}(\va_0=\va)]$ averages over the within-chunk arrivals $(s_1,\dots,s_H)$ generated by committing open loop to $\va$ from $s_0=s$, $\E^\pi_{s,\va}[\,\cdot\,]$ averages over the infinite-horizon trajectory that executes $\va$ from $s_0=s$ and draws subsequent chunks $\va_k\sim\pi(\cdot\mid s_{kH})$ at boundaries $t=kH$ for $k\ge1$, and $\E_{\mathcal D}[\,\cdot\mid s_0=s,\ \va_0=\va]$ denotes conditional expectation under the offline dataset law. Writing $C_H:=\sum_{j=1}^{H}\gamma^{j-1}\psi(s_j)$ and $C_\infty:=\sum_{t\ge0}\gamma^t\psi(s_{t+1})$ for a feature map $\psi:\mathcal S\to\R^d$, the chunk successor representation and fixed-policy backup in \cref{sec:method} take the forms $F^\beta_\psi(s,\va)=\E^\beta_{s,\va}[C_\infty]$ and $(\T^\pi G)(s,\va)=\E_{s,\va}\big[C_H+\gamma^H\int G(s_H,\va')\,\pi(\mathrm d\va'\mid s_H)\big]$, with per-chunk discount $\gamma^H<1$. Throughout, $\beta$ denotes the data-collecting behavior reference, $\mu$ the candidate proposer, $\pi$ a stationary continuation policy used for comparison, and $\nu$ the closed-loop policy induced by test-time selection or guidance. Chunks are scored at horizon $H_{\mathrm{score}}=H$, while execution applies $H_{\mathrm{exec}}=H$ under whole-chunk control or $H_{\mathrm{exec}}=1$ under rolling control. Throughout our analysis, we impose five standing assumptions.

\begin{enumerate}[label=(S\arabic*),leftmargin=*,nosep]
\item \emph{Markov chunk-boundary process.} For a boundary state $s$, a chunk $\va$, and a stationary continuation $\pi$, executing $\va$ during steps $0,\dots,H-1$ and drawing $\va_k\sim\pi(\cdot\mid s_{kH})$ at boundaries $t=kH$ ($k\ge1$) renders the trajectory after step $kH$ conditionally independent of its past given $(s_{kH},\va_k)$. Consequently, $W_k:=(s_{kH},\va_k)$ with $W_0=(s,\va)$ forms a Markov chain with transition operator $(\mathsf P^\pi G)(s,\va):=\E^\pi_{s,\va}[G(s_H,\va_1)]=\E_{s,\va}\big[\int G(s_H,\va')\,\pi(\mathrm d\va'\mid s_H)\big]$.
\item \emph{Bounded features.} The feature map $\psi:\mathcal S\to\R^d$ is measurable with $\norm{\psi(s)}_2\le B_\psi<\infty$ for all $s\in\mathcal S$.
\item \emph{Measurability.} State, action, and chunk spaces are standard Borel spaces, policies are Markov kernels, and all value and feature maps are measurable. We write $\mathfrak B_d$ for the Banach space of bounded measurable maps $\mathcal S\times\mathcal A^H\to\R^d$ equipped with $\supnorm{G}:=\sup_{s,\va}\norm{G(s,\va)}_2$.
\item \emph{Stationary fixed continuation.} In every application of $\T^\pi$ or $F^\pi_\psi$, the continuation kernel $\pi$ is held fixed as the operand $G$ varies.
\item \emph{Empirical chunk-transition consistency.} Under the offline distribution $\mathcal D$, the conditional law of $(s_1,\dots,s_H,\va')$ given $(s_0,\va)$ coincides with that of the chunked process under $\beta$, with $\va'\sim\beta(\cdot\mid s_H)$ conditionally independent of $(s_0,\va,s_{1:H-1})$ given $s_H$.
\end{enumerate}

Assumptions (S1)--(S4) govern the population chunk operators in \cref{app:prop1,app:ranking-sec}, whereas (S5) enters only when relating sampled regression objectives on recorded tuples to those population operators in \cref{app:loss-not-cert,app:terminal,app:density-theory}. Consistent with \cref{sec:method}, scalar values and critic scores use $(1-\gamma)$-normalization unless noted otherwise: $J^\pi_r(s,\va):=(1-\gamma)Q^\pi_r(s,\va)$ denotes normalized chunk value. With the default successor target, the predictor $T_\phi(\phi(s),\va)$ estimates the normalized successor $\tilde F^\beta_\psi:=(1-\gamma)F^\beta_\psi$, and we write $\hF(s,\va):=T_\phi(\phi(s),\va)/(1-\gamma)$ for its unnormalized counterpart. Under this normalization, the feature-moment score in \cref{sec:method-readout} is $\hJ_\psi(s,\va)=T_\phi(\phi(s),\va)^\top z=(1-\gamma)\hF(s,\va)^\top z$, the categorical density score $\hJ_D(s,\va)$ in \cref{app:density-theory} is normalized by construction, and unnormalized value statements follow upon dividing by $1-\gamma$ without altering candidate rankings or normalized gradient directions. To compare distributions, we write $\TV(p,q):=\sup_A|p(A)-q(A)|$ on probability measures and $\norm{\vartheta}_{\TV}:=\sup_A|\vartheta(A)|=\sup_{0\le h\le1}|\int h\,\mathrm d\vartheta|$ on zero-mass signed measures $\vartheta$, which gives $|\int r\,\mathrm d\vartheta|\le(\sup r-\inf r)\norm{\vartheta}_{\TV}$ for every bounded measurable $r$. Single-state derivations in \cref{app:ranking-sec,app:grad-sec} follow \cref{sec:method-limits} in suppressing fixed $(s,r)$ arguments, and section-local symbols are introduced where needed.

Following the assumptions above, \cref{app:prop1} establishes fixed-behavior successor contraction and connects the population backup to multi-horizon sampled regression. \Cref{app:ranking-sec,app:rolling} derive candidate-ranking preservation and closed-loop performance bounds under whole-chunk and rolling execution. \Cref{app:grad-sec,app:local-residual} characterize support identifiability, subspace gradient guidance, and localized Bellman residuals, while \cref{app:latent-sec,app:density-theory,app:fb} analyze the feature-moment and density readouts and compare against actor-coupled backups.

\subsection{Identifiability conditions for fixed-behavior evaluation}
\label{app:data}
\label{app:interventional}

Assumptions (S1) and (S5) specify structural conditions on the supplied state representation and trajectory recordings rather than holding automatically whenever trajectories are sliced into $H$-step segments. Whereas one-step offline evaluation requires transitions to follow the primitive kernel $\mathcal D(s,a,s')=d_{\mathcal D}(s,a)\mathcal P(s'\mid s,a)$ without outcome-based filtering, chunk-level evaluation requires both interventional consistency within each chunk and conditional independence across consecutive chunks. First, for almost every recorded boundary pair, conditioning on $(s_0,\va)$ in $\mathcal D$ must match open-loop execution of $\va$, which implies
\begin{equation}
\Law_{\mathcal D}\big(C_H,s_H\,\big|\,s_0=s,\ \va_0=\va\big)=\Law\big(C_H,s_H\,\big|\,s_0=s,\ \mathrm{do}(\va_0=\va)\big).
\label{eq:app-interventional}
\end{equation}
On the data support, \cref{eq:app-interventional} holds whenever transitions are deterministic in the supplied sufficient state or whenever the collector commits to whole chunks open loop without post-hoc trajectory selection, whereas under stochastic transitions a reactive per-step collector can correlate later actions in $\va$ with intermediate disturbances and violate \cref{eq:app-interventional}. Second, substituting the next recorded chunk $\va'$ into the bootstrap target in \cref{sec:method-arch} requires $\Law_{\mathcal D}(\va'\mid s_0,\va_0,s_{1:H})=\beta(\cdot\mid s_H)$ almost surely. Across single or mixed collection sources in \cref{sec:method}, this Markov boundary condition holds when the boundary representation $s_H$ incorporates enough recent history to resolve the active behavior mode, motivating the 16-step state-history window in \cref{app:impl}.

\section{Fixed-behavior successor operators and training targets}
\label{app:prop1}

We first relate the chunk-level successor occupancy measure to its feature moments, then prove $\gamma^H$-contraction and residual bounds for fixed-policy evaluation on the chunk-boundary process \cite{sutton1999between}, and finally connect the population backup to multi-horizon sampled regression and alternative targets.

\subsection{The chunk successor measure and its feature moments}
\begin{lemma}[Successor measure and feature moments]\leavevmode
\label{lem:measure-latent}
Let $M^\beta(X\mid s,\va):=\E^\beta_{s,\va}[\sum_{t\ge0}\gamma^t\Ind\{s_{t+1}\in X\}]$ denote the unnormalized chunk-level successor measure, related to the normalized occupancy $m_\beta$ of \cref{sec:method} by $m_\beta=(1-\gamma)M^\beta$. Under \textnormal{(S1)--(S4)}, $F^\beta_\psi(s,\va)=\int\psi(s^+)\,M^\beta(\mathrm ds^+\mid s,\va)$ and $\tilde F^\beta_\psi(s,\va)=\int\psi(s^+)\,m_\beta(\mathrm ds^+\mid s,\va)$. Moreover, on measure-valued kernels $M$ with uniformly bounded total mass $\sup_{s,\va}M(\mathcal S\mid s,\va)<\infty$, feature integration commutes with the measure-level Bellman backup, and $M^\beta$ is a fixed point of that backup.
\end{lemma}
\begin{proof}
For each $(s,\va)$, monotone convergence defines a nonnegative measure $M^\beta(\cdot\mid s,\va)$ of total mass $\sum_{t\ge0}\gamma^t=(1-\gamma)^{-1}$. Every bounded measurable $f:\mathcal S\to\R$ satisfies
\begin{equation}
\int f\,\mathrm dM^\beta(\cdot\mid s,\va)=\E^\beta_{s,\va}\Big[\textstyle\sum_{t\ge0}\gamma^tf(s_{t+1})\Big],
\label{eq:app-integral}
\end{equation}
which holds for indicator functions by definition, extends to simple functions by linearity, and holds for bounded measurable $f$ by uniform simple-function approximation and dominated convergence. Applying \cref{eq:app-integral} coordinatewise to $\psi$ yields the moment identities for $F^\beta_\psi$ and $\tilde F^\beta_\psi$. Next, define the measure-valued backup
\[
(\T^\beta_MM)(X\mid s,\va):=\E_{s,\va}\Big[\textstyle\sum_{j=1}^{H}\gamma^{j-1}\Ind\{s_j\in X\}\Big]+\gamma^H\,\E_{s,\va}\Big[\int M(X\mid s_H,\va')\,\beta(\mathrm d\va'\mid s_H)\Big].
\]
Integrating $\psi$ against the within-chunk sum gives $\E_{s,\va}[C_H]$ by the argument of \cref{eq:app-integral}, while Fubini's theorem, justified by boundedness of $\psi$ and uniform mass boundedness of $M$, gives $\gamma^H\E_{s,\va}[\int\int\psi\,\mathrm dM(\cdot\mid s_H,\va')\,\beta(\mathrm d\va'\mid s_H)]$ for the continuation term, whence $\int\psi\,\mathrm d(\T^\beta_MM)=\T^\beta(\int\psi\,\mathrm dM)$. Splitting the discounted indicator sum at step $H$ and conditioning on $(s_H,\va_1)$ via (S1) and the tower property yields $M^\beta=\T^\beta_MM^\beta$.
\end{proof}

Extending classical successor representations \cite{dayan1993improving,barreto2017successor} to committed $H$-step chunks, the normalized occupancy $m_\beta=(1-\gamma)M^\beta$ has unit mass. For any bounded arrival reward $r$, the common evaluation functional is $J^\beta_r(s,\va)=\int r\,\mathrm dm_\beta(\cdot\mid s,\va)$, which admits multiple downstream task readouts. In this work we instantiate two estimators of this integral: the feature-moment readout $\hJ_\psi$ in \cref{sec:method-readout,app:latent-sec} and the categorical density readout $\hJ_D$ in \cref{app:density-theory}.

\subsection{Chunk-level fixed-behavior evaluation}
\label{app:proof-prop1}
\label{app:proof-contraction}
\label{app:fixed-contraction}
\label{app:chunk}
\label{app:bellman}

Extending fixed-behavior one-step successor evaluation \cite{dayan1993improving,barreto2017successor,zheng2026can} to $H$-step boundary transitions yields a contraction modulus $\gamma^H$ per backup step and a $1/(1-\gamma^H)$ Bellman-residual bound.

\begin{proposition}[Chunk-level contraction]\leavevmode
\label{prop:contraction}
Under \textnormal{(S1)--(S4)} with fixed bounded measurable features $\psi$ and stationary continuation $\beta$ in \cref{app:notation}: (a)~$\T^\beta$ maps $\mathfrak B_d$ into itself and is a $\gamma^H$-contraction in $\supnorm{\cdot}$ with unique bounded fixed point $F^\beta_\psi$, and (b)~every $\hF\in\mathfrak B_d$ satisfies $\supnorm{\hF-F^\beta_\psi}\le\supnorm{\hF-\T^\beta\hF}/(1-\gamma^H)$.
\end{proposition}
\begin{proof}
Under (S2), $\sum_{t\ge0}\gamma^t\norm{\psi(s_{t+1})}_2\le B_\psi/(1-\gamma)$, so $C_\infty$ converges absolutely and splits as $C_\infty=C_H+\gamma^H C'_\infty$ with $C'_\infty:=\sum_{t\ge0}\gamma^t\psi(s_{H+t+1})$. Conditioning on $\mathcal G_H:=\sigma(s_0,\va_0,s_1,\dots,s_H,\va_1)$ with $\va_1\sim\beta(\cdot\mid s_H)$, (S1) gives $\E^\beta_{s,\va}[C'_\infty\mid\mathcal G_H]=F^\beta_\psi(s_H,\va_1)$, and the tower property yields
\[
F^\beta_\psi(s,\va)
=\E_{s,\va}[C_H]+\gamma^H(\mathsf P^\beta F^\beta_\psi)(s,\va)
=(\T^\beta F^\beta_\psi)(s,\va),
\]
with $F^\beta_\psi\in\mathfrak B_d$ by (S3) and $\supnorm{F^\beta_\psi}\le B_\psi/(1-\gamma)$. For any $G\in\mathfrak B_d$, $\supnorm{\T^\beta G}\le B_\psi(1-\gamma^H)/(1-\gamma)+\gamma^H\supnorm G$ and Markov-kernel integration preserves measurability, so $\T^\beta(\mathfrak B_d)\subset\mathfrak B_d$. For any $G,G'\in\mathfrak B_d$, (S4) fixes the same continuation $\beta$ in both evaluations, so within-chunk terms cancel and Jensen's inequality gives
\[
\norm{(\T^\beta G-\T^\beta G')(s,\va)}_2
=\gamma^H\norm{(\mathsf P^\beta(G-G'))(s,\va)}_2
\le\gamma^H(\mathsf P^\beta\norm{G-G'}_2)(s,\va)
\le\gamma^H\supnorm{G-G'}.
\]
Taking the supremum over $(s,\va)$ proves the $\gamma^H$-contraction, and Banach's fixed-point theorem on the complete space $\mathfrak B_d$ yields uniqueness of $F^\beta_\psi$, establishing~(a). For~(b), the triangle inequality and $F^\beta_\psi=\T^\beta F^\beta_\psi$ give
\[
\supnorm{\hF-F^\beta_\psi}
\le\supnorm{\hF-\T^\beta\hF}+\supnorm{\T^\beta\hF-\T^\beta F^\beta_\psi}
\le\supnorm{\hF-\T^\beta\hF}+\gamma^H\supnorm{\hF-F^\beta_\psi},
\]
and rearranging this finite inequality yields $\supnorm{\hF-F^\beta_\psi}\le\supnorm{\hF-\T^\beta\hF}/(1-\gamma^H)$.
\end{proof}

\label{app:not-qstar}
Rescaling the within-chunk sum to $(1-\gamma)C_H$ defines the normalized backup $\tilde\T^\beta$, which has unique fixed point $\tilde F^\beta_\psi=(1-\gamma)F^\beta_\psi=\int\psi\,\mathrm dm_\beta$ and identical contraction modulus $\gamma^H$. For any task vector $z\in\R^d$ and linear reward $r_z=\psi^\top z$, boundedness of $\psi$ permits exchanging expectation and summation as in \cref{sec:method}:
\begin{equation}
F^\beta_\psi(s,\va)^\top z
=\E^\beta_{s,\va}\sum_{t\ge0}\gamma^t\psi(s_{t+1})^\top z
=Q^\beta_{r_z}(s,\va),
\label{eq:app-readout}
\end{equation}
with normalized counterpart $\tilde F^\beta_\psi(s,\va)^\top z=J^\beta_{r_z}(s,\va)$. Notably, $F^\beta_\psi$ evaluates the fixed continuation $\beta$ rather than an optimal value function $Q^\star$, and open-loop chunk execution agrees with a closed-loop collector under the consistency conditions in \cref{app:interventional}.

\subsection{Population backup and sampled regression}
\label{app:loss-not-cert}

Fix a target predictor $G_k\in\mathfrak B_d$ and feature map $\psi$, and set $\Upsilon_k:=C_H+\gamma^H G_k(s_H,\va')$ for a recorded tuple $(s_0,\va,s_{1:H},\va')$.

\begin{lemma}\leavevmode
\label{lem:loss-decomp}
Let $G_k\in\mathfrak B_d$, so that $\Upsilon_k$ is square integrable. Under \textnormal{(S5)}, every $F\in\mathfrak B_d$ satisfies
\begin{equation}
\E_{\mathcal D}\norm{F(s_0,\va)-\Upsilon_k}_2^2=\E_{\mathcal D}\norm{F(s_0,\va)-(\T^\beta G_k)(s_0,\va)}_2^2+\E_{\mathcal D}\norm{\Upsilon_k-\E_{\mathcal D}[\Upsilon_k\mid s_0,\va]}_2^2 .
\label{eq:app-loss-decomp}
\end{equation}
\end{lemma}
\begin{proof}
Under (S5), $\E_{\mathcal D}[\Upsilon_k\mid s_0,\va]=(\T^\beta G_k)(s_0,\va)$. Writing $F-\Upsilon_k=(F-\T^\beta G_k)+(\T^\beta G_k-\Upsilon_k)$ at $(s_0,\va)$, the first term is $(s_0,\va)$-measurable and both are square integrable, so the cross expectation vanishes.
\end{proof}

Because the conditional-variance term does not depend on $F$, if the fitted class contains $\T^\beta G_k$ and population regression is solved exactly, the minimizer equals $\T^\beta G_k$ almost everywhere under $\mathcal D$ while leaving zero-probability inputs unconstrained. Whenever $\Upsilon_k$ has positive conditional variance on a set of positive data probability, the minimal population squared loss remains strictly positive, and when target networks and encoders evolve jointly, this variance term also depends on the representation parameters. Consequently, a small finite-sample regression loss alone does not bound the uniform residual in \cref{prop:contraction}(b).

\subsection{Multi-horizon successor family}
\label{app:multi-h}

Following \cref{sec:method-arch}, the predictor is conditioned on $H\in\mathcal H=\{4,8,16,32\}$ with base sampling probabilities $(\tfrac16,\tfrac16,\tfrac16,\tfrac12)$, renormalized over horizons fitting before trajectory end, and bootstrapped at a length-matched recorded chunk $\va'$ of length $H$. Shared parameters therefore fit an $H$-indexed family of fixed-horizon successors queried at $H=8$.

\begin{corollary}[$H$-conditioned family of fixed-continuation operators]\leavevmode
\label{cor:multi-h}
For each $H\in\mathcal H$, assume \textnormal{(S1)--(S4)} for the chunked process of length $H$ under continuation $\beta_H$, and let $\tilde\T_H$ denote its normalized backup. On $\prod_{H\in\mathcal H}\mathfrak B_d^{(H)}$ equipped with $\norm{G}:=\max_{H\in\mathcal H}\supnorm{G_H}$, the block operator $(\tilde\T G)_H:=\tilde\T_HG_H$ is a contraction of modulus $\max_{H\in\mathcal H}\gamma^H=\gamma^4$ with unique fixed point $(\tilde F^{\beta_H}_\psi)_{H\in\mathcal H}$, and the $H=8$ operator has contraction modulus $\gamma^8$.
\end{corollary}
\begin{proof}
For each $H\in\mathcal H$, the contraction proof in \cref{app:proof-contraction} gives $\supnorm{\tilde\T_HG_H-\tilde\T_HG'_H}\le\gamma^H\supnorm{G_H-G'_H}$ with unique fixed point $\tilde F^{\beta_H}_\psi$. Because $\tilde\T$ acts blockwise without cross-horizon coupling at the operator level, taking the maximum over $H\in\mathcal H$ yields modulus $\max_{H\in\mathcal H}\gamma^H=\gamma^4$.
\end{proof}

Horizon sampling weights scale the empirical regression loss without replacing these operator moduli by an averaged discount, while parameter sharing couples function approximation across $H\in\mathcal H$.

\subsection{Normalized regression target, trajectory truncation and terminal states}
\label{app:terminal}

For a recorded tuple $(s_0,\va,s_{1:H},\va')$ with normalized arrival mean $\overline\psi_H:=\sum_{j=1}^{H}\gamma^{j-1}\overline\psi(s_j)/\sum_{j=1}^{H}\gamma^{j-1}$ as in \cref{sec:method-arch}, the predictor regresses
\begin{equation}
y=(1-\gamma^H)\overline\psi_H+\gamma^H\,\overline T_\phi\big(\overline\phi(s_H),\va'\big)
=(1-\gamma)\textstyle\sum_{j=1}^{H}\gamma^{j-1}\,\overline\psi(s_j)+\gamma^H\,\overline T_\phi\big(\overline\phi(s_H),\va'\big),
\label{eq:app-td-target}
\end{equation}
which samples $\tilde\T^\beta$ applied to the EMA target networks under (S5), while the symmetric branch exchanges $\phi$ and $\psi$. Because ExORL and OGBench recordings terminate at a time limit rather than an absorbing state, the underlying process continues beyond the cut and no terminal indicator is applied. An index serves as a regression anchor only when at least $2H$ steps follow it so that both $\va$ and $\va'$ are complete, whereas a bootstrap pair $(s_H,\va')$ within $2H$ steps of a cut is evaluated by $\overline T_\phi$ without serving as an anchor there. At the population level, controlling value error on a covered subset $\Omega$ requires controlling exits from $\Omega$ under $\mathsf P^\beta$, as formalized by \cref{lem:exit} in \cref{app:local-residual}. Conversely, in environments with genuine absorbing terminal states $s_T$ whose sufficient state, including any history window, remains invariant and yields constant reward thereafter, absorbing continuation satisfies (S1)--(S3) with $F^\beta_\psi(s_T,\va)=\psi(s_T)/(1-\gamma)$.

\label{app:within-chunk}
The within-chunk target variant in \cref{sec:method-readout} replaces \cref{eq:app-td-target} by $y=\overline\psi_H$, omitting the bootstrap branch while keeping $\gamma$ unchanged. When the conditional arrival mean lies in the predictor class for fixed features and $r_z=\psi^\top z$, its population-optimal linear readout is the normalized finite-horizon return
\[
T_\phi(\phi(s),\va)^\top z
=\frac{\E_{s,\va}\sum_{j=1}^{H}\gamma^{j-1} r_z(s_j)}{\sum_{j=1}^{H}\gamma^{j-1}},
\]
included as a linear variant in \cref{app:zs-pixels}. On state OGBench, the density head trained on frozen within-chunk features estimates the infinite-horizon occupancy $m_\beta$ through its own $\gamma^H$ distributional backup in \cref{app:density-theory}, and the same reward-free head training applies equally to the default successor backbone prior to task readout.

\section{Ranking preservation and closed-loop control}
\label{app:ranking-sec}

Following \cref{sec:method-limits,sec:method-improvement}, we first bound pairwise ranking error across candidates under a target continuation \cite{singh1994upper}, and then accumulate per-boundary selection gains into closed-loop return via the performance-difference identity \cite{kakade2002approximately}.

\subsection{Proof of Proposition~\ref{prop:ranking}}
\label{app:proof-ranking}

\begin{proof}
Fix a boundary state $s$, a bounded reward $r$, a stationary continuation $\pi$ satisfying (S1) and (S3)--(S4), and a nonempty finite candidate set $\mathcal A_s$. Suppressing fixed $(s,r)$ as in \cref{sec:method-limits}, write $J^\pi(\va):=(1-\gamma)Q^\pi_r(s,\va)$, $J^\beta(\va):=(1-\gamma)Q^\beta_r(s,\va)$, $e(\va):=\hJ(\va)-J^\beta(\va)$, $\Delta V:=V^\beta_r-V^\pi_r$, and $u(\va):=\E_{s,\va}\Delta V(s_H)=\int\Delta V\,\mathrm dP_{s,\va}$, where $P_{s,\va}$ denotes the law of $s_H$ under $(s,\va)$. Because $Q^\beta_r(s,\va)$ and $Q^\pi_r(s,\va)$ share the within-chunk return $R_H:=\sum_{t=0}^{H-1}\gamma^t r_t$, conditioning on $s_H$ cancels $R_H$ and gives
\begin{equation}
J^\beta(\va)-J^\pi(\va)=(1-\gamma)\gamma^H u(\va)=(1-\gamma)\gamma^H\int\Delta V\,\mathrm dP_{s,\va}.
\label{eq:app-continuation}
\end{equation}
Setting $\hat e_\pi:=\hJ-J^\pi$, adding $e=\hJ-J^\beta$ to \cref{eq:app-continuation} recovers \cref{eq:reuse-split}, and subtracting \cref{eq:reuse-split} across two candidates $\va,\vb\in\mathcal A_s$ yields
\begin{equation}
\hat e_\pi(\va)-\hat e_\pi(\vb)
=e(\va)-e(\vb)+(1-\gamma)\gamma^H\big[u(\va)-u(\vb)\big]
=e(\va)-e(\vb)+(1-\gamma)\gamma^H\int\Delta V\,\mathrm d(P_{s,\va}-P_{s,\vb}).
\label{eq:app-gap-split}
\end{equation}
For any real-valued map $f$ on a nonempty finite set, $\max_{a,b}|f(a)-f(b)|=\max f-\min f$ with attained extrema, so the continuation spread in \cref{eq:gap-terms} equals $\delta_\pi(\mathcal A_s)=(1-\gamma)\gamma^H\max_{\va,\vb\in\mathcal A_s}|u(\va)-u(\vb)|<\infty$. On $\{\va,\vb\}$, $\eta(\{\va,\vb\})=|e(\va)-e(\vb)|$ and $\delta_\pi(\{\va,\vb\})=(1-\gamma)\gamma^H|u(\va)-u(\vb)|$, so the triangle inequality and monotonicity of set ranges give
\begin{equation}
|\hat e_\pi(\va)-\hat e_\pi(\vb)|
\le\varepsilon_\Delta(\{\va,\vb\})
\le\varepsilon_\Delta(\mathcal A_s),\qquad
\varepsilon_\Delta(B):=\eta(B)+\delta_\pi(B).
\label{eq:app-pair-bound}
\end{equation}
Since $[\hJ(\va)-\hJ(\vb)]-[J^\pi(\va)-J^\pi(\vb)]=\hat e_\pi(\va)-\hat e_\pi(\vb)$, \cref{eq:app-pair-bound} proves the pairwise bound, and whenever either gap has magnitude strictly greater than $\varepsilon_\Delta(\mathcal A_s)$, both scores share the same strict ordering.
\end{proof}

\begin{corollary}[Selection loss]\leavevmode
\label{cor:selection-loss}
Under the assumptions of \cref{prop:ranking}, if $\hat\va\in\mathcal A_s$ satisfies $\hJ(\hat\va)\ge\max_{\mathcal A_s}\hJ-\xi$ with $\xi\ge0$, then
\[
\max_{\mathcal A_s}J^\pi-J^\pi(\hat\va)\le\varepsilon_\Delta(\mathcal A_s)+\xi.
\]
For any included reference chunk $\vb\in\mathcal A_s$, writing $G=\max_{\mathcal A_s}J^\pi-J^\pi(\vb)$ gives $J^\pi(\hat\va)-J^\pi(\vb)\ge G-\varepsilon_\Delta(\mathcal A_s)-\xi$.
\end{corollary}
\begin{proof}
Choose $\va^\star\in\argmax_{\mathcal A_s}J^\pi$, so approximate maximization gives $\hJ(\va^\star)-\hJ(\hat\va)\le\xi$. Applying \cref{eq:app-pair-bound} yields
\[
0\le J^\pi(\va^\star)-J^\pi(\hat\va)
=\hJ(\va^\star)-\hJ(\hat\va)+\hat e_\pi(\hat\va)-\hat e_\pi(\va^\star)
\le\xi+|\hat e_\pi(\hat\va)-\hat e_\pi(\va^\star)|
\le\xi+\varepsilon_\Delta(\{\va^\star,\hat\va\})
\le\xi+\varepsilon_\Delta(\mathcal A_s),
\]
which proves the selection-loss bound, and subtracting this loss from $G$ recovers \cref{eq:selection-gain} at $\pi=\mu$ and $\xi=0$. The bound holds pathwise for every realized candidate set, including score-dependent proposals.
\end{proof}

\subsection{Whole-chunk performance difference}
\label{app:closed-sec}
\label{app:closed-loop}

We next extend single-state selection in \cref{cor:selection-loss} to whole-chunk closed-loop execution in \cref{sec:method-improvement}. Set $\pi=\mu$, write $J^\mu(s,\va):=(1-\gamma)Q^\mu_r(s,\va)$ and $\hat e_\mu(s,\va):=\hJ(s,\va)-J^\mu(s,\va)$, and condition throughout on any episode-shared randomness such as the cached reference set. At each boundary state $s$, the procedure draws a nonempty finite candidate set $\mathcal A_s$, whose elements may depend on one another and on $\hJ$, using fresh randomness independent of the boundary history given $s$, and selects $\hat\va\in\mathcal A_s$ satisfying $\hJ(s,\hat\va)\ge\max_{\mathcal A_s}\hJ(s,\cdot)-\xi$ ($\xi\ge0$) via a fixed measurable tie-breaking rule, executing full $H$-step chunks under both $\nu$ and $\mu$.

On each realized draw $\mathcal A_s$, define the selection loss $L$ and, whenever a reference chunk $\vb$ is drawn at $s$, the realized maximum candidate gain $G$ as in \cref{sec:method-improvement}:
\begin{equation}
G:=\max_{\va\in\mathcal A_s}J^\mu(s,\va)-J^\mu(s,\vb),\qquad
L:=\max_{\va\in\mathcal A_s}J^\mu(s,\va)-J^\mu(s,\hat\va)\ge0,
\label{eq:app-headroom}
\end{equation}
so $J^\mu(s,\hat\va)-J^\mu(s,\vb)=G-L$ identically, and $\vb\in\mathcal A_s$ guarantees $G\ge0$ pathwise even when candidates are correlated. Defining the realized candidate gap error
\begin{equation}
\hat\varepsilon_\Delta(s,\mathcal A_s):=\max_{\va,\va'\in\mathcal A_s}\big|[\hJ(s,\va)-\hJ(s,\va')]-[J^\mu(s,\va)-J^\mu(s,\va')]\big|=\max_{\mathcal A_s}\hat e_\mu(s,\cdot)-\min_{\mathcal A_s}\hat e_\mu(s,\cdot),
\label{eq:app-gap-error}
\end{equation}
\cref{eq:app-pair-bound} at $\pi=\mu$ and the proof of \cref{cor:selection-loss} give $L\le\hat\varepsilon_\Delta(s,\mathcal A_s)+\xi\le\varepsilon_\Delta(\mathcal A_s)+\xi$ on every realization.

Because candidate generation and tie-breaking use fresh randomness at each boundary under a stationary rule, the conditional law of $\hat\va$ given the boundary history depends only on $s$, inducing a stationary chunk policy $\nu(\mathrm d\va\mid s)$. With one-boundary advantage $A^\mu_\nu(s):=\int Q^\mu_r(s,\va)\,\nu(\mathrm d\va\mid s)-V^\mu_r(s)$ and $\bar G(s):=\E[\max_{\mathcal A_s}J^\mu(s,\cdot)\mid s]-(1-\gamma)V^\mu_r(s)$, we always have $(1-\gamma)A^\mu_\nu(s)=\bar G(s)-\E[L\mid s]$. Whenever a reference draw satisfies $\vb\mid s\sim\mu(\cdot\mid s)$, the chunk Bellman equation for $\mu$ gives $\E[J^\mu(s,\vb)\mid s]=(1-\gamma)V^\mu_r(s)$, so $\bar G(s)=\E[G\mid s]$ and $(1-\gamma)A^\mu_\nu(s)=\E[G-L\mid s]$. Writing $\mathsf P^\nu_{\mathcal S}$ for the boundary-state transition kernel of $\nu$ and $\bar r^\nu$ for its expected one-chunk discounted reward, the chunk Bellman equations give
\[
V^\nu_r-V^\mu_r
=\bar r^\nu+\gamma^H\mathsf P^\nu_{\mathcal S}V^\nu_r-V^\mu_r
=\underbrace{\bar r^\nu+\gamma^H\mathsf P^\nu_{\mathcal S}V^\mu_r-V^\mu_r}_{A^\mu_\nu}
 +\gamma^H\mathsf P^\nu_{\mathcal S}(V^\nu_r-V^\mu_r).
\]
Iterating $n$ times leaves a remainder bounded by $\gamma^{Hn}\supnorm{V^\nu_r-V^\mu_r}\to0$ as $n\to\infty$, yielding the chunk-level performance-difference identity \cite{kakade2002approximately}:
\begin{equation}
V^\nu_r-V^\mu_r
=\sum_{k\ge0}\gamma^{Hk}(\mathsf P^\nu_{\mathcal S})^kA^\mu_\nu,
\label{eq:app-pdl}
\end{equation}
or $(V^\nu_r-V^\mu_r)(s_0)=(1-\gamma^H)^{-1}\E_{s\sim d^\nu_{s_0}}[A^\mu_\nu(s)]$ under the normalized boundary occupancy $d^\nu_{s_0}:=(1-\gamma^H)\sum_{k\ge0}\gamma^{Hk}(\mathsf P^\nu_{\mathcal S})^k(s_0,\cdot)$.

\subsection{Maximum candidate gain and selection loss}
\label{app:headroom}

Combining the derivations in \cref{app:closed-loop} yields the closed-loop identity of \cref{sec:method-improvement} and its corresponding lower bounds.

\begin{proposition}[Maximum gain minus selection loss]\leavevmode
\label{prop:headroom}
In the setting of \cref{app:closed-loop}, let $\bar\varepsilon_\Delta(s):=\E[\hat\varepsilon_\Delta(s,\mathcal A_s)\mid s]\le\E[\eta(\mathcal A_s)+\delta_\mu(\mathcal A_s)\mid s]$.
(a)~Every realization satisfies $L\le\hat\varepsilon_\Delta(s,\mathcal A_s)+\xi$, so $\E[L\mid s]\le\bar\varepsilon_\Delta(s)+\xi$.
(b)~If at each boundary the candidate set $\mathcal A_s$ contains a fresh reference draw $\vb\in\mathcal A_s$ with conditional law $\mu(\cdot\mid s)$ given $s$ and the boundary history, then $G\ge0$ pathwise, $(1-\gamma)A^\mu_\nu(s)=\E[G-L\mid s]$, and
\begin{equation}
(1-\gamma)\big(V^\nu_r-V^\mu_r\big)(s_0)=\frac{\E_{s\sim d^\nu_{s_0}}\E\big[G-L\mid s\big]}{1-\gamma^H}\ge\frac{\E_{s\sim d^\nu_{s_0}}\big[\E[G\mid s]-\bar\varepsilon_\Delta(s)\big]-\xi}{1-\gamma^H}.
\label{eq:app-headroom-pdl}
\end{equation}
Without a reference draw from $\mu$, replacing $\E[G-L\mid s]$ by $\bar G(s)-\E[L\mid s]$ and $\E[G\mid s]$ by $\bar G(s)$ as defined in \cref{app:closed-loop} preserves both the equality and the lower bound in \cref{eq:app-headroom-pdl}, though $\bar G(s)$ need not be nonnegative.
\end{proposition}
\begin{proof}
Part~(a) averages the pathwise selection-loss bound of \cref{app:closed-loop} over the candidate draw at $s$, and part~(b) substitutes $(1-\gamma)A^\mu_\nu(s)=\E[G-L\mid s]$ into the occupancy form of \cref{eq:app-pdl} multiplied by $1-\gamma$, where $\vb\in\mathcal A_s$ ensures $G\ge0$ and applying~(a) yields the lower bound.
\end{proof}

Dropping the nonnegative gain $\E[G\mid s]\ge0$ in \cref{eq:app-headroom-pdl} yields a worst-case floor under bounded mean spreads.

\begin{corollary}[Closed-loop comparison]\leavevmode
\label{cor:closed-loop}
Under the assumptions of \cref{prop:headroom}(b), if at every boundary state reachable by $\nu$ we have $\E[\eta(\mathcal A_s)\mid s]\le\bar\eta$ and $\E[\delta_\mu(\mathcal A_s)\mid s]\le\bar\delta_\mu$, then
\[
(1-\gamma)(V^\nu_r-V^\mu_r)\ge-\frac{\bar\eta+\bar\delta_\mu+\xi}{1-\gamma^H}.
\]
\end{corollary}
\begin{proof}
At every reachable state, $\E[G\mid s]\ge0$ and $\bar\varepsilon_\Delta(s)\le\bar\eta+\bar\delta_\mu$, so \cref{eq:app-headroom-pdl} yields the stated floor.
\end{proof}

The right-hand side of \cref{eq:app-headroom-pdl} is strictly positive whenever $\E[G\mid s]$ exceeds $\bar\varepsilon_\Delta(s)+\xi$ in expectation under $d^\nu_{s_0}$. Four structural properties follow directly. First, any state-only offset $\hat e_\mu(s,\va)=c(s)$ cancels from $\hat\varepsilon_\Delta(s,\mathcal A_s)$, leaving both selection and the bound invariant. Second, continuation mismatch enters solely through endpoint differences across candidates via $\delta_\mu(\mathcal A_s)$ in \cref{eq:app-gap-split}, with total-variation and Lipschitz forms given in \cref{rem:eps-q-compare}. Third, $G$ reflects proposal quality under $\mu$, depending on $\hJ$ only when candidates are gradient-guided, whereas $L$ captures ranking loss on realized proposals, with both $\max_{\mathcal A_s}J^\mu(s,\cdot)$ and $\hat\varepsilon_\Delta(s,\mathcal A_s)$ nondecreasing across nested sets. Fourth, both $G$ and $L$ are averaged under the selected policy's boundary occupancy $d^\nu_{s_0}$ in \cref{eq:app-headroom-pdl}. In deployment, BFN16 draws all candidates from $\mu$ and QGF+BFN4 augments four guided draws with one unguided draw from $\mu$, so both satisfy $\vb\in\mathcal A_s$ and $\vb\mid s\sim\mu(\cdot\mid s)$ in \cref{prop:headroom}(b), whereas QGF alone and guided-only QGF+BFN4$^\circ$ omit the unguided draw and obey the general $\bar G(s)$ form.

\subsection{Necessity of the continuation spread}
\begin{example}[No continuation-free score equals every $Q^\pi$]\leavevmode
\label{ex:two-branch}
Let $\gamma=0.9$ and $H=1$. From $s$, chunk $A$ transitions to $s_A$ and chunk $B$ to $s_B$, both with arrival reward $0$. From both $s_A$ and $s_B$, one action enters an absorbing state of per-step reward $1$ and another enters an absorbing state of reward $0$. Policy $\pi_1$ chooses the rewarding action at $s_A$ and the zero-reward action at $s_B$, whereas $\pi_2$ reverses both choices. Because $(Q^{\pi_1}_r(s,A),Q^{\pi_1}_r(s,B))=(9,0)$ and $(Q^{\pi_2}_r(s,A),Q^{\pi_2}_r(s,B))=(0,9)$, no policy-agnostic score can rank $A$ and $B$ identically to both continuations. An exact behavior critic for $\beta=\pi_1$ has $e\equiv0$ and $\eta=0$, while against $\pi=\pi_2$ we have $\Delta V(s_A)=10$, $\Delta V(s_B)=-10$, and $\delta_{\pi_2}(\{A,B\})=(1-\gamma)\gamma\cdot20=1.8$, matching the exact error split $\hat e_{\pi_2}(A)-\hat e_{\pi_2}(B)=0.9-(-0.9)=1.8=\delta_{\pi_2}$ in \cref{eq:app-gap-split} and bounding the realized normalized regret $0.9$ via \cref{cor:selection-loss}.
\end{example}

\subsection{Score-error budgets for feature-moment and density readouts}
\label{app:score-budgets}

The ranking bounds in \cref{app:proof-ranking} depend only on scalar scores. For the feature-moment readout with $r_z=\psi^\top z$, \cref{eq:app-readout} yields
\[
e(\va)=(1-\gamma)(\hF-F^\beta_\psi)(s,\va)^\top z
       +(1-\gamma)(Q^\beta_{r_z}-Q^\beta_r)(s,\va).
\]
If $(1-\gamma)\norm{\hF-F^\beta_\psi}_2\le\varepsilon_F(\va)$ and $|r-r_z|\le\varepsilon_r$ on all relevant arrivals, Cauchy--Schwarz and the discounted sum give $|e(\va)|\le\varepsilon_\beta(\va):=\norm z_2\varepsilon_F(\va)+\varepsilon_r$. For the density readout, \cref{cor:density-value} in \cref{app:density-theory} supplies the corresponding scalar behavior-score bound $|e(\va)|\le\varepsilon_\beta(\va)$ from occupancy, reference-sampling, and reward-proxy errors.

If $|\Delta V|\le D_V$ across reachable post-chunk states, \cref{eq:app-continuation} further gives
\begin{equation}
|\hJ(\va)-J^\pi(\va)|\le\varepsilon_Q(\va)
:=\varepsilon_\beta(\va)+(1-\gamma)\gamma^H D_V,
\qquad
\varepsilon_\Delta(\mathcal A_s)\le2\max_{\mathcal A_s}\varepsilon_Q,
\label{eq:eps-q}
\end{equation}
bounding the error spread by $2\max_{\mathcal A_s}\varepsilon_\beta$ and the continuation spread by $2(1-\gamma)\gamma^H D_V$, with pairwise refinement $\varepsilon_\Delta(\{\va,\vb\})\le\varepsilon_Q(\va)+\varepsilon_Q(\vb)$. At $\pi=\mu$, every realization satisfies $\hat\varepsilon_\Delta(s,\mathcal A_s)\le\varepsilon_\Delta(\mathcal A_s)\le2\max_{\mathcal A_s}\varepsilon_Q$, so retaining the same gain term $\E[G\mid s]$ under $d^\nu_{s_0}$ makes the spread-based lower bound in \cref{eq:app-headroom-pdl} strictly tighter than the uniform-error bound $\big(\E_{s\sim d^\nu_{s_0}}\big[\E[G\mid s]-2\E[\max_{\mathcal A_s}\varepsilon_Q\mid s]\big]-\xi\big)/(1-\gamma^H)$ whenever $\E_{s\sim d^\nu_{s_0}}[\bar\varepsilon_\Delta(s)]<2\,\E_{s\sim d^\nu_{s_0}}\E[\max_{\mathcal A_s}\varepsilon_Q\mid s]$.

\begin{remark}[Relative versus uniform error bounds]\leavevmode
\label{rem:eps-q-compare}
If $e(\va)=c(s)+\tilde e(\va)$, the state-only offset cancels from $\eta$. Likewise, a continuation gap constant on the union $E_s:=\bigcup_{\va\in\mathcal A_s}\operatorname{supp}(P_{s,\va})$ of candidate endpoint supports contributes zero to $\delta_\pi$. More generally,
\[
\delta_\pi(\mathcal A_s)\le(1-\gamma)\gamma^H\,
\big(\sup_{E_s}\Delta V-\inf_{E_s}\Delta V\big)\max_{\va,\vb\in\mathcal A_s}\TV(P_{s,\va},P_{s,\vb}).
\]
Under deterministic dynamics it equals $(1-\gamma)\gamma^H$ times $\max_{E_s}\Delta V-\min_{E_s}\Delta V$ over the reached endpoints $E_s$. If $\Delta V$ is $L_V$-Lipschitz on $E_s$, this is at most $(1-\gamma)\gamma^H L_V$ times the endpoint diameter. These are conditions under which the candidate comparison is sharper than an absolute-error bound.
\end{remark}

\section{Rolling first-action execution and tail correction}
\label{app:rolling}

Whereas \cref{cor:closed-loop} analyzes open-loop execution of whole chunks, rolling execution applies only the first action and replans at every step. Below we relate committed-chunk and first-action values, derive the closed-loop rolling bound of \cref{sec:method-improvement}, demonstrate why the tail term is indispensable, and formalize completion-averaged scoring.

\subsection{First-action and committed-chunk values}
\label{app:r7-setup}
\label{app:r7-pdl}

Assume a sufficient state at every primitive step, bounded rewards, and the measurability conditions of (S3). Let $\mu$ be the frozen chunk proposer and $\mu_1$ its first-action marginal, so the unguided rolling base policy has value $V^{\mu_1}_r$ with normalized value $v^{\mu_1}:=(1-\gamma)V^{\mu_1}_r$, and write $v^\nu:=(1-\gamma)V^\nu_r$ for the rolling selector $\nu$. Because \cref{eq:reuse-split} depends on $\pi$ solely through endpoint values $V^\pi_r(s_H)$, it applies directly when the post-chunk continuation is the primitive-step policy $\pi=\mu_1$. For $0\le\ell\le H$, define the $\ell$-step committed value
\[
J^{\mu_1}_\ell(s,\va):=(1-\gamma)\E_{s,\va}\left[
\sum_{j=0}^{\ell-1}\gamma^j r_j+\gamma^\ell V^{\mu_1}_r(s_\ell)\right],
\]
with $J^{\mu_1}_0(s,\va)=v^{\mu_1}(s)$, and define the normalized tail advantage
\begin{equation}
\tau(s,\va):=\E_{s,a_0}\left[
J^{\mu_1}_{H-1}(s_1,\va_{1:H-1})-v^{\mu_1}(s_1)\right].
\label{eq:rolling-tail}
\end{equation}
Splitting $J^{\mu_1}_H(s,\va)=(1-\gamma)\E_{s,a_0}r_0+\gamma\E_{s,a_0}J^{\mu_1}_{H-1}(s_1,\va_{1:H-1})$ and subtracting $J^{\mu_1}_1(s,a_0)=(1-\gamma)\E_{s,a_0}r_0+\gamma\E_{s,a_0}v^{\mu_1}(s_1)$ yields the exact identity
\begin{equation}
J^{\mu_1}_1(s,a_0)=J^{\mu_1}_H(s,\va)-\gamma\tau(s,\va).
\label{eq:rolling-split}
\end{equation}
Comparing committed-chunk values therefore compares executed first actions up to their tail difference $\gamma\tau(s,\va)$.

\subsection{Rolling performance with a relative tail correction}
\label{app:r7-no-reuse}
\label{rem:rolling-spread}

\begin{proposition}[Rolling execution]\leavevmode
\label{prop:r7-rolling}
\label{prop:r7-relative-tail}
At every reachable pre-selection history with current state $s$, let the candidate set $\mathcal A_s$ contain a fresh draw $\vb$ with conditional law $\mu(\cdot\mid s)$, and let $\hat\va\in\mathcal A_s$ satisfy $\hJ(s,\hat\va)\ge\max_{\mathcal A_s}\hJ(s,\cdot)-\xi$ ($\xi\ge0$). Evaluate \cref{eq:gap-terms} at $\pi=\mu_1$ with comparison value $J_H^{\mu_1}$, maximum candidate gain $G:=\max_{\mathcal A_s}J_H^{\mu_1}(s,\cdot)-J_H^{\mu_1}(s,\vb)\ge0$, and $\varepsilon_\Delta(\mathcal A_s):=\eta(\mathcal A_s)+\delta_{\mu_1}(\mathcal A_s)$.
(a)~Every realization satisfies
\[
J_1^{\mu_1}(s,\hat a_0)-J_1^{\mu_1}(s,b_0)
\ge G-\varepsilon_\Delta(\mathcal A_s)-\xi-\gamma\big[\tau(s,\hat\va)-\tau(s,\vb)\big].
\]
(b)~If along every reachable pre-selection history $\E\big[G-\varepsilon_\Delta(\mathcal A_s)-\gamma\big(\tau(s,\hat\va)-\tau(s,\vb)\big)\mid\text{history}\big]\ge\underline\Delta$, then $(1-\gamma)\big(V^\nu_r-V^{\mu_1}_r\big)(s_0)\ge(\underline\Delta-\xi)/(1-\gamma)$. In particular, if the conditional means of $\eta(\mathcal A_s)$, $\delta_{\mu_1}(\mathcal A_s)$, and $\tau(s,\hat\va)-\tau(s,\vb)$ are bounded above by $\bar\eta$, $\bar\delta_{\mu_1}$, and $\delta_{\mathrm{rel}}$, respectively, then
\[
(1-\gamma)\big(V^\nu_r-V^{\mu_1}_r\big)(s_0)\ge-
\frac{\bar\eta+\bar\delta_{\mu_1}+\xi+\gamma\delta_{\mathrm{rel}}}{1-\gamma}.
\]
\end{proposition}
\begin{proof}
For part~(a), $\vb\in\mathcal A_s$ implies $G\ge0$, and \cref{cor:selection-loss} at $\pi=\mu_1$ gives $J_H^{\mu_1}(s,\hat\va)-J_H^{\mu_1}(s,\vb)\ge G-\varepsilon_\Delta(\mathcal A_s)-\xi$ on each realization. Subtracting \cref{eq:rolling-split} between $\hat\va$ and $\vb$ gives
\[
J_1^{\mu_1}(s,\hat a_0)-J_1^{\mu_1}(s,b_0)
=J_H^{\mu_1}(s,\hat\va)-J_H^{\mu_1}(s,\vb)
-\gamma\big[\tau(s,\hat\va)-\tau(s,\vb)\big],
\]
which combines with the chunk inequality to prove~(a). For part~(b), because $b_0$ has conditional law $\mu_1(\cdot\mid s)$, the primitive Bellman equation gives $\E[J_1^{\mu_1}(s,b_0)\mid\text{history}]=v^{\mu_1}(s)$ without assuming equality between committed-chunk and rolling base values. Taking conditional expectations in~(a) yields $\E[J_1^{\mu_1}(s,\hat a_0)\mid\text{history}]-v^{\mu_1}(s)\ge\underline\Delta-\xi$, with $\underline\Delta=-(\bar\eta+\bar\delta_{\mu_1}+\gamma\delta_{\mathrm{rel}})$ recovered upon dropping $G\ge0$. Since $\E[J_1^{\mu_1}(s_t,\hat a_t)\mid\text{history}]-v^{\mu_1}(s_t)=\E[(1-\gamma)r_t+\gamma v^{\mu_1}(s_{t+1})-v^{\mu_1}(s_t)\mid\text{history}]$, multiplying by $\gamma^t$, taking expectations, and telescoping through $n-1$ gives
\[
(1-\gamma)\E^\nu_{s_0}\sum_{t=0}^{n-1}\gamma^t r_t
-v^{\mu_1}(s_0)+\gamma^n\E^\nu_{s_0} v^{\mu_1}(s_n)
\ge(\underline\Delta-\xi)\frac{1-\gamma^n}{1-\gamma}.
\]
Boundedness of $v^{\mu_1}$ sends $\gamma^n\E^\nu_{s_0} v^{\mu_1}(s_n)\to0$ as $n\to\infty$, proving~(b) including under history-dependent selection.
\end{proof}

At $H=1$, $\tau\equiv0$ recovers the standard one-step approximate-greedy floor, whereas under a uniform score error $|\hJ-J_H^{\mu_1}|\le\bar\varepsilon_Q$ across queried candidates one obtains $(1-\gamma)(V^\nu_r-V^{\mu_1}_r)(s_0)\ge-(2\bar\varepsilon_Q+\xi+\gamma\delta_{\mathrm{rel}})/(1-\gamma)$.

\subsection{Necessity of the tail correction}
\label{app:r7-counterexample}
\label{app:r7-pure-bfn}

Even with exact behavior-critic scores and full data support, selecting the highest-value committed chunk can lower closed-loop return under rolling execution when its value comes from an unexecuted tail, motivating the comparison of both execution modes in \cref{sec:exp-swap}.

\begin{example}[On-support BFN under rolling execution]\leavevmode
\label{ex:r7-pure-bfn}
Consider two states $s_0,s_1$ with arrival rewards $r(s_0)=0$ and $r(s_1)=1$, where primitive action $i\in\{0,1\}$ transitions deterministically to $s_i$ from either state. Let $\gamma=0.9$, $H=3$, and at both states set
\[
\beta(\cdot\mid s)=\mu(\cdot\mid s)=\tfrac12\delta_{\va_{\mathrm A}}+\tfrac12\delta_{\va_{\mathrm B}},
\qquad \va_{\mathrm A}=(1,0,0),\quad \va_{\mathrm B}=(0,1,1).
\]
Each step of a fresh chunk visits $s_1$ with probability $1/2$, so both the committed-chunk base $\beta$ and its rolling marginal $\mu_1$ achieve $V^\beta_r(s)=V^{\mu_1}_r(s)=5$ at either state $s\in\{s_0,s_1\}$. Exact committed-chunk values from either state $s$ are
\[
Q^\beta_r(s,\va_{\mathrm A})=1+0.9^3\cdot5=4.645,\qquad
Q^\beta_r(s,\va_{\mathrm B})=0.9+0.9^2+0.9^3\cdot5=5.355,
\]
with opposite normalized tail advantages $\tau(s,\va_{\mathrm A})=-0.095$ and $\tau(s,\va_{\mathrm B})=+0.095$. Drawing $N\ge2$ independent chunks at every step and selecting their exact maximizer executes first action $1$ only when all $N$ draws equal $\va_{\mathrm A}$ (probability $2^{-N}$), yielding rolling value $V^\nu_r(s)=2^{-N}/(1-\gamma)=10\cdot 2^{-N}<5$ ($2.5$ at $N=2$).
\end{example}

\subsection{Completion-averaged first-action scores}
\label{app:completion-readout}

To guide single-action actors in \cref{app:zs-steer-actors}, fix a target primitive policy $\pi$ and a completion kernel $q(\mathrm d\va_{\mathrm{tail}}\mid s,a_0)$ over $(H-1)$-step suffixes $\va_{\mathrm{tail}}$ generated by Flow BC inpainting. Defining $J_H^\pi$, $J_1^\pi$, and $\tau_\pi$ as in \cref{app:r7-setup} with $\pi$ replacing $\mu_1$, set
\[
\bar J_q(s,a_0):=\E_{\va_{\mathrm{tail}}\sim q}J_H^\pi\big(s,(a_0,\va_{\mathrm{tail}})\big),\qquad
\bar{\hJ}_q(s,a_0):=\E_{\va_{\mathrm{tail}}\sim q}\hJ\big(s,(a_0,\va_{\mathrm{tail}})\big).
\]
Because $J_1^\pi(s,a_0)$ does not depend on $\va_{\mathrm{tail}}$, averaging \cref{eq:rolling-split} gives
\begin{equation}
\bar J_q(s,a_0)-J_1^\pi(s,a_0)
=\gamma\E_{\va_{\mathrm{tail}}\sim q}\tau_\pi\big(s,(a_0,\va_{\mathrm{tail}})\big),
\label{eq:completion-tail}
\end{equation}
and adding and subtracting $\bar J_q(s,a_0)$ yields the exact score-error decomposition
\begin{equation}
\bar{\hJ}_q(s,a_0)-J_1^\pi(s,a_0)
=\E_{\va_{\mathrm{tail}}\sim q}\big[\hJ\big(s,(a_0,\va_{\mathrm{tail}})\big)-J_H^\pi\big(s,(a_0,\va_{\mathrm{tail}})\big)\big]
+\gamma\E_{\va_{\mathrm{tail}}\sim q}\tau_\pi\big(s,(a_0,\va_{\mathrm{tail}})\big).
\label{eq:completion-error}
\end{equation}
Here the first expectation combines behavior-critic and continuation errors via \cref{eq:reuse-split}, while the second vanishes whenever the completion law $q(\cdot\mid s,a_0)$ matches the conditional within-chunk reward and endpoint distribution induced by $\pi$ from $s_1$. Averaging $K=4$ completions with shared tail noise across candidate actions $a_0$ correlates Monte Carlo completion variance across candidates, so only cross-candidate differences enter selection.

\section{Offline support, gradient identifiability and subspace guidance}
\label{app:grad-sec}

\subsection{Data support and identifiability}
\label{app:nonid}

Population support is taken in fixed compatible Polish topologies on the state and chunk spaces, with the Euclidean topology on $\mathcal A^H$ in part~(b). Perturbations below act on the unnormalized predictor $\hF=T_\phi/(1-\gamma)$ and its target network.

\begin{proposition}[Support identifiability]\leavevmode
\label{prop:nonid}
Fix the feature coordinates and let $\mathcal S_{\mathcal D}$ be the population support of the pairs at which the TD loss evaluates the online predictor or the target predictor. (a)~If $h$ is bounded, measurable, and zero on $\mathcal S_{\mathcal D}$, adding $h$ to the online and target successor functions leaves every sampled TD term unchanged while shifting the true Bellman residual by $h-\gamma^H\mathsf P^\beta h$. (b)~Suppose chunks lie in a Euclidean space, $\mathcal S_{\mathcal D}$ is contained in the graph of a measurable map $s\mapsto\va_{\mathcal D}(s)$, and the predictor class is closed under adding the bounded measurable chunk-differentiable perturbations constructed below. Then for $z\neq0$ and every vector $v$, some admissible $h$ shifts $\nabla_{\va}\hJ_\psi$ at every demonstrated pair by $v$, and for any $(s_*,\va_*)\notin\mathcal S_{\mathcal D}$ and finite $c\in\R$, another shifts the normalized score at $(s_*,\va_*)$ by $c$ without changing any sampled TD term.
\end{proposition}
\begin{proof}
For~(a), both $(s_0,\va)$ and $(s_H,\va')$ lie in $\mathcal S_{\mathcal D}$ almost surely under $\mathcal D$, so $h=0$ on $\mathcal S_{\mathcal D}$ leaves every sampled loss term unchanged, whereas affinity of $\T^\beta$ shifts the population residual by $h-\gamma^H\mathsf P^\beta h$. For~(b), fix $A\in\R^{d\times\dim\mathcal A^H}$ and a smooth cutoff $\varsigma:\R_{\ge0}\to[0,1]$ equal to $1$ on $[0,1]$ and $0$ on $[2,\infty)$, and set $h(s,\va):=\varsigma(\norm{\va-\va_{\mathcal D}(s)}_2)\,A(\va-\va_{\mathcal D}(s))$. Then $\supnorm{h}\le 2\opnorm A$, $h=0$ on the graph, and $\partial_{\va}h(s,\va_{\mathcal D}(s))=A$, shifting $\nabla_{\va}\hJ_\psi$ at demonstrated pairs by $(1-\gamma)A^\top z=v$ upon setting $A=zv^\top/((1-\gamma)\norm{z}_2^2)$. For $(s_*,\va_*)\notin\mathcal S_{\mathcal D}$, closedness of $\mathcal S_{\mathcal D}$ yields a disjoint product neighborhood $U\times B$, and a smooth bump $b$ supported in $B$ with $b(\va_*)=1$ defines $h(s,\va)=\Ind_U(s)b(\va)cz/((1-\gamma)\norm{z}_2^2)$, which vanishes on $\mathcal S_{\mathcal D}$ and shifts $\hJ_\psi(s_*,\va_*)$ by $c$.
\end{proof}

When $\mathcal S_{\mathcal D}$ instead contains an open chunk neighborhood around $\va$, any differentiable perturbation vanishing on $\mathcal S_{\mathcal D}$ has zero chunk gradient there and part~(b) is ruled out, showing that valid action gradients arise from local action coverage together with network smoothness.

\subsection{Uncontrolled gradients under uniform value accuracy}
\label{app:grad-counter}

Fix $(s,r,\pi)$ and write $J(\va):=J^\pi_r(s,\va)$. Even on a covered interval $a\in[-1,1]$, small uniform score error alone does not control action gradients: for any $\epsilon>0$, setting $J(a)=-a$ and $\hJ(a)=-a+\epsilon\sin(ka)$ with $k=2\pi n>1/\epsilon$ ($n\in\mathbb N$) gives $\sup_a|\hJ(a)-J(a)|\le\epsilon$ and exact agreement $\hJ(a)=J(a)$ on $\{-1,0,1\}$, yet $J'(0)=-1$ while $\hJ'(0)=-1+\epsilon k>0$ points toward strictly decreasing true value. Controlling action gradients therefore requires smoothness of the score error on a neighborhood.

\subsection{From value error to gradient error}
\label{app:bridge}

\begin{lemma}[Value-to-gradient bridge]\leavevmode
\label{lem:bridge}
Let $e_J:=\hJ-J$, and suppose that on the closed ball $B_b(\va)$ of radius $b>0$ around $\va$, $|e_J|\le\epsilon$ and $\norm{\nabla e_J(\vb)-\nabla e_J(\vb')}_2\le L_e\norm{\vb-\vb'}_2$. Then for every $0<\ell\le b$,
\begin{equation}
\norm{\nabla e_J(\va)}_2\le\frac{\epsilon}{\ell}+\frac{L_e\ell}{2}.
\label{eq:app-bridge}
\end{equation}
In particular, $\eta_g:=\norm{\nabla e_J(\va)}_2\le\sqrt{2L_e\epsilon}$ when $L_e,\epsilon>0$ and $\sqrt{2\epsilon/L_e}\le b$, with $\eta_g=0$ if $\epsilon=0$ and $\eta_g\le\epsilon/b$ if $L_e=0$.
\end{lemma}
\begin{proof}
For any unit vector $v$, integrating the Lipschitz gradient along $[\va,\va\pm\ell v]$ gives $e_J(\va\pm\ell v)=e_J(\va)\pm\ell\nabla e_J(\va)^\top v+R_\pm$ with $|R_\pm|\le L_e\ell^2/2$. Subtracting the two expansions yields $|2\ell\,\nabla e_J(\va)^\top v|\le 2\epsilon+L_e\ell^2$, and dividing by $2\ell$ over all unit $v$ proves \cref{eq:app-bridge}. Minimizing at $\ell=\sqrt{2\epsilon/L_e}$ (or taking $\ell\to0$ and $\ell=b$ in the degenerate cases) completes the proof.
\end{proof}

\Cref{lem:bridge} applies at interior chunks $\va$ where $e_J$ has an $L_e$-Lipschitz gradient across a two-sided ball $B_b(\va)$, converting pairwise ranking control directly into a gradient-error bound.

\begin{corollary}[Relative scores to action gradients]\leavevmode
\label{cor:ranking-gradient}
Fix $s,\pi$ and a closed ball $B_b(\va)$ inside the action domain. Let $q:=\hJ-J^\pi$ be continuously differentiable on a neighborhood of $B_b(\va)$ with $L_e$-Lipschitz gradient on $B_b(\va)$, and suppose $|q(\vb)-q(\vc)|\le\varepsilon_\Delta$ for all $\vb,\vc\in B_b(\va)$. Then for every $0<\ell\le b$,
\[
\norm{\nabla q(\va)}_2\le\frac{\varepsilon_\Delta}{2\ell}+\frac{L_e\ell}{2},
\]
which is at most $\sqrt{L_e\varepsilon_\Delta}$ whenever $L_e,\varepsilon_\Delta>0$ and $\sqrt{\varepsilon_\Delta/L_e}\le b$ (zero if $\varepsilon_\Delta=0$, and $\varepsilon_\Delta/(2b)$ if $L_e=0$).
\end{corollary}
\begin{proof}
Continuity on compact $B_b(\va)$ attains $\max_{B_b}q$ and $\min_{B_b}q$. Setting $c=(\max_{B_b}q+\min_{B_b}q)/2$ gives $|q-c|\le\varepsilon_\Delta/2$ on $B_b(\va)$ with $\nabla(q-c)=\nabla q$, so \cref{lem:bridge} with $\epsilon=\varepsilon_\Delta/2$ yields the result.
\end{proof}

Extending the candidate ranges in \cref{eq:reuse-split,eq:gap-terms} to $B_b(\va)$ via supremum minus infimum supplies $\varepsilon_\Delta=\eta(B_b)+\delta_\pi(B_b)$, and orthogonal subspace projection can only reduce $\norm{\nabla q(\va)}_2$.

\subsection{Local improvement under gradient error}
\label{app:improve-sec}
\label{app:proof-improve}

With $(s,r,\pi)$ fixed, let $\hg=\nabla\hJ(\va)$ at proposal $\va$. For a nonempty closed convex trust set $\mathcal C\ni\va$ and step parameter $\lambda>0$, consider the projected update
\begin{equation}
\va^+=\argmax_{\vb\in\mathcal C}\Big\{\hg^\top(\vb-\va)-\tfrac{1}{2\lambda}\norm{\vb-\va}_2^2\Big\}=\Pi_{\mathcal C}(\va+\lambda\hg).
\label{eq:app-local-step}
\end{equation}

\begin{proposition}[Local value improvement]\leavevmode
\label{prop:improve}
Let $J$ be differentiable with an $L_J$-Lipschitz gradient on a neighborhood of $[\va,\va^+]$, let $\norm{\hg-\nabla J(\va)}_2\le\eta_g$, and set $\delta\va:=\va^+-\va$. Then
\begin{equation}
J(\va^+)-J(\va)\ge\Big(\frac1\lambda-\frac{L_J}{2}\Big)\norm{\delta\va}_2^2-\eta_g\norm{\delta\va}_2 ,
\label{eq:improve}
\end{equation}
so $J(\va^+)>J(\va)$ whenever $0<\lambda<2/L_J$ and $\norm{\delta\va}_2>\eta_g/(1/\lambda-L_J/2)$.
\end{proposition}
\begin{proof}
Completing the square identifies $\va^+=\Pi_{\mathcal C}(\va+\lambda\hg)$ uniquely, and the projection variational inequality $(\va+\lambda\hg-\va^+)^\top(\vb-\va^+)\le0$ evaluated at $\vb=\va\in\mathcal C$ gives
\begin{equation}
\hg^\top\delta\va\ge\norm{\delta\va}_2^2/\lambda .
\label{eq:app-vi}
\end{equation}
Along $[\va,\va^+]$, $L_J$-smoothness of $J$ gives $J(\va+\delta\va)-J(\va)-\nabla J(\va)^\top\delta\va\ge-\tfrac{L_J}{2}\norm{\delta\va}_2^2$, while Cauchy--Schwarz gives $\nabla J(\va)^\top\delta\va\ge\hg^\top\delta\va-\eta_g\norm{\delta\va}_2$. Combining both bounds with \cref{eq:app-vi} yields \cref{eq:improve}.
\end{proof}

\subsection{Fixed-length steps in a subspace}
\label{app:fixed-step}

\begin{corollary}[Fixed-length subspace step]\leavevmode
\label{cor:fixed-step}
Under the assumptions of \cref{prop:improve}, let $\mathcal C=\va+\mathcal V$ for a linear subspace $\mathcal V$ with orthogonal projector $P_{\mathcal V}$, assume $P_{\mathcal V}\hg\neq0$, and set $\delta\va=\ell P_{\mathcal V}\hg/\norm{P_{\mathcal V}\hg}_2$ for $\ell>0$ with $\eta_g^{\mathcal V}:=\norm{P_{\mathcal V}(\hg-\nabla J(\va))}_2$. Then
\begin{equation}
J(\va+\delta\va)-J(\va)\ge\ell\big(\norm{P_{\mathcal V}\hg}_2-\eta^{\mathcal V}_g-L_J\ell/2\big),
\label{eq:improve-fixed}
\end{equation}
which is strictly positive whenever $\norm{P_{\mathcal V}\hg}_2>\eta_g^{\mathcal V}$ and $0<\ell<2(\norm{P_{\mathcal V}\hg}_2-\eta_g^{\mathcal V})/L_J$.
\end{corollary}
\begin{proof}
Setting $\lambda=\ell/\norm{P_{\mathcal V}\hg}_2$ in \cref{eq:app-local-step} gives $\Pi_{\va+\mathcal V}(\va+\lambda\hg)=\va+\delta\va$. Symmetry and idempotence of $P_{\mathcal V}$ give $\hg^\top\delta\va=\ell\norm{P_{\mathcal V}\hg}_2$ and $|(\hg-\nabla J(\va))^\top\delta\va|=|(P_{\mathcal V}(\hg-\nabla J(\va)))^\top\delta\va|\le\eta_g^{\mathcal V}\ell$, which combines with $L_J$-smoothness of $J$ to prove \cref{eq:improve-fixed}.
\end{proof}

\subsection{Flow-ODE guidance condition and QGF configuration}
\label{app:flow-sec}
\label{app:flow}
\label{app:qgf-error}
\label{app:qgf-impl}

Whereas \cref{cor:fixed-step} perturbs a generated chunk directly, guiding a flow-matching proposer $\dot\vb=v(\theta,\vb)$ along $\theta\in[0,1]$ requires aligning the guidance direction with the pulled-back value of the terminal flow map $\Psi_\theta(\vb)$ ($\Psi_1=\mathrm{id}$). For continuously differentiable $(\theta,\vb)\mapsto\Psi_\theta(\vb)$ and $J$, define $\mathscr H(\theta,\vb):=J(\Psi_\theta(\vb))$.

\begin{proposition}[Guided flow endpoint]\leavevmode
\label{prop:flow}
Let $\va^\theta$, $\theta\in[0,1]$, be an absolutely continuous solution of $\dot\va^\theta=v(\theta,\va^\theta)+\alpha_\theta P_{\mathcal V}\tilde g_\theta$ with measurable $\alpha_\theta\ge0$, integrable right-hand side below, and $\Delta_\theta:=\norm{P_{\mathcal V}(\tilde g_\theta-\nabla\mathscr H(\theta,\va^\theta))}_2$. Then
\begin{equation}
J(\va^1)-J(\Psi_0(\va^0))\ge\int_0^1\alpha_\theta\norm{P_{\mathcal V}\nabla\mathscr H(\theta,\va^\theta)}_2\big(\norm{P_{\mathcal V}\nabla\mathscr H(\theta,\va^\theta)}_2-\Delta_\theta\big)\,\mathrm d\theta ,
\label{eq:flow}
\end{equation}
so $J(\va^1)\ge J(\Psi_0(\va^0))$ whenever $\norm{P_{\mathcal V}\nabla\mathscr H(\theta,\va^\theta)}_2\ge\Delta_\theta$ for almost every $\theta\in[0,1]$.
\end{proposition}
\begin{proof}
Along any unguided trajectory $\vb^\theta$, $\mathscr H(\theta,\vb^\theta)=J(\vb^1)$ is constant, so $\partial_\theta\mathscr H+\nabla\mathscr H^\top v=0$ everywhere on the domain. Along the guided path $\va^\theta$, the chain rule therefore gives $\frac{\mathrm d}{\mathrm d\theta}\mathscr H(\theta,\va^\theta)=\alpha_\theta\nabla\mathscr H^\top P_{\mathcal V}\tilde g_\theta=\alpha_\theta\norm{P_{\mathcal V}\nabla\mathscr H}_2^2+\alpha_\theta(P_{\mathcal V}\nabla\mathscr H)^\top P_{\mathcal V}(\tilde g_\theta-\nabla\mathscr H)\ge\alpha_\theta\norm{P_{\mathcal V}\nabla\mathscr H}_2(\norm{P_{\mathcal V}\nabla\mathscr H}_2-\Delta_\theta)$ almost everywhere, and integrating over $\theta\in[0,1]$ proves \cref{eq:flow}.
\end{proof}

To avoid differentiating through the multi-step ODE, we adopt the QGF approximation \citep{zhou2026test}, which evaluates $\tilde g_\theta=\nabla\hJ(\hat\va^1)$ at the one-step Euler endpoint $\hat\va^1=\va^\theta+(1-\theta)v(\theta,\va^\theta)$ and approximates $\partial\Psi_\theta(\va^\theta)$ by $I$. Under critic gradient error $\eta_g$ at $\hat\va^1$, $L_J$-smoothness and gradient bound $\norm{\nabla J}_2\le\Lambda_J$, endpoint lookahead error $\norm{\hat\va^1-\Psi_\theta(\va^\theta)}_2\le\varepsilon_D$, and Jacobian deviation $\opnorm{I-\partial\Psi_\theta(\va^\theta)}\le\varepsilon_\Psi$, the resulting flow-guidance error satisfies
\begin{equation}
\Delta_\theta\le\norm{\tilde g_\theta-\nabla\mathscr H(\theta,\va^\theta)}_2\le\eta_g+L_J\varepsilon_D+\varepsilon_\Psi\Lambda_J .
\label{eq:app-qgf-error}
\end{equation}
In discrete deployment, we integrate 10 Euler steps ($i=0,\dots,9$), evaluate $\hg_i=\nabla\hJ\big(\mathrm{clip}(\va_i+(1-i/10)v_i,-1,1)\big)$, and apply the normalized subspace update
\[
\va_{i+1}=\va_i+\tfrac1{10}v_i+\alpha\sqrt{H\dim\mathcal A}\;\frac{P_{\mathcal V}\hg_i}{\max(\norm{P_{\mathcal V}\hg_i}_2,10^{-12})},
\]
with $P_{\mathcal V}\hg_i=0$ giving zero push and the terminal chunk $\va_{10}$ clipped to $[-1,1]$:
\begin{center}
\small
\begin{tabular}{@{}ll@{}}
\toprule
QGF setting & Value \\
\midrule
Base-flow integration & 10 Euler steps ($\Delta\theta=0.1$) \\
Temporal subspace $P_{\mathcal V}$ & Lowest 3 temporal DCT modes (gripper coordinates guided or masked on manipulation) \\
Normalized step scale & $\alpha\sqrt{H\dim\mathcal A}$ with norm floor $10^{-12}$, invariant to positive critic rescaling \\
Guidance strength grid & $\alpha\in\{0,0.03,0.1,0.3,1\}$ selected per task on seed-3917 development episodes \\
Range clipping & Euler endpoint estimate and final chunk $\va_{10}$ clipped to $[-1,1]$ \\
\bottomrule
\end{tabular}
\end{center}

\section{Localized Bellman residual bound on covered regions}
\label{app:local-residual}

\Cref{prop:contraction}(b) bounds successor error via the global supremum residual, whereas offline pretraining in \cref{sec:method-arch} supervises only data-covered boundary pairs $\Omega\subset\mathcal S\times\mathcal A^H$. Writing $e_F:=\hF-F^\beta_\psi\in\mathfrak B_d$ and $\delta:=\hF-\T^\beta\hF\in\mathfrak B_d$, the fixed-point identity in \cref{app:bellman} gives
\begin{equation}
e_F=\delta+\gamma^H\,\mathsf P^\beta e_F .
\label{eq:app-error-recursion}
\end{equation}
For the boundary Markov chain $(W_k)$ under $\beta$ started at $W_0=w\in\mathcal S\times\mathcal A^H$ with natural filtration $(\mathcal F_k)$, let $K_\Omega:=\inf\{k\ge0:W_k\notin\Omega\}$ denote the first exit time from measurable $\Omega$.

\begin{lemma}[Exit decomposition]\leavevmode
\label{lem:exit}
For every boundary pair $w$,
\begin{equation}
e_F(w)=\E^\beta_w\Big[\textstyle\sum_{k<K_\Omega}\gamma^{Hk}\delta(W_k)\Big]+\E^\beta_w\big[\gamma^{H K_\Omega}e_F(W_{K_\Omega})\Ind\{K_\Omega<\infty\}\big].
\label{eq:app-exit}
\end{equation}
If $\norm{\delta}_2\le\delta_\Omega$ on $\Omega$, $\norm{e_F}_2\le B_\partial$ outside $\Omega$, and $p_{\mathrm{exit}}(w):=\E^\beta_w[\gamma^{H K_\Omega}\Ind\{K_\Omega<\infty\}]$, then
\begin{equation}
\norm{e_F(w)}_2\le\frac{\delta_\Omega}{1-\gamma^H}\big(1-p_{\mathrm{exit}}(w)\big)+B_\partial\,p_{\mathrm{exit}}(w).
\label{eq:app-exit-bound}
\end{equation}
\end{lemma}
\begin{proof}
Setting $S_n:=\sum_{k<K_\Omega\wedge n}\gamma^{Hk}\delta(W_k)+\gamma^{H(K_\Omega\wedge n)}e_F(W_{K_\Omega\wedge n})$, \cref{eq:app-error-recursion} and the Markov property give $\E^\beta_w[S_{n+1}-S_n\mid\mathcal F_n]=0$, so $\E^\beta_w S_n=e_F(w)$ for all $n\ge0$. Letting $n\to\infty$ via dominated convergence yields \cref{eq:app-exit}, and bounding the pre-exit sum by $\delta_\Omega(1-\gamma^{H K_\Omega})/(1-\gamma^H)$ and the exit term on $\{W_{K_\Omega}\notin\Omega\}$ by $B_\partial\gamma^{H K_\Omega}$ proves \cref{eq:app-exit-bound}.
\end{proof}

Here $p_{\mathrm{exit}}(w)$ weights the first out-of-region error by its discounted hitting factor, recovering $\norm{e_F(w)}_2\le\delta_\Omega/(1-\gamma^H)$ whenever $\Omega$ is closed under $\mathsf P^\beta$ ($p_{\mathrm{exit}}=0$).

\section{Feature-moment readout}
\label{app:latent-sec}

\subsection{Task vectors for dense and sparse-goal rewards}
\label{app:task-code}

Let $\rho$ be a reference probability distribution over arrival states with invertible feature second-moment matrix $\Sigma_\psi:=\E_\rho[\psi\psi^\top]$. For any reward $r\in L^2(\rho)$, the population least-squares task vector $z_r:=\Sigma_\psi^{-1}\E_\rho[\psi\,r]$ defines the $L^2(\rho)$-orthogonal projection $r_{z_r}=\psi^\top z_r$ onto $\spn(\psi)$, recovering $z_r=z$ whenever $r=\psi^\top z$ and otherwise entering the uniform reward-approximation term $\varepsilon_r$ in \cref{app:score-budgets}. On dense-reward DMC tasks, \cref{sec:method-readout} estimates $z_r$ by ridge regression on 10k cached arrivals as detailed in \cref{app:zs-protocol}.

For sparse-goal OGBench suites under the feature-moment readout, with temperature $T>0$, signed weights $w_i:=\mathrm{softmax}(T\,r)_i\,r_i$, and empirical feature mean $m:=\psi_{\mathrm{ref}}$ over cached arrivals $\{s'_i\}$, the centered task vector is
\[
z=\sqrt d\,\frac{u}{\norm{u}_2},\qquad
u:=\sum_i w_i\big(\psi(s'_i)-m\big)=\sum_i w_i\psi(s'_i)-\Big(\sum_i w_i\Big)m,
\]
whenever $u\neq0$ (if all cached rewards coincide, $u=0$). Unlike centering $\psi$ under a fixed $z$, subtracting $m$ inside $u$ removes the state-dependent component $(\sum_i w_i)\psi(s)^\top m$ and aligns $z$ with the displacement from the dataset mean toward high-reward states, whereas positive rescaling to radius $\sqrt d$ preserves candidate orderings and normalized gradients. In FB coordinates with backward map $B$ and goal distribution $p_{\mathrm g}\ll\rho$, the analogous goal code is
\begin{equation}
\E_{s\sim p_{\mathrm g}}B(s)=\E_\rho\Big[B(s)\,\frac{\mathrm dp_{\mathrm g}}{\mathrm d\rho}(s)\Big],
\label{eq:app-demo-code}
\end{equation}
corresponding to the density-ratio reward $\mathrm dp_{\mathrm g}/\mathrm d\rho$ up to approximation residuals captured in $\varepsilon_r$.

\subsection{Moment regression and occupancy projection}
\label{app:pythagoras}

When $M^\beta(\cdot\mid s,\va)$ admits a density $q=\mathrm dM^\beta/\mathrm d\rho\in L^2(\rho)$ of total mass $(1-\gamma)^{-1}$, \cref{lem:measure-latent} identifies $F^\beta_\psi(s,\va)=\E_\rho[\psi\,q]$, and fitting this feature moment is equivalent in the metric $\Sigma_\psi^{-1}$ to fitting the $\spn(\psi)$-projection $\Pi_\psi q$ of $q$.

\begin{lemma}\leavevmode
\label{lem:pythagoras}
For $q\in L^2(\rho)$ with moment $F_q:=\E_\rho[\psi\,q]$ and $L^2(\rho)$-orthogonal projector $\Pi_\psi$ onto $\spn(\psi)$, every $v\in\R^d$ satisfies
\begin{equation}
\norm{q-\psi^\top \Sigma_\psi^{-1}v}_{L^2(\rho)}^2=\norm{q-\Pi_\psi q}_{L^2(\rho)}^2+(v-F_q)^\top \Sigma_\psi^{-1}(v-F_q).
\label{eq:app-pythagoras}
\end{equation}
\end{lemma}
\begin{proof}
Writing $\Pi_\psi q=\psi^\top c$, orthogonality $\E_\rho[\psi(q-\psi^\top c)]=0$ gives $c=\Sigma_\psi^{-1}F_q$, so $\Pi_\psi q=\psi^\top \Sigma_\psi^{-1}F_q$. Splitting $q-\psi^\top \Sigma_\psi^{-1}v=(q-\Pi_\psi q)+\psi^\top \Sigma_\psi^{-1}(F_q-v)$ into orthogonal components in $L^2(\rho)$ and evaluating $\norm{\psi^\top \Sigma_\psi^{-1}(F_q-v)}_{L^2(\rho)}^2=(v-F_q)^\top \Sigma_\psi^{-1}(v-F_q)$ proves \cref{eq:app-pythagoras}.
\end{proof}

\subsection{Linear composition of task vectors}
\label{app:composition}

Under fixed continuation $\beta$, linearity of expectation in the reward implies that for any linear combination $z_{\mathrm{mix}}=\sum_ic_iz_i$ with $r_{z_{\mathrm{mix}}}=\sum_ic_i r_{z_i}$,
\begin{equation}
Q^\beta_{r_{z_{\mathrm{mix}}}}=\sum_ic_iQ^\beta_{r_{z_i}},\qquad
\nabla_{\va}Q^\beta_{r_{z_{\mathrm{mix}}}}=\sum_ic_i\nabla_{\va}Q^\beta_{r_{z_i}},
\label{eq:app-composition}
\end{equation}
whenever differentiable. Because the predictor $T_\phi(\phi(s),\va)$ takes no task vector as input, the learned feature-moment scores $\hJ_\psi(s,\va)=(1-\gamma)\hF(s,\va)^\top z=T_\phi(\phi(s),\va)^\top z$ (as well as fixed-support density scores in the reward vector) compose identically, whereas argmax selection, normalized gradient steps, and sequential multi-stage tasks do not compose linearly.

\section{Categorical density readout}
\label{app:density-theory}

The categorical Bellman density readout builds on density-ratio and contrastive occupancy estimation \citep{anonymous2026density,blier2021learning,eysenbach2021clearning,eysenbach2022contrastive,zheng2024contrastive}. Below we state the chunk-level estimator on frozen representations and summarize the finite-support and reference-sampling bounds supplying the score-error budget in \cref{app:score-budgets}.

\subsection{Chunk-level estimator formulation}
\label{app:density-estimator}

With the pretrained backbone frozen under either target in \cref{app:terminal}, write $x=(s,\va)$, $x'=(s_H,\va')$, $\E_x:=\E_{s,\va}$, and $Y_j:=s_j$ ($j=1,\dots,H$). Two forward heads read the fused chunk features at $x$ and a backward head reads the standardized arrival features $\psi(s^+)$ at an arrival state $s^+$, producing mean inner-product logits $f(x,s^+)$. On each training anchor $x$, the support multiset is $S=(Y_1,\dots,Y_H,Z_1,\dots,Z_K)$ ($K=2048$) with $Z_i\stackrel{\mathrm{i.i.d.}}{\sim}\rho$ drawn with replacement from the empirical training-arrival distribution $\rho$. Writing $p_f(i\mid x,S):=\exp f(x,S_i)/\sum_j\exp f(x,S_j)$ and $\bar f$ for the mean EMA teacher logits, each online forward head $h$ minimizes $-\sum_{i=1}^{H+K}t_i\log p_{f_h}(i\mid x,S)$ against the unit-mass distributional backup target
\[
t_i=(1-\gamma)\gamma^{i-1}\Ind\{i\le H\}+\gamma^H\,p_{\bar f}(i\mid x',S).
\]
At test time, identifying the retained reference arrivals $Z_1,\dots,Z_{K_{\mathrm{ref}}}\stackrel{\mathrm{i.i.d.}}{\sim}\rho$ ($1\le K_{\mathrm{ref}}\le 10^4$) with the cached states $\{s'_i\}$ and softmax weights $p(i\mid s,\va)$ in \cref{sec:method-readout}, the score for labels $r_i$ is
\[
\hJ_D(x)=\sum_{i=1}^{K_{\mathrm{ref}}}\frac{\exp f(x,Z_i)}{\sum_j\exp f(x,Z_j)}\,r_i ,
\]
which forms a convex combination of $\{r_i\}$. A constant reward shift therefore moves every score equally and preserves candidate rankings, while pair-dependent logit offsets $c(x)$ leave the score unchanged. For any logit map $f$ with finite normalizer $Z_f(x):=\int e^{f(x,s^+)}\rho(\mathrm ds^+)\in(0,\infty)$, define $q_f(x,s^+):=e^{f(x,s^+)}/Z_f(x)$ and $C_f(x,\mathrm ds^+):=q_f(x,s^+)\rho(\mathrm ds^+)$, and let $\mathcal T C:=a+\gamma^H\mathsf P^\beta C$ with $a_x:=(1-\gamma)\E_x\sum_{j=1}^H\gamma^{j-1}\delta_{Y_j}$ denote the probability-measure backup. By \cref{lem:measure-latent}, $m_\beta=\mathcal Tm_\beta$, and this fixed point is unique because $\mathcal T$ contracts $\sup_x\TV$ with modulus $\gamma^H$. Under exact population-normalized cross-entropy minimization \citep{anonymous2026density}, iterates satisfy $\sup_x\TV(C_{f_n}(x),m_\beta(x))\le\gamma^{Hn}\sup_x\TV(C_{f_0}(x),m_\beta(x))$ and determine $\log(\mathrm dm_\beta(x)/\mathrm d\rho)$ up to a pair-dependent log-normalizer, since a limit $f^\ast$ with $C_{f^\ast}=m_\beta$ satisfies $f^\ast(x,s^+)=\log(\mathrm dm_\beta(x)/\mathrm d\rho)(s^+)+c(x)$ with $c(x)=\log Z_{f^\ast}(x)$, which cancels in each softmax.

\subsection{Finite-support cross-entropy certificate}
\label{app:density-finite}

In practice, softmax normalization acts over the finite support $S=(Y_1,\dots,Y_H,Z_1,\dots,Z_K)$, minimizing $L_x^{(K)}(f\mid g):=-\E[\sum_{i=1}^{H+K}t_i\log p_f(i\mid x,S)\mid x]$ against frozen teacher logits $g$.

\begin{lemma}[Sampled cross-entropy stationarity]\leavevmode
\label{lem:sampled-ce}
Assume \textnormal{(M1)}: $S=(Y_1,\dots,Y_H,Z_1,\dots,Z_K)$ keeps repeated slots with $Z_i\stackrel{\mathrm{i.i.d.}}{\sim}\rho$ independent of $(Y_{1:H},x')$. With $R_f(x,S):=\sum_i p_f(i\mid x,S)\delta_{S_i}$ and $a^{\mathrm{path}}:=(1-\gamma)\sum_{j=1}^H\gamma^{j-1}\delta_{Y_j}$, every bounded measurable perturbation $v:\mathcal S\to\R$ at $x$ satisfies
\[
\frac{\mathrm d}{\mathrm d\epsilon}L_x^{(K)}(f+\epsilon v\mid g)\Big|_{\epsilon=0}=\int v\,\mathrm d\Gamma_{f,g}(x),\qquad
\Gamma_{f,g}(x):=\E\big[R_f(x,S)-a^{\mathrm{path}}-\gamma^H R_g(x',S)\,\big|\,x\big].
\]
\end{lemma}
\begin{proof}
Differentiating $-\sum_i t_i\log p_{f+\epsilon v}(i\mid x,S)$ at $\epsilon=0$ using $\sum_i t_i=1$ yields $\sum_i(p_f(i\mid x,S)-t_i)v(S_i)$, bounded by $2\supnorm v$, so differentiating under the expectation gives $\int v\,\mathrm d\Gamma_{f,g}(x)$.
\end{proof}

\begin{proposition}[Finite-support residual certificate]\leavevmode
\label{thm:finite-support}
Assume \textnormal{(S1)--(S5)}, \textnormal{(M1)}, and the following across a $\mathsf P^\beta$-closed continuation domain $\Omega$ (including bootstrap-only pairs): \textnormal{(M2)}~$0<q_f(x,s^+),q_g(x,s^+)\le B<\infty$ for mean online and EMA teacher logits $f,g$, \textnormal{(M3)}~$\norm{\E[R_f(x,S)\mid x]-\frac12\sum_{h=1}^2\E[R_{f_h}(x,S)\mid x]}_{\TV}\le\epsilon_{\mathrm{head}}$, and \textnormal{(M4)}~$\norm{\Gamma_{f_h,g}(x)}_{\TV}\le\epsilon_{\mathrm{fit}}$ for $h\in\{1,2\}$ with $\TV(C_f(x),C_g(x))\le\epsilon_{\mathrm{EMA}}$. With $b_{H,K}:=\min\{1,\sqrt{(B-1)/K}+HB/K\}$,
\[
\sup_{x\in\Omega}\TV\big(C_f(x),m_\beta(x)\big)\le\delta_D:=\min\Big\{1,\frac{\epsilon_{\mathrm{fit}}+\epsilon_{\mathrm{head}}+(1+\gamma^H)\,b_{H,K}+\gamma^H\,\epsilon_{\mathrm{EMA}}}{1-\gamma^H}\Big\}.
\]
\end{proposition}
\begin{proof}
By (M3)--(M4), $\norm{\Gamma_{f,g}(x)}_{\TV}\le\epsilon_{\mathrm{fit}}+\epsilon_{\mathrm{head}}$. For $u\in\{x,x'\}$, $w=q_f(u,\cdot)$ (or $q_g(u,\cdot)$), and measurable $0\le h\le1$, setting $D=K^{-1}\sum_{i=1}^K w(Z_i)$, $A=K^{-1}\sum_{i=1}^K w(Z_i)h(Z_i)$, $U=K^{-1}\sum_{j=1}^H w(Y_j)$, and $V=K^{-1}\sum_{j=1}^H w(Y_j)h(Y_j)$ writes $R:=R_f(u,S)(h)=(A+V)/(D+U)$ as
\[
R-\textstyle\int wh\,\mathrm d\rho=\big(A-\int wh\,\mathrm d\rho\big)+R(1-D)+(V-UR).
\]
Since $\E_Z A=\int wh\,\mathrm d\rho$, $0\le R\le1$, $|V-UR|\le U\le HB/K$, and $\E_Z|D-1|\le\sqrt{\Var_\rho(w)/K}\le\sqrt{(B-1)/K}$ by $\E_\rho w=1$ and $w^2\le Bw$, independence of $Z_{1:K}$ from $(Y_{1:H},x')$ gives $\norm{\E[R_f(x,S)\mid x]-C_f(x)}_{\TV}\le b_{H,K}$ and $\norm{\E[R_g(x',S)\mid x]-(\mathsf P^\beta C_g)_x}_{\TV}\le b_{H,K}$ under (S5). Decomposing $C_f-\mathcal TC_f=\Gamma_{f,g}-(\E[R_f\mid x]-C_f)+\gamma^H(\E[R_g\mid x]-\mathsf P^\beta C_g)+\gamma^H\mathsf P^\beta(C_g-C_f)$ and applying $\gamma^H$-contraction of $\mathcal T$ in $\TV$ on $\Omega$ yields the bound.
\end{proof}

As $K\to\infty$ for fixed $(H,B)$, $b_{H,K}\to0$, showing that the finite-support normalization error vanishes asymptotically as support size grows.

\subsection{Self-normalized reference readout and score budget}
\label{app:density-readout}
\label{app:density-value}

\begin{proposition}[Self-normalized reference readout]\leavevmode
\label{thm:snis}
Assume \textnormal{(E1)}: $Z_1,\dots,Z_K\stackrel{\mathrm{i.i.d.}}{\sim}\rho$ ($K=K_{\mathrm{ref}}\ge1$) are independent of the logits, fixed query $x$, and reward, and \textnormal{(E2)}: $r:\mathcal S\to[r_-,r_+]$ has range $R:=r_+-r_-$. For $w\ge0$ with $\E_\rho w=1$ and $C(\mathrm ds^+):=w(s^+)\rho(\mathrm ds^+)$, set $v:=\int r\,\mathrm dC$ and $\hat v_K:=\sum_{i=1}^K w(Z_i)r(Z_i)/\sum_{i=1}^K w(Z_i)$ when $\sum_i w(Z_i)>0$ ($0$ otherwise). Then $\hat v_K\to v$ almost surely as $K\to\infty$, and for any $\delta\in(0,1)$, with probability at least $1-\delta$:
\begin{enumerate}[label=(\alph*),leftmargin=*,nosep]
\item if $w\le B$ and $Bt<1$ with $t:=\sqrt{\log(4/\delta)/(2K)}$, then $|\hat v_K-v|\le BRt/(1-Bt)$, and
\item if $\E_\rho w^2\le C_2$ and $a:=\sqrt{2(C_2-1)/(K\delta)}<1$, then $|\hat v_K-v|\le R\sqrt{2C_2/(K\delta)}/(1-a)$.
\end{enumerate}
\end{proposition}
\begin{proof}
Shifting $r$ by $-r_-$ places $r,v\in[0,R]$ and writes $\hat v_K-v=A_K/D_K$ with $D_K:=K^{-1}\sum_i w(Z_i)$ and $A_K:=K^{-1}\sum_i w(Z_i)(r(Z_i)-v)$. The strong law gives $D_K\to1$ and $A_K\to0$ almost surely, while Hoeffding's inequality in~(a) and Chebyshev's inequality in~(b) bound $|D_K-1|$ and $|A_K|$ simultaneously via a union bound.
\end{proof}

For $N$ candidates drawn independently of $Z_{1:K}$ at a fixed state, replacing $\delta$ by $\delta/N$ bounds all $N$ simultaneously, whereas states visited after adaptive selections depend on the cached reference set.

\begin{corollary}[Density score-error budget]\leavevmode
\label{cor:density-value}
Assume \textnormal{(S1)--(S5)}, $\TV(C_f(x),m_\beta(x))\le\delta_D$ on $\Omega$ from \cref{thm:finite-support}, and \textnormal{(V1)}: the transition reward admits an arrival-measurable proxy $r_0$ of range $R_0:=\sup r_0-\inf r_0$ with $|r_t-r_0(s_{t+1})|\le\varepsilon_{\mathrm{proxy}}$ and label error $|r_i-r_0(Z_i)|\le\epsilon_{\mathrm{lab}}$. On an event where \cref{thm:snis} applied to $r_0$ gives $|\hat v_K-\int r_0\,\mathrm dC_f(x)|\le\epsilon_K(x)$, we have $|\hJ_D(x)-J^\beta_r(x)|\le\varepsilon_\beta(x):=R_0\,\delta_D+\epsilon_K(x)+\epsilon_{\mathrm{lab}}+\varepsilon_{\mathrm{proxy}}$. If in addition \textnormal{(V2)}~$|V^\beta_r-V^\pi_r|\le D_V$ across reachable post-chunk states for a continuation $\pi$, then
\[
|\hJ_D(x)-J^\pi_r(x)|\le\epsilon_D(x):=R_0\,\delta_D+\epsilon_K(x)+\epsilon_{\mathrm{lab}}+\varepsilon_{\mathrm{proxy}}+(1-\gamma)\gamma^H D_V .
\]
\end{corollary}
\begin{proof}
Convexity of softmax weights bounds label error by $\epsilon_{\mathrm{lab}}$, \cref{thm:snis} contributes $\epsilon_K(x)$, $|\int r_0\,\mathrm d(C_f(x)-m_\beta(x))|\le R_0\delta_D$, $|J^\beta_{r_0}(x)-J^\beta_r(x)|\le\varepsilon_{\mathrm{proxy}}$, and \cref{eq:app-continuation} contributes $(1-\gamma)\gamma^H D_V$.
\end{proof}

\section{Comparison with actor-coupled successor methods}
\label{app:fb}

Whereas actor-coupled successor methods bootstrap at actions sampled from a jointly trained policy whose greedy operator need not be a $\gamma$-contraction \citep[Prop.~3]{zheng2026can}, \method{} replaces actor-sampled bootstrap actions with the next recorded chunk $\va'$ under $\beta$ so that \cref{prop:contraction} applies directly. \Cref{tab:inherit} summarizes the architectural and training differences between TD-JEPA \cite{bagatella2026td} and \method{}.

\begin{table}[!htbp]
\centering
\caption{\textbf{\method{} retains TD-JEPA's joint-embedding backbone while replacing actor-coupled bootstrapping with recorded action chunks and removing task conditioning from the predictor.} Component comparison between TD-JEPA \cite{bagatella2026td} and \method{}.}
\label{tab:inherit}
\footnotesize
\setlength{\tabcolsep}{4pt}
\begin{tabular}{@{}p{2.6cm}p{4.9cm}p{5.4cm}@{}}
\toprule
Component & TD-JEPA & \method{} \\
\midrule
Encoders & State encoder $\phi$, arrival encoder $\psi$ (single state or 3-frame DrQ stack) & Conditioning encoder $\phi$, arrival encoder $\psi$ (16-step state window or 3-frame DrQ stack) \\
Predictor & $T_\phi(\phi(s),a,z)$ conditioned on primitive action $a$ and task $z$ & $T_\phi(\phi(s),\va)$ conditioned on $H$-step action chunk $\va$ with no task input \\
Symmetric predictor & $T_\psi(\psi(s),a,z)$ & $T_\psi(\psi(s),\va)$ \\
Target networks & EMA of $\phi,\psi,T_\phi,T_\psi$ & EMA of $\phi,\psi,T_\phi,T_\psi$ (decay $0.995$) \\
Bootstrap action & $a'\sim\pi_z(\cdot\mid s')$ from a jointly trained actor & Next recorded chunk $\va'$ from $\mathcal D$ under fixed behavior $\beta$ \\
Horizon & Single step, discount $\gamma$ & Multi-horizon $H\in\{4,8,16,32\}$ in training ($H{=}8$ at deployment), per-chunk discount $\gamma^H$ \\
Regularizer & Orthonormalization of $\phi$ and $\psi$ & Variance--covariance penalties, plus proprioceptive state grounding \\
Task readout & $T_\phi(\phi(s),a,z)^\top z$ & $T_\phi(\phi(s),\va)^\top z$ or categorical density score $\hJ_D(s,\va)$ \\
Actor training & Jointly trained task-conditioned actor $\pi_z$ & None (zero-update test-time guidance of unseen policies) \\
\bottomrule
\end{tabular}
\end{table}

\section{Discounted value bounds and benchmark evaluation metrics}
\label{app:not-claimed}

The theoretical results in \cref{app:prop1,app:ranking-sec,app:rolling,app:grad-sec} analyze the continuing $\gamma$-discounted objective under the stated Markov-boundary, coverage, and approximation conditions. Standard zero-shot RL reporting evaluates closed-loop finite-episode undiscounted return on DMC and any-step episode goal completion on OGBench. Complementing closed-loop benchmark rollouts, the counterfactual diagnostic in \cref{app:zs-rank} directly measures discounted return gains from a single chunk decision followed by base-policy continuation, while finite-$K$ regression and readout terms follow \cref{app:loss-not-cert,app:density-finite,app:density-readout}.

\section{Architecture, training objectives and algorithm}
\label{app:impl}

This section details the network architecture, pretraining losses, and deployment procedure of \cref{sec:method}, complementing the state-suite hyperparameters in \cref{tab:zs-abi}.

\subsection{Architecture}
\label{app:arch}

\Cref{fig:supp-arch} summarizes the observation encoders, task-free chunk predictor, and frozen-backbone readout modules of \cref{sec:method-arch}. On proprioceptive benchmarks (\cref{fig:supp-arch}a, top), $\phi$ and $\psi$ encode a 16-step window of normalized simulator states ($1$ step in the no-history ablation). Each temporal transformer uses two pre-norm layers of width 128 and 4 attention heads over the concatenated normalized states, presence masks, and step-timing features (relative offset, control period, and validity). On visual benchmarks, observations consist of three stacked $64\times64$ RGB frames mapped by an end-to-end DrQ convolutional encoder \citep{yarats2021image} into the same stream.

In each cross-attention fusion block, 4 learned query tokens attend independently to each active input stream, sum the normalized updates scaled by $1/\sqrt{\#\text{streams}}$, and pass through a feed-forward layer to output unconstrained embeddings $\phi(s),\psi(s)\in\R^d$ ($d=256$). In the task-free predictor $T_\phi(\phi(s),\va)$, the query tokens of $\phi(s)$ cross-attend to encoded tokens of the $H$-step chunk $\va$ (incorporating normalized step count, physical duration, and sinusoidal horizon features of $H$), followed by two MLP heads of hidden width 256 whose outputs are averaged at test time.

When using the categorical density readout \cite{anonymous2026density}, two forward heads read the predictor's fused $(s,\va)$ features and one backward head reads standardized $\psi(s^+)$ features. Each head is a 3-layer MLP of width 512 (GELU, LayerNorm) and output dimension 128 (with $\sqrt{128}$-radius output normalization on the backward head for ExORL), trained on the frozen backbone as specified in \cref{app:density-estimator,app:zs-abi} (\cref{fig:supp-arch}c).

\begin{figure}[!htbp]
\centering
\includegraphics[width=\textwidth]{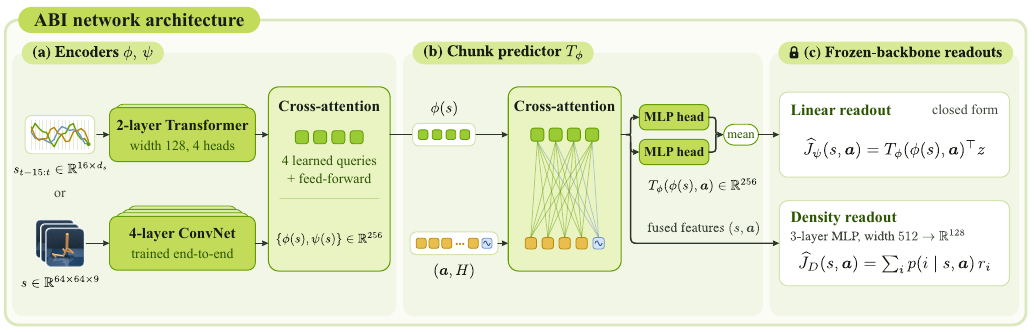}
\caption{\textbf{\method{} decouples reward-free multi-horizon representation pretraining from zero-update task readouts.} \emph{(a)} Proprioceptive state windows ($16$ steps) or stacked $64\times 64$ RGB frames pass through modality-specific encoders and 4-query cross-attention into unconstrained embeddings $\phi(s),\psi(s)\in\R^{256}$. \emph{(b)} The task-free predictor cross-attends query tokens to encoded $H$-step chunks $(\va,H)$ without task inputs and averages two width-256 MLP heads. \emph{(c)} At deployment, the frozen backbone supports closed-form linear scoring $\hJ_\psi$ or categorical Bellman density scoring $\hJ_D$ (\cref{app:density-estimator}).}
\label{fig:supp-arch}
\end{figure}

\subsection{Training objectives}
\label{app:losses}

All regression targets are evaluated without gradients through EMA copies $(\overline\phi,\overline\psi,\overline T_\phi,\overline T_\psi)$ with horizon $H\in\mathcal H$ sampled as in \cref{app:multi-h}.

\paragraph{Successor and within-chunk TD losses.} Under the default successor target, $\mathcal L_{\mathrm{TD}}$ regresses $T_\phi(\phi(s_0),\va)$ onto \cref{eq:app-td-target} with mean squared error averaged across coordinates and both predictor heads, while $\mathcal L^{\mathrm{sym}}_{\mathrm{TD}}$ applies the symmetric regression to $T_\psi(\psi(s_0),\va)$ against $\overline\phi$ arrivals and $\overline T_\psi(\overline\psi(s_H),\va')$. Under the within-chunk target variant of \cref{app:within-chunk}, both losses regress directly onto the normalized within-chunk mean $\overline\psi_H$ (and $\overline\phi_H$) without querying target predictors.

\paragraph{Anti-collapse regularization.} Both encoders receive variance hinge $\frac1d\sum_i(1-\sqrt{\Var f_i+10^{-4}})_+$ and off-diagonal covariance penalty $\frac{\lambda_{\mathrm c}}{d}\sum_{i\ne j}\mathrm{Cov}(f)_{ij}^2$ \citep{bardes2022vicreg} with $\lambda_{\mathrm c}=0.01$. On state inputs, an auxiliary head $g$ reconstructs the normalized current state and 8-step state displacement from $\psi$ with weight $\lambda_{\mathrm g}=0.1$, whereas pixel inputs omit state reconstruction. Together these terms form $\mathcal L_{\mathrm{reg}}$.

\subsection{Algorithm}
\label{app:algorithm}

\begin{algorithm}[!htbp]
\caption{\method{}: actor-free chunk representation pretraining and zero-update policy guidance}
\label{alg:abi}
\begin{algorithmic}[1]
\Statex \textbf{Pretraining} on unlabeled trajectories $\mathcal D$ (no rewards, no task inputs, no actor)
\For{each update}
  \State Sample anchors $s_0$ with $\ge 2H$ remaining steps, horizons $H\in\mathcal H$, chunks $\va$, and next recorded chunks $\va'$ of length $H$ from $\mathcal D$
  \State Form default targets via \cref{eq:app-td-target} or within-chunk targets $y=\overline\psi_H$ using EMA networks
  \State Update online parameters on $\mathcal L_{\mathrm{TD}}+\mathcal L^{\mathrm{sym}}_{\mathrm{TD}}+\mathcal L_{\mathrm{reg}}$ and update EMA networks with decay $\omega$
\EndFor
\State Freeze EMA backbone $(\phi,\psi,T_\phi)$ and optionally train reward-free density heads as in \cref{app:density-estimator}
\Statex \textbf{Test-time deployment} with unseen proposer $\mu$ and task rewards on cached arrivals $\{s'_i\}$
\State Construct task vector $z$ for $\hJ_\psi$ or cache $\{s'_i\}$ with rewards $\{r_i\}$ for $\hJ_D$ as in \cref{sec:method-readout}
\For{each replanning step at state $s$}
  \State Sample candidates at $H{=}8$: 16 draws from $\mu$ (BFN16), 1 guided draw (QGF), or 4 guided draws plus 1 unguided draw from $\mu$ (QGF+BFN4), integrating guided draws with $P_{\mathcal V}$ as in \cref{app:qgf-impl}
  \State Select $\hat\va\in\argmax_{\va\in\mathcal A_s}\hJ(s,\va)$ and execute $\hat a_0$ ($H_{\mathrm{exec}}{=}1$, replanning every step) or $\hat\va$ ($H_{\mathrm{exec}}{=}8$)
\EndFor
\end{algorithmic}
\end{algorithm}

\FloatBarrier

\section{Experimental details and additional results}
\label{app:zeroshot}

This section provides experimental protocols, hyperparameter configurations, and extended empirical evaluations complementing \cref{sec:exp}. \Cref{app:zs-protocol,app:zs-abi,app:zs-guidance} specify datasets, task inference, pretraining configurations, and proposer guidance settings, while \cref{app:zs-pixels} details visual control and object-sensitivity probing. Critic comparisons, ranking, and external-actor reuse appear in \cref{app:zs-swap,app:zs-rank,app:zs-steer-actors}, with horizon and proposer ablations in \cref{app:zs-lengths,app:zs-flowbcplus}. Baseline reproductions, sample-scaling curves, and full tables follow in \cref{app:zs-baselines,app:zs-convergence,app:zs-all}, matched-readout baseline evaluations in \cref{app:zs-baseline-density}, and rollout progressions alongside OGBench data and tasks in \cref{app:zs-qualitative,app:zs-ogbench}.

\subsection{Data, tasks and evaluation protocol}
\label{app:zs-protocol}

We follow the data processing, task-inference interface, and evaluation harness of \citet{bagatella2026td} at its public release, quoting reported baseline numbers in \cref{tab:zeroshot} from \citet{bagatella2026td}.

\paragraph{Datasets and tasks.} ExORL \cite{yarats2022don} comprises the first 5{,}000 RND exploratory episodes (MuJoCo 3.3.7) across 20 dense-reward tasks: \texttt{walker} (stand, walk, run, spin), \texttt{cheetah} (walk, walk backward, run, run backward), \texttt{quadruped} (stand, walk, run, jump), and \texttt{pointmass} (reach top-left, top-right, bottom-left, bottom-right, bottom-left-long, loop, square, fast--slow). OGBench \cite{park2025ogbench} comprises the full trajectories of \texttt{antmaze-\{medium,large\}-navigate}, \texttt{antmaze-\{medium,large\}-stitch}, \texttt{antmaze-medium-explore}, \texttt{cube-\{single,double\}-play}, \texttt{scene-play}, and \texttt{puzzle-3x3-play}, evaluated on the five standard goals per dataset for 45 tasks in total, with data and task layouts illustrated in \cref{app:zs-ogbench}. None of the OGBench datasets demonstrates the evaluation tasks, which accounts for the low success rate of unguided Flow BC in \cref{sec:exp-zeroshot}.

\paragraph{Task inference.} For each training seed, 10k transitions $(s,a,s')$ are sampled from the training set, with task rewards evaluated from $(a,s')$ and OGBench rewards shifted by $+1$. The density readout $\hJ_D$ \cite{anonymous2026density} in \cref{sec:method-readout} scores cached states directly via $\hJ_D(s,\va)=\sum_k p(k\mid s,\va)r_k$. For the feature-moment readout $\hJ_\psi$, dense DMC tasks use the ridge vector $z=(\Psi^\top\Psi/n+\lambda I)^{-1}\Psi^\top r/n$ on $\psi(s'_i)$ with $\lambda=10^{-4}\lambda_{\max}(\Psi^\top\Psi/n)$, whereas sparse-goal OGBench tasks use the centered vector $z=\sqrt d\,u/\norm{u}_2$ with $u=\sum_i\mathrm{softmax}(Tr)_i r_i(\psi(s'_i)-\psi_{\mathrm{ref}})$ from \cref{app:task-code}. One-step FB applies the same softmax weighting to its backward features with its native temperatures ($T{=}10$ on \texttt{walker}, \texttt{pointmass}, and \texttt{antmaze}, $T{=}3$ on \texttt{cheetah} and \texttt{quadruped}, and $T{=}300$ on manipulation).

\paragraph{Splits and metrics.} Hyperparameters and guidance strengths are selected on 10 development episodes per task (reset seeds $30000,\dots,30009$) and frozen prior to test evaluation on 20 disjoint test episodes per task (reset seeds $40000,\dots,40019$) across 3 training seeds (3917, 3502, 8948). Unguided Flow BC, reproduced TD-JEPA, and one-step FB are evaluated on the same test split (using 10 episodes per task on DMC for rerun actor baselines, matching the released TD-JEPA launchers). We report episode return on DMC and any-step goal completion (\%) on OGBench. Dataset entries report mean\,$\pm$\,s.e.\ across the 3 training seeds. Suite averages report s.e.\ across the 3 per-seed suite means, except in \cref{tab:zeroshot-full-h8,tab:zeroshot-variants}, which propagate per-dataset s.e.\ in quadrature.

\FloatBarrier
\subsection{\method{} configuration on state inputs}
\label{app:zs-abi}

\Cref{tab:zs-abi} lists the state-benchmark pretraining configuration of \cref{sec:method-arch,app:impl} (\cref{fig:supp-arch}). Pretraining minimizes $\mathcal L_{\mathrm{TD}}+\mathcal L^{\mathrm{sym}}_{\mathrm{TD}}+\mathcal L_{\mathrm{reg}}$ from \cref{app:losses} with $\lambda_{\mathrm c}=0.01$ and $\lambda_{\mathrm g}=0.1$ using no rewards, task inputs, or policy updates, and deployment queries the frozen EMA critic at $H{=}8$ from the 16-step state window.

For $\hJ_D$, the reward-free density heads of \cref{app:arch,app:density-estimator} train for $62.5\text{k}$ updates on the frozen backbone with $K{=}2048$ negative arrivals per anchor and evaluate at test time on the retained non-held-out subset ($7{,}746$--$8{,}200$ states) of the 10k task-inference states analyzed in \cref{thm:snis}, whereas $\hJ_\psi$ uses all 10k states in closed form. In \cref{tab:zeroshot-variants}, \emph{No history} restricts $\phi$ and $\psi$ to the current state, and \emph{Within-chunk target} sets $y=\overline\psi_H$ in backbone pretraining while retaining $\gamma^H$ distributional bootstrapping in the density heads as described in \cref{app:within-chunk}. The suite configuration pairs the default successor backbone with $\hJ_\psi$ on dense-reward DMC and pairs the within-chunk backbone with $\hJ_D$ on sparse-goal OGBench, with all variants reported in \cref{tab:zeroshot-variants}. In \cref{fig:valuemap}, the left panel plots $\hJ_D(s,\va)$ (log scale, clipped) across 4{,}000 recorded pairs on \texttt{antmaze-mn}, and the right panels show same-start development rollouts (seed 3917, BFN16, $H_{\mathrm{exec}}{=}8$), extended across eight distinct tasks in \cref{app:zs-qualitative} (\cref{fig:supp-rollouts}).

\begin{table}[!htbp]
\centering
\caption{\textbf{All 13 state benchmarks share a single \method{} architecture and optimization schedule.} Pretraining hyperparameters for proprioceptive state suites, which differ across suites only in the discount factor $\gamma$ and the backbone target.}
\label{tab:zs-abi}
\small
\begin{tabular}{@{}ll@{}}
\toprule
Latent dimension $d$, predictor MLP width & 256, 256 (unconstrained norm), temporal transformer width 128 \\
Parameters & 7.7M (backbone), 3.0M (density heads) \\
State-history window & 16 steps (no-history ablation: 1 step) \\
Training chunk horizons $\mathcal H$ & $\{4,8,16,32\}$ sampled with probabilities $(\tfrac16,\tfrac16,\tfrac16,\tfrac12)$ \\
Backbone target & Default successor (DMC) or within-chunk arrival mean (OGBench) \\
Discount $\gamma$ & 0.98 (DMC), 0.99 (OGBench) per primitive step \\
Batch size and updates & 2048 anchors (512 episodes $\times$ 4), $62.5\text{k}$ updates ($1.28\times10^8$ samples) \\
Optimizer and schedule & AdamW (weight decay $10^{-4}$, clip 1), lr $3\times10^{-4}$ cosine to $3\times10^{-5}$ after 10\% warmup \\
EMA decay $\omega$ and precision & 0.995, bf16 mixed precision \\
\bottomrule
\end{tabular}
\end{table}

\FloatBarrier
\subsection{Flow BC proposer, selection and guidance}
\label{app:zs-guidance}

\paragraph{Flow BC proposers.} A residual MLP with LayerNorm and GELU maps the normalized state, noisy 8-step chunk, and flow time $\theta\in[0,1]$ to a velocity field integrated over 10 Euler steps. DMC uses a 2-layer width-256 MLP ($2\times10^5$ updates, batch 256, Adam lr $10^{-4}$), and OGBench uses a 4-layer width-512 MLP ($10^6$ updates, batch 256, Adam lr $3\times10^{-4}$), denoted \emph{2-layer Flow BC} and \emph{4-layer Flow BC} when comparing capacities in \cref{app:zs-flowbcplus}.

\paragraph{Guidance modes.} At each replanning step, \textbf{BFN16} scores 16 independent Flow BC draws, \textbf{QGF} guides a single draw via the 10-step DCT-projected update in \cref{app:qgf-impl}, \textbf{QGF+BFN4} selects among 4 guided draws and 1 unguided Flow BC draw supplying the reference draw of \cref{prop:headroom}, and \textbf{QGF+BFN4$^\circ$} denotes the guided-only ablation without the unguided draw. All variants within a mode share reset-seeded proposal noise streams.

\paragraph{Hyperparameter granularity and support sensitivity.} Guidance strengths $\alpha\in\{0,0.03,0.1,0.3,1\}$ (and gripper masking on manipulation) are selected per task on seed-3917 development episodes under $H_{\mathrm{exec}}{=}8$ and transferred unchanged to all seeds and to $H_{\mathrm{exec}}{=}1$. Coarser $\alpha$ granularity yields similar DMC QGF+BFN4 rolling returns on the test split: a single suite-wide $\alpha{=}0.3$ reaches $645.0\pm6.7$ and per-domain $\alpha$ (matching TD-JEPA's tuning granularity: \texttt{walker} 0.3, \texttt{cheetah} 1.0, \texttt{quadruped} and \texttt{pointmass} 0.03) reaches $646.1\pm3.4$, versus $655.2\pm4.2$ per task (and $643.3$, $643.4$, and $652.8$ on held-out seeds). Consistent with \cref{prop:nonid}, seed-3917 development sweeps show that on narrow OGBench data success peaks at $\alpha{=}0.03$ (7.8\% under QGF and 12.0\% under QGF+BFN4) and drops to 0.0\% at $\alpha{=}1$, whereas broad exploratory DMC data sustain high return even at $\alpha{=}1$ (\texttt{quadruped} under QGF moves from 703.3 at $\alpha{=}0.1$ to 624.2 at $\alpha{=}1$).

\paragraph{Held-out seed and split verification.} Retraining Flow BC, critics, density heads, and readouts on three fresh training seeds (4271, 5381, 6491) and evaluating on disjoint initial states (reset seeds $70000,\dots,70019$, 20 episodes per task) confirms the suite configuration without reselection: with 2-layer Flow BC, the suite configuration gains $+54.3$ over the default critic on DMC under QGF+BFN4$^\circ$ on every seed and $+7.4$ ($[+4.8,+10.0]$, 95\% paired $t$-interval) on OGBench under BFN16, and gains $+4.0$ ($[+2.7,+5.3]$) on OGBench over per-dataset development selection while retaining 98\% of DMC return. Evaluating the full suite configuration (with 4-layer Flow BC on OGBench and QGF+BFN4) on these held-out seeds and states confirms consistent improvements across both execution modes in \cref{tab:zs-heldout}.

\begin{table}[!htbp]
\centering
\caption{\textbf{On held-out training seeds and fresh initial states, the frozen suite configuration improves Flow BC across every mode under both rolling and whole-chunk execution.} Evaluation on fresh training seeds (4271, 5381, 6491) and disjoint reset seeds ($70000,\dots,70019$) with unchanged hyperparameters, reported as mean\,$\pm$\,s.e.\ across the 3 per-seed suite means.}
\label{tab:zs-heldout}
\small
\setlength{\tabcolsep}{6pt}
\begin{tabular}{@{}llcrrrr@{}}
\toprule
Suite & Mode & $H_{\rm exec}$ & Mean $\pm$ s.e. & Seed 4271 & Seed 5381 & Seed 6491 \\
\midrule
DMC & Unguided & 1 & 71.1{\scriptsize$\,\pm\,$3.1} & 65.0 & 75.5 & 72.7 \\
DMC & Unguided & 8 & 72.4{\scriptsize$\,\pm\,$2.0} & 73.9 & 68.5 & 74.7 \\
DMC & BFN16 & 1 & 578.3{\scriptsize$\,\pm\,$8.8} & 593.5 & 578.3 & 563.2 \\
DMC & BFN16 & 8 & 551.0{\scriptsize$\,\pm\,$9.2} & 569.1 & 544.5 & 539.3 \\
DMC & QGF & 1 & 627.7{\scriptsize$\,\pm\,$4.5} & 628.1 & 635.3 & 619.6 \\
DMC & QGF & 8 & 619.5{\scriptsize$\,\pm\,$5.5} & 608.9 & 627.3 & 622.3 \\
DMC & QGF+BFN4 & 1 & 652.8{\scriptsize$\,\pm\,$6.0} & 662.2 & 654.4 & 641.7 \\
DMC & QGF+BFN4 & 8 & 636.6{\scriptsize$\,\pm\,$3.0} & 638.7 & 640.4 & 630.7 \\
\midrule
OGBench & Unguided & 1 & 2.5{\scriptsize$\,\pm\,$0.4} & 3.2 & 2.0 & 2.3 \\
OGBench & Unguided & 8 & 2.2{\scriptsize$\,\pm\,$0.2} & 2.2 & 2.6 & 1.9 \\
OGBench & BFN16 & 1 & 37.0{\scriptsize$\,\pm\,$1.5} & 39.0 & 38.1 & 34.0 \\
OGBench & BFN16 & 8 & 31.0{\scriptsize$\,\pm\,$0.8} & 30.2 & 30.2 & 32.6 \\
OGBench & QGF & 1 & 23.7{\scriptsize$\,\pm\,$0.6} & 22.6 & 24.8 & 23.8 \\
OGBench & QGF & 8 & 21.2{\scriptsize$\,\pm\,$0.1} & 21.2 & 21.4 & 21.0 \\
OGBench & QGF+BFN4 & 1 & 33.1{\scriptsize$\,\pm\,$0.8} & 32.2 & 34.7 & 32.6 \\
OGBench & QGF+BFN4 & 8 & 25.6{\scriptsize$\,\pm\,$0.2} & 25.7 & 25.2 & 26.0 \\
\bottomrule
\end{tabular}

\end{table}

\FloatBarrier
\subsection{Pixel inputs and object-level sensitivity}
\label{app:zs-pixels}
\label{app:zs-pixel-focus}

DMC-RGB renders the ExORL trajectories of \cref{app:zs-protocol} at $64\times64$ (camera 2 on \texttt{quadruped}, camera 0 otherwise), and visual OGBench uses the official $64\times64$ RGB versions of all nine datasets, stacked over 3 frames without proprioception. Both $\phi$ and $\psi$ use end-to-end DrQ encoders from \cref{app:arch} under the same $62.5\text{k}$-update budget and horizons as states, paired with linear readouts $\hJ_\psi$ (ridge or goal vector) for default and within-chunk targets as in \cref{app:within-chunk}. Visual Flow BC trains its own end-to-end DrQ encoder with the 2-layer head on DMC-RGB and 4-layer head on visual OGBench. For each dataset, the (target, readout) pair with the highest seed-3917 development mean ($30000,\dots,30009$, $H_{\mathrm{exec}}{=}8$, BFN16) is selected and evaluated across 3 seeds on test episodes ($20000,\dots,20009$) under whole-chunk execution in \cref{tab:zeroshot-pixels-full} and under rolling execution in \cref{tab:zeroshot-pixels}, reusing BFN16 on visual OGBench and QGF+BFN4 with per-task development $\alpha$ on DMC-RGB. Upgrading visual Flow BC from 2 layers to 4 layers on visual OGBench under the identical frozen critics raises $H_{\mathrm{exec}}{=}8$ test success from 21.1\% to 31.6\% across every seed, and development selection picks the default successor target on 9 of 13 datasets in \cref{tab:zeroshot-pixels-full}.

\begin{table}[!htbp]
\centering
\caption{\textbf{From raw pixels, all four target--readout pairs improve frozen visual Flow BC on both suites.} Test-episode performance across 3 training seeds, reported as mean\,$\pm$\,s.e.\ across per-seed suite means. All four pairs use BFN16 under whole-chunk execution ($H_{\mathrm{exec}}{=}8$). The rolling Selected column ($H_{\mathrm{exec}}{=}1$) reuses the guidance of \cref{tab:zeroshot-pixels}, which is QGF+BFN4 on DMC-RGB and BFN16 on visual OGBench. Underlined entries mark the development-selected pair repeated under Selected, and the bottom row reports the 2-layer visual Flow BC ablation.}
\label{tab:zeroshot-pixels-full}
\small
\begingroup
\small
\setlength{\tabcolsep}{3pt}
\begin{tabular}{@{}l cc cccc cc@{}}
\toprule
 & \multicolumn{2}{c}{Base, $H_{\mathrm{exec}}$} & \multicolumn{4}{c}{All pairs, $H_{\mathrm{exec}}{=}8$} & \multicolumn{2}{c}{Selected, $H_{\mathrm{exec}}$} \\
\cmidrule(lr){2-3}\cmidrule(lr){4-7}\cmidrule(l){8-9}
 &  &  & \multicolumn{2}{c}{default} & \multicolumn{2}{c}{within-chunk target} &  &  \\
\cmidrule(lr){4-5}\cmidrule(lr){6-7}
Dataset & 8 & 1 & ridge & goal & ridge & goal & 8 & 1 \\
\midrule
\multicolumn{9}{@{}l}{\emph{DMC-RGB, base policy Flow BC}} \\
walker & 163.3 & 159.2 & \underline{293.0} & 286.1 & 222.8 & 207.3 & 293.0 & 357.1 \\
cheetah & 113.3 & 115.7 & \underline{523.4} & 533.9 & 279.5 & 281.4 & 523.4 & 663.2 \\
quadruped & 103.4 & 212.7 & 217.1 & \underline{187.0} & 119.8 & 100.4 & 187.0 & 242.5 \\
pointmass & $-$60.7 & $-$71.9 & 125.2 & 91.3 & \underline{79.5} & 27.5 & 79.5 & 130.8 \\
\emph{DMC-RGB (avg)} & 79.8\,$\pm$\,3.4 & 103.9\,$\pm$\,5.7 & 289.7\,$\pm$\,38.1 & 274.6\,$\pm$\,22.3 & 175.4\,$\pm$\,9.1 & 154.2\,$\pm$\,5.1 & 270.7\,$\pm$\,10.2 & 348.4\,$\pm$\,8.9 \\
\midrule
\multicolumn{9}{@{}l}{\emph{OGBench visual, base policy Flow BC (4-layer)}} \\
antmaze-mn & 7.3 & 20.7 & \underline{74.7} & 74.0 & 14.7 & 13.3 & 74.7 & 82.0 \\
antmaze-ln & 3.3 & 8.7 & \underline{32.7} & 24.0 & 1.3 & 0.7 & 32.7 & 44.0 \\
antmaze-ms & 4.7 & 4.7 & \underline{27.3} & 22.0 & 2.7 & 4.7 & 27.3 & 16.7 \\
antmaze-ls & 0.0 & 0.0 & 0.0 & 1.3 & \underline{0.0} & 0.0 & 0.0 & 0.7 \\
antmaze-me & 0.0 & 0.0 & 0.0 & \underline{0.0} & 0.0 & 0.0 & 0.0 & 0.0 \\
cube-single & 4.7 & 1.3 & \underline{93.3} & 96.7 & 17.3 & 24.0 & 93.3 & 82.7 \\
cube-double & 0.0 & 1.3 & 5.3 & \underline{13.3} & 8.0 & 8.0 & 13.3 & 15.3 \\
scene & 6.0 & 6.0 & 37.3 & 39.3 & \underline{37.3} & 28.7 & 37.3 & 38.7 \\
puzzle-3x3 & 2.0 & 4.7 & 4.7 & 3.3 & \underline{6.0} & 6.0 & 6.0 & 11.3 \\
\emph{OGBench visual (avg)} & 3.1\,$\pm$\,0.3 & 5.3\,$\pm$\,0.3 & 30.6\,$\pm$\,0.9 & 30.4\,$\pm$\,1.4 & 9.7\,$\pm$\,1.0 & 9.5\,$\pm$\,0.9 & 31.6\,$\pm$\,0.7 & 32.4\,$\pm$\,0.8 \\
\midrule
\multicolumn{9}{@{}l}{\emph{OGBench visual, 2-layer Flow BC ablation ($H_{\mathrm{exec}}{=}1$ not evaluated)}} \\
\emph{OGBench visual (avg)} & 2.5\,$\pm$\,0.3 & -- & 19.6\,$\pm$\,0.6 & 19.0\,$\pm$\,0.9 & 5.8\,$\pm$\,0.1 & 6.7\,$\pm$\,0.4 & 21.1\,$\pm$\,0.8 & -- \\
\bottomrule
\end{tabular}
\endgroup

\end{table}

To test whether visual pretraining grounds task scores in task-relevant objects in \cref{fig:pixel-focus,tab:zs-pixel-focus}, we probe frozen pixel critics on 64 recorded visual \texttt{scene} frames (8 frames from each of 8 random episodes across all 3 seeds) without environment rollouts. Defining indicator rewards for \emph{open drawer}, \emph{open window}, and \emph{lift cube} on the 10k task-inference states via the ridge vector $z$, we occlude $5\times5$ Gaussian patches at stride 2 (blending a blurred frame with the dataset median frame) and compute the fraction of squared score change $(\Delta\hJ_\psi)^2$ falling inside each object's simulator segmentation mask, while the task-free column in \cref{fig:pixel-focus} records Euclidean norm change $\norm{\Delta T_\phi}_2$. With the reported within-chunk target, every object receives a larger sensitivity share under its own task than under either other task on every seed in \cref{tab:zs-pixel-focus}, including on the 44 frames without cube--drawer overlap (27\% vs.\ 16\% and 14\%), and the default successor target preserves the same ordering on every seed. At initialization, no object's share under its own task exceeds its largest share under another task by more than 1 percentage point.

\begin{table}[!htbp]
\centering
\caption{\textbf{After reward-free pretraining, each object captures a larger share of score sensitivity under its own task than under other tasks.} Occlusion-sensitivity share (\%, mean\,$\pm$\,s.e.\ over 3 seeds across 64 visual \texttt{scene} frames) inside each simulator object mask for $\hJ_\psi$. Bold marks each task's target object, and Area gives mean pixel coverage (\%).}
\label{tab:zs-pixel-focus}
\small
\begin{tabular}{@{}l ccc@{}}
\toprule
Task & Drawer & Window & Cube \\
\midrule
\multicolumn{4}{@{}l}{\emph{Within-chunk target (reported on scene)}} \\
open drawer & $\mathbf{37.4}\pm3.9$ & $4.1\pm0.4$ & $23.9\pm4.8$ \\
open window & $23.6\pm0.9$ & $\mathbf{13.0}\pm1.4$ & $19.2\pm1.9$ \\
lift cube & $26.4\pm4.7$ & $4.7\pm0.8$ & $\mathbf{42.7}\pm7.3$ \\
\midrule
\multicolumn{4}{@{}l}{\emph{Default successor target}} \\
open drawer & $\mathbf{38.8}\pm1.4$ & $7.1\pm0.6$ & $23.6\pm2.4$ \\
open window & $31.9\pm1.1$ & $\mathbf{10.0}\pm1.8$ & $23.2\pm2.5$ \\
lift cube & $31.5\pm1.3$ & $4.6\pm0.4$ & $\mathbf{39.6}\pm3.2$ \\
\midrule
\multicolumn{4}{@{}l}{\emph{Same network at initialization}} \\
open drawer & $22.1\pm1.8$ & $7.6\pm0.2$ & $2.3\pm0.1$ \\
open window & $24.0\pm2.1$ & $7.7\pm0.2$ & $2.5\pm0.2$ \\
lift cube & $30.0\pm1.3$ & $7.1\pm0.2$ & $2.9\pm0.1$ \\
\midrule
Area & 5.9 & 5.7 & 0.6 \\
\bottomrule
\end{tabular}
\end{table}

\FloatBarrier
\subsection{Critic swap under a shared Flow BC proposer}
\label{app:zs-swap}

To isolate critic quality on identical proposals as in \cref{sec:exp-swap}, we retrain TD-JEPA with its unmodified official code and launcher budgets ($3\times10^6$ updates at batch 1024 on DMC and $10^6$ updates at batch 256 on OGBench), sweeping its full per-dataset grid ($\mathrm{lr}_\psi\in\{10^{-4},10^{-5}\}$ crossed with its released $\lambda_\psi$ values) on seed 3917 and retraining the best configuration per dataset on seeds 3502 and 8948. In \cref{tab:zeroshot-swap}, \method{} leads in all four cells on every seed, and replicating the BFN16 rolling comparison on an independent block of initial states ($90000,\dots,90019$) confirms the same ordering on every seed (569.8 vs.\ 558.6 on DMC and 38.6\% vs.\ 15.5\% on OGBench). One-step FB's critic $F(s,a)^\top z$ comes from the agents in \cref{app:zs-baselines}. Both single-action baseline critics score each proposed chunk by its first action $a_0$ using their native 10k-transition task inference on shared noise streams.

Under rolling gradient guidance ($H_{\mathrm{exec}}{=}1$) with per-task development step sizes selected over $\alpha\in\{0,0.03,0.1,0.3,1,3\}$, the reproduced TD-JEPA critic reaches 595.0\,$\pm$\,7.0 (QGF) and 598.4\,$\pm$\,12.3 (QGF+BFN4) on DMC versus 629.5 and 655.2 for \method{} (which transfers $\alpha\le1$ from $H_{\mathrm{exec}}{=}8$), and one-step FB's critic reaches 205.9\,$\pm$\,4.1 and 200.2\,$\pm$\,10.8 on DMC and 8.9\,$\pm$\,0.2\% and 15.0\,$\pm$\,1.1\% on OGBench, with \method{} ahead on every seed across all comparisons. Under whole-chunk execution ($H_{\mathrm{exec}}{=}8$) in \cref{tab:zeroshot-swap-full}, \method{} likewise leads both baseline critics on every seed across all three guidance modes on DMC and on OGBench with 4-layer Flow BC, and leads the reproduced TD-JEPA critic on every seed in every mode with 2-layer Flow BC. Moving from $H_{\mathrm{exec}}{=}8$ to per-step replanning ($H_{\mathrm{exec}}{=}1$) in \cref{tab:zeroshot-swap} raises every critic on every seed while leaving unguided Flow BC nearly unchanged.

\Cref{tab:zeroshot-nsweep,fig:supp-nsweep} sweep the candidate count $N\in\{1,2,4,8,16,32\}$ at $H_{\mathrm{exec}}{=}8$. \Cref{tab:zeroshot-nsweep} reports both the per-dataset-swept reproduced TD-JEPA critic and our fixed-budget TD-JEPA retrain (trained under the same launcher budgets without per-dataset hyperparameter search), while \cref{fig:supp-nsweep} expands the 5-panel summary of \cref{fig:n-scaling} to 10 panels covering all four DMC domains and representative OGBench navigation, stitching, exploration, and manipulation datasets. In \cref{fig:supp-nsweep}, \method{} improves monotonically with $N$ on every DMC panel, the OGBench mean, \texttt{antmaze-ms}, and \texttt{cube-single}, and levels off from $N{=}4$ on \texttt{antmaze-mn} and from $N{=}8$ on \texttt{antmaze-me}, whereas the single-step TD-JEPA critic gains little on \texttt{quadruped} and \texttt{antmaze-me}, where evaluating only the first action fails to distinguish multi-step action chunks. Evaluated with its own trained actors on the test split of \cref{tab:zeroshot}, reproduced TD-JEPA reaches 638.9 on DMC and 32.0\% on OGBench (versus 661.2 and 38.0\% reported over larger seed pools by \citet{bagatella2026td}), which \method{}-guided Flow BC exceeds on every seed in both suites, via BFN16 on OGBench and QGF+BFN4 on DMC. \Cref{app:zs-baseline-density} further compares frozen \method{}, reproduced TD-JEPA, and one-step FB representations on OGBench under the shared density readout of \cite{anonymous2026density}.

\begin{figure}[!htbp]
\centering
\includegraphics[width=\textwidth]{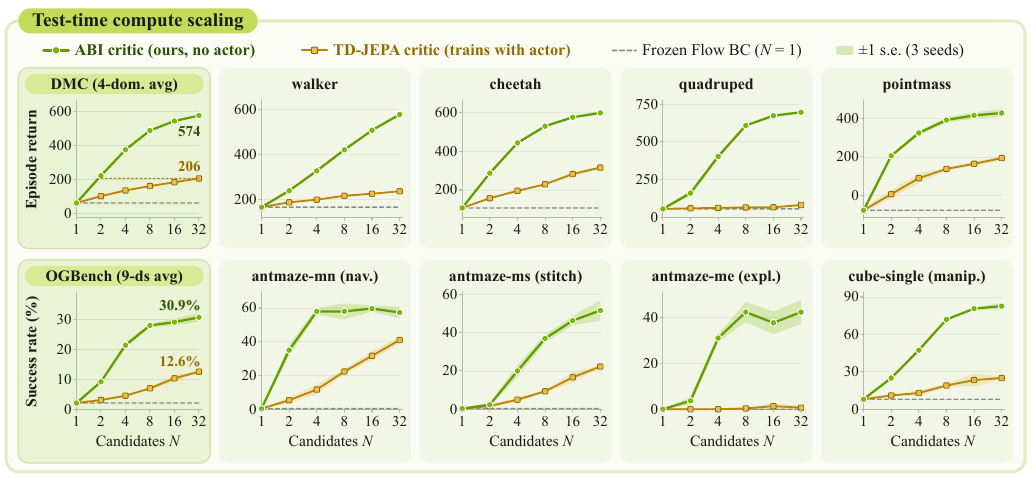}
\caption{\textbf{Expanding \cref{fig:n-scaling} to 10 suite and domain panels confirms that test-time compute scaling under whole-chunk execution ($H_{\mathrm{exec}}{=}8$) requires a chunk-level critic across locomotion, navigation, stitching, exploration, and manipulation.} Best-of-$N$ selection ($N\in\{1,2,4,8,16,32\}$) over shared frozen Flow BC proposals on the test split of \cref{tab:zeroshot} (mean\,$\pm$\,1 s.e.\ across 3 training seeds). TD-JEPA curves show the fixed-budget retrain, and \cref{tab:zeroshot-nsweep} lists the suite values.}
\label{fig:supp-nsweep}
\end{figure}

\begin{table}[!htbp]
\centering
\caption{\textbf{At every candidate budget $N\ge 2$, the \method{} critic outperforms both TD-JEPA critics on every training seed in both suites.} Whole-chunk best-of-$N$ selection ($H_{\mathrm{exec}}{=}8$) over frozen Flow BC proposals on the test split of \cref{tab:zeroshot}, reporting DMC return and OGBench success (\%) as mean\,$\pm$\,s.e.\ across 3 per-seed suite means. $N{=}1$ is unguided Flow BC. \Cref{fig:n-scaling} plots the reproduced TD-JEPA curve, and \cref{fig:supp-nsweep} plots the fixed-budget curve on per-domain panels.}
\label{tab:zeroshot-nsweep}
\small
\setlength{\tabcolsep}{3pt}
\begin{tabular}{@{}l l c c c c c c@{}}
\toprule
Suite & Critic & $N{=}1$ & $N{=}2$ & $N{=}4$ & $N{=}8$ & $N{=}16$ & $N{=}32$ \\
\midrule
DMC & Reproduced TD-JEPA & 63.3{\scriptsize$\,\pm\,$3.7} & 104.9{\scriptsize$\,\pm\,$3.0} & 140.4{\scriptsize$\,\pm\,$6.2} & 162.8{\scriptsize$\,\pm\,$6.8} & 191.8{\scriptsize$\,\pm\,$9.1} & 209.6{\scriptsize$\,\pm\,$5.4} \\
 & Fixed-budget TD-JEPA & 63.3{\scriptsize$\,\pm\,$3.7} & 102.8{\scriptsize$\,\pm\,$4.7} & 136.4{\scriptsize$\,\pm\,$6.4} & 162.0{\scriptsize$\,\pm\,$1.7} & 184.1{\scriptsize$\,\pm\,$6.2} & 206.4{\scriptsize$\,\pm\,$3.6} \\
 & \method{} & 63.3{\scriptsize$\,\pm\,$3.7} & 222.4{\scriptsize$\,\pm\,$5.2} & 373.6{\scriptsize$\,\pm\,$6.3} & 486.2{\scriptsize$\,\pm\,$3.3} & 541.9{\scriptsize$\,\pm\,$3.6} & 573.7{\scriptsize$\,\pm\,$3.7} \\
\midrule
OGBench & Reproduced TD-JEPA & 2.1{\scriptsize$\,\pm\,$0.5} & 3.7{\scriptsize$\,\pm\,$0.1} & 6.3{\scriptsize$\,\pm\,$0.3} & 7.2{\scriptsize$\,\pm\,$0.7} & 11.7{\scriptsize$\,\pm\,$0.1} & 13.5{\scriptsize$\,\pm\,$1.3} \\
 & Fixed-budget TD-JEPA & 2.1{\scriptsize$\,\pm\,$0.5} & 3.1{\scriptsize$\,\pm\,$0.4} & 4.5{\scriptsize$\,\pm\,$0.3} & 7.0{\scriptsize$\,\pm\,$0.4} & 10.4{\scriptsize$\,\pm\,$1.1} & 12.6{\scriptsize$\,\pm\,$0.4} \\
 & \method{} & 2.1{\scriptsize$\,\pm\,$0.5} & 9.2{\scriptsize$\,\pm\,$0.3} & 21.5{\scriptsize$\,\pm\,$0.9} & 28.1{\scriptsize$\,\pm\,$0.6} & 29.2{\scriptsize$\,\pm\,$1.0} & 30.9{\scriptsize$\,\pm\,$1.5} \\
\bottomrule
\end{tabular}
\end{table}

\begin{table}[!htbp]
\centering
\caption{\textbf{Under whole-chunk execution ($H_{\mathrm{exec}}{=}8$), the \method{} critic outperforms reproduced TD-JEPA and one-step FB critics across every guidance mode on both suites.} Suite averages across 3 training seeds on the test split using 2-layer Flow BC (top two blocks) and 4-layer Flow BC (bottom block). \method{} (default) pairs the default backbone with the density readout on all datasets, \method{} uses the suite configuration of \cref{tab:zeroshot}, and QGF+BFN4$^\circ$ uses four guided draws.}
\label{tab:zeroshot-swap-full}
\small
\setlength{\tabcolsep}{4pt}
\begin{tabular}{@{}l l c c c@{}}
\toprule
Suite & Critic & BFN16 & QGF & QGF+BFN4$^\circ$ \\
\midrule
DMC (avg) & TD-JEPA & 191.8 & 240.0 & 275.8 \\
 & One-step FB & 143.5 & 145.8 & 152.6 \\
 & \method{} (default) & 515.8 & 571.0 & 587.9 \\
 & \method{} & \textbf{541.9} & \textbf{615.6} & \textbf{639.8} \\
\midrule
OGBench (avg), & TD-JEPA & 7.3 & 6.5 & 8.7 \\
2-layer Flow BC & \method{} (default) & 16.1 & 8.0 & 10.5 \\
 & \method{} & \textbf{24.8} & \textbf{15.0} & \textbf{18.4} \\
\midrule
OGBench (avg), & TD-JEPA & 11.7 & 8.6 & 11.1 \\
4-layer Flow BC & One-step FB & 7.6 & 3.1 & 5.7 \\
 & \method{} & \textbf{29.2} & \textbf{22.2} & \textbf{23.7} \\
\bottomrule
\end{tabular}
\end{table}

\subsection{Counterfactual ranking of fixed candidates}
\label{app:zs-rank}

Complementing closed-loop rollouts, this counterfactual rollout diagnostic directly measures how accurately each critic of \cref{app:zs-swap} ranks a shared set of candidate chunks on true environment returns as analyzed in \cref{prop:ranking,sec:exp-rank}. Across \texttt{walker}, \texttt{antmaze-mn}, and \texttt{scene}, we sample 48 root states per domain (8 from unguided Flow BC and 8 from \method{}-guided rollouts per training seed, reset seeds $120000,\dots,120008$). At each root, 8 Flow BC chunks of length 8 are frozen and executed either in full (whole) or for their first action only (first), each across two independent rolling Flow BC continuations to episode end (4{,}608 branches total). A critic's gain is the discounted task return of its top-scored chunk minus the 8-chunk mean, and the estimated maximum gain $G$ cross-fits chunk selection and evaluation across the two continuation replicates, with 95\% stratified bootstrap intervals over 10{,}000 root resamples.

In \cref{fig:rank,tab:zs-rank}, on dense-reward \texttt{walker} under whole-chunk execution, \method{} achieves positive gain in all six seed $\times$ state-source cells (0.41 to 2.55), exceeding reproduced TD-JEPA's critic by 1.02 [0.03, 2.13] and one-step FB's critic by 1.62 [0.54, 2.76] with return-gap-weighted pairwise accuracy 0.68 versus 0.55 and 0.49. \Cref{tab:zs-rank} reports all three domains under both execution modes: on sparse-goal \texttt{antmaze-mn}, single 8-step branches yield identical terminal returns at 30 of 48 roots under whole-chunk execution and 32 of 48 roots under first-action execution.

\begin{table}[!htbp]
\centering
\caption{\textbf{On dense-reward \texttt{walker} with whole chunks, the \method{} critic recovers most of the cross-fitted maximum candidate gain.} Discounted return gain over a random pick among 8 frozen Flow BC chunks under whole-chunk (whole) and first-action (first) execution followed by rolling Flow BC continuation, reported as mean and 95\% stratified bootstrap interval across 48 roots over 3 training seeds. Bold marks the critic gain whose 95\% interval lies above zero.}
\label{tab:zs-rank}
\small
\setlength{\tabcolsep}{4pt}
\begin{tabular}{@{}l l c c c c@{}}
\toprule
Domain & Execution & Est.\ max gain & \method{} critic & TD-JEPA's critic & One-step FB's critic \\
\midrule
walker & whole & 1.87 [1.03, 2.81] & \textbf{1.63} [0.85, 2.48] & 0.60 [$-0.16$, 1.43] & 0.00 [$-0.50$, 0.55] \\
 & first & $-0.02$ [$-0.56$, 0.56] & 0.23 [$-0.19$, 0.65] & 0.04 [$-0.45$, 0.60] & $-0.01$ [$-0.55$, 0.52] \\
\midrule
antmaze-mn ($\times10^{-3}$) & whole & 1.2 [$-2.2$, 5.4] & 3.6 [$-1.9$, 13.5] & $-1.5$ [$-3.4$, $-0.0$] & 5.0 [$-1.3$, 15.2] \\
 & first & $-1.4$ [$-2.3$, $-0.6$] & 0.6 [$-1.7$, 4.3] & $-0.3$ [$-2.0$, 1.9] & 0.5 [$-1.5$, 3.5] \\
\midrule
scene & whole & $-0.48$ [$-7.54$, 7.13] & 0.86 [$-5.19$, 7.52] & 3.60 [$-3.10$, 10.47] & $-8.26$ [$-14.96$, $-1.85$] \\
 & first & 4.15 [$-1.26$, 9.83] & 0.14 [$-6.09$, 6.87] & 3.59 [$-2.73$, 10.45] & $-0.57$ [$-5.53$, 4.28] \\
\bottomrule
\end{tabular}
\end{table}

\FloatBarrier
\subsection{Guiding actors trained by other methods}
\label{app:zs-steer-actors}

Without retraining or hyperparameter reselection, the frozen \method{} critic of \cref{tab:zeroshot} guides the single-step actors of reproduced TD-JEPA from \cref{app:zs-swap} and one-step FB from \cref{app:zs-baselines} at test time in \cref{tab:zeroshot-actors}. At every step, 16 action candidates are drawn: the actor's deterministic mode plus 15 stochastic samples (from TD-JEPA's actor distribution or Gaussian noise of scale $0.2$ around one-step FB's mode). Each candidate $a_i$ is completed into $K{=}4$ 8-step chunks via shared-noise Flow BC inpainting as formalized in \cref{app:completion-readout} (with $K$ fixed on development episodes from score rank stability), scored by $\hJ$, and executed for one step ($H_{\mathrm{exec}}{=}1$).

In \cref{tab:zs-steer-released}, guiding reproduced TD-JEPA actors on both the primary test split ($40000,\dots,40019$) and fresh initial states ($80000,\dots,80019$) beats both the unguided actor and a random candidate pick on every seed on OGBench (exceeding the 38.0\% reported by \citet{bagatella2026td}) and beats the random control on every seed in both DMC blocks, with pooled paired gains of $+9.8$ [7.3, 12.3] on DMC and $+8.6$ [7.4, 9.8] points on OGBench over the unguided actor ($+16.1$ [13.4, 18.8] and $+7.6$ [6.4, 8.8] over random selection, 95\% bootstrap intervals). Similarly, guiding one-step FB actors on the test split outperforms the unguided actor and random selection in \cref{tab:zeroshot-actors}, as well as one-step FB's own critic, on every seed in both suites, selecting a non-mode sample on 88--100\% of steps.

\begin{table}[!htbp]
\centering
\caption{\textbf{The frozen \method{} critic improves reproduced TD-JEPA actors over both unguided execution and random candidate selection across both evaluation blocks.} Rolling single-step execution ($N{=}16$ actor proposals scored via $K{=}4$ shared-noise Flow BC completions) across 3 training seeds on the primary test split ($40000,\dots,40019$) and fresh initial states ($80000,\dots,80019$). Unguided rows rerun each actor in this evaluation and match the separate actor evaluation of \cref{tab:zs-convergence} on DMC, whereas AntMaze, scene, and puzzle rollouts differ between the two runs (OGBench 30.3\% here and 32.0\% there).}
\label{tab:zs-steer-released}
\small
\begin{tabular}{@{}l l c c c r l r l@{}}
\toprule
 & & & & & \multicolumn{2}{c}{\method{} $-$ unguided} & \multicolumn{2}{c}{\method{} $-$ random} \\
\cmidrule(lr){6-7}\cmidrule(l){8-9}
Suite & Block & Unguided & Random & \method{} & Mean & Per seed & Mean & Per seed \\
\midrule
DMC & test & 638.9 & 630.8 & \textbf{646.8} & $+7.9$ & $+14.3$, $+12.8$, $-3.2$ & $+16.0$ & $+13.7$, $+24.2$, $+10.2$ \\
 & new & 636.2 & 631.6 & \textbf{647.8} & $+11.6$ & $+16.6$, $+15.9$, $+2.5$ & $+16.2$ & $+17.6$, $+20.3$, $+10.7$ \\
\midrule
OGBench & test & 30.3 & 31.7 & \textbf{39.7} & $+9.4$ & $+8.8$, $+10.6$, $+8.9$ & $+8.0$ & $+6.6$, $+9.2$, $+8.2$ \\
 & new & 31.3 & 31.9 & \textbf{39.1} & $+7.8$ & $+6.9$, $+9.6$, $+6.9$ & $+7.2$ & $+7.9$, $+10.2$, $+3.4$ \\
\bottomrule
\end{tabular}
\end{table}

\FloatBarrier
\subsection{Training chunk lengths and single-step critics}
\label{app:zs-lengths}

Training one shared multi-horizon predictor across $\mathcal H=\{4,8,16,32\}$ (queried at $H{=}8$) outperforms fixed $H{=}8$ training across all three guidance modes on DMC and substantially raises long-horizon maze navigation success on \texttt{antmaze-mn} and \texttt{antmaze-ms} in \cref{tab:zeroshot-variants}. Single-step training with $H_{\mathrm{score}}{=}H_{\mathrm{exec}}{=}1$, where the tail term of \cref{prop:r7-rolling} vanishes, already lifts DMC Flow BC from 58.4\,$\pm$\,3.5 to 497.9\,$\pm$\,9.2 with $\hJ_\psi$ and to 553.3\,$\pm$\,2.5 with $\hJ_D$ when scoring the first action of 16 Flow BC proposals on the test split. Multi-horizon chunk training reaches 572.5 in \cref{tab:zeroshot-swap} and outperforms $H{=}1$ training on every seed in the DMC suite mean.

\FloatBarrier
\subsection{Proposer capacity: 2-layer versus 4-layer Flow BC}
\label{app:zs-flowbcplus}

To evaluate zero-retraining critic reuse across proposer capacities, we pair the identical frozen \method{} critics and guidance settings with the 2-layer ($2\times256$) and 4-layer ($4\times512$) Flow BC proposers of \cref{app:zs-guidance}. At $H_{\mathrm{exec}}{=}8$ in \cref{tab:zs-flowbcplus}, upgrading to 4-layer Flow BC substantially raises multi-modal OGBench success, most notably on \texttt{scene} and \texttt{cube-single}, and independent confirmation on held-out training seeds (4271, 5381, 6491) and fresh states ($70000,\dots,70019$) shows 4-layer Flow BC winning on every seed under both BFN16 and QGF+BFN4, whereas on exploratory DMC data 2-layer Flow BC already matches 4-layer Flow BC under BFN16 and achieves higher QGF+BFN4 return. A paired rolling-execution run ($H_{\mathrm{exec}}{=}1$, BFN16) on OGBench confirms the gain on every seed: \method{}-guided success rises from 32.3\,$\pm$\,1.1\% with 2-layer Flow BC (unguided 1.4\,$\pm$\,0.2\%) to 38.6\,$\pm$\,0.6\% with 4-layer Flow BC in \cref{tab:zeroshot}, and in the paired run 4-layer Flow BC is higher on every seed.

\begin{table}[!htbp]
\centering
\caption{\textbf{Without retraining the critic, upgrading from 2-layer to 4-layer Flow BC raises OGBench success across primary and held-out seeds.} Whole-chunk execution ($H_{\mathrm{exec}}{=}8$) suite means and paired differences (4-layer minus 2-layer, with 95\% Student-$t$ intervals and per-seed differences) on the primary test split and on held-out training seeds and initial states. Both columns come from one paired rerun, whose 2-layer OGBench BFN16 mean of 24.7 differs by 0.1 from \cref{tab:zeroshot-swap-full}.}
\label{tab:zs-flowbcplus}
\small
\setlength{\tabcolsep}{4pt}
\begin{tabular}{@{}l l c c c l@{}}
\toprule
Suite & Mode & 2-layer & 4-layer & Difference & Per seed \\
\midrule
DMC & BFN16 & 541.9 & 541.0 & $-0.9$ $[-22.8,+21.0]$ & $-2.9$, $+8.8$, $-8.5$ \\
 & QGF+BFN4 & 644.3 & 615.5 & $-28.8$ $[-42.5,-15.2]$ & $-29.2$, $-23.2$, $-34.1$ \\
\midrule
OGBench & BFN16 & 24.7 & 29.2 & $+4.6$ $[-1.9,+11.0]$ & $+3.8$, $+7.4$, $+2.4$ \\
 & QGF+BFN4 & 18.3 & 25.5 & $+7.2$ $[+0.7,+13.7]$ & $+9.2$, $+8.1$, $+4.2$ \\
\midrule
OGBench, & BFN16 & 23.7 & 31.0 & $+7.3$ $[+5.2,+9.4]$ & $+6.6$, $+7.2$, $+8.2$ \\
new seeds & QGF+BFN4 & 17.7 & 25.6 & $+7.9$ $[+5.7,+10.1]$ & $+8.0$, $+7.0$, $+8.8$ \\
\bottomrule
\end{tabular}
\end{table}

\FloatBarrier
\subsection{Baseline reproduction}
\label{app:zs-baselines}

One-step FB \citep{zheng2026can} in \cref{tab:zeroshot} executes its own trained actor across all 3 seeds and all 13 datasets: \texttt{walker}, \texttt{cheetah}, and \texttt{quadruped} use the final checkpoints ($10^6$ updates at batch 1024) from the authors' unmodified training pipeline, while \texttt{pointmass} and OGBench run the same JAX agent code and README hyperparameters ($10^6$ updates at batch 1024, using the nearest listed domain settings where unlisted) through our data loader. Reproduced TD-JEPA follows the official-code training protocol detailed in \cref{app:zs-swap}.

\FloatBarrier
\subsection{Pretraining sample scaling and convergence}
\label{app:zs-convergence}

To evaluate downstream control as a function of reward-free pretraining samples ($\text{updates}\times\text{batch}$), \cref{fig:supp-convergence,tab:zs-convergence} evaluate intermediate checkpoints across all 13 state datasets and 3 seeds (3917, 3502, 8948) on the primary test split ($40000,\dots,40019$) with no snapshot selection. For \method{}, sample counts sum backbone pretraining ($62.5\text{k}\times 2048 = 128.0\text{M}$ at full budget) plus matching-fraction density-head training on OGBench ($256.0\text{M}$ total), evaluated from one independent cosine-schedule run matching \cref{tab:zeroshot} at full budget (572.5 vs.\ 572.5 on DMC and 38.4\% vs.\ 38.6\% on OGBench). Reproduced TD-JEPA ($3072.0\text{M}$ full budget on DMC, $256.0\text{M}$ on OGBench) and one-step FB ($1024.0\text{M}$ on DMC) execute their own trained actors at constant learning rates.

As visualized in \cref{fig:supp-convergence}(a) and detailed in \cref{tab:zs-convergence}, \method{} pretrains with $24\times$ fewer samples than reproduced TD-JEPA's full DMC budget (128.0M vs.\ 3072.0M), outperforms reproduced TD-JEPA at matched 128.0M samples on every seed, and crosses 80\% of TD-JEPA's full-budget DMC return (dotted line) with $12\times$ fewer samples (32.0M vs.\ 384.0M). Even one-step FB's highest intermediate DMC checkpoint (329.2 at 409.6M samples) remains well below \method{} at 16.0M samples (451.3) before degrading later in actor-coupled training. On OGBench in \cref{fig:supp-convergence}(b), \method{} at 256.0M total samples outperforms both reproduced TD-JEPA and one-step FB at matched sample budgets on every seed and crosses 80\% of TD-JEPA's full-budget success at half the samples (64.0M vs.\ 128.0M).

\begin{figure}[!htbp]
\centering
\includegraphics[width=\textwidth]{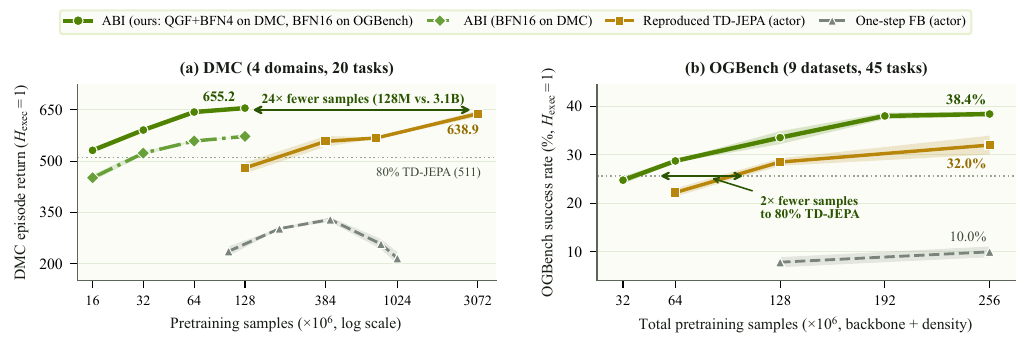}
\caption{\textbf{Eliminating joint actor training yields faster and monotonic sample scaling on both DMC and OGBench.} Downstream rolling test performance ($H_{\mathrm{exec}}{=}1$, mean\,$\pm$\,1 s.e.\ across 3 training seeds) versus total reward-free pretraining samples ($\text{updates}\times\text{batch}$), plotting the checkpoint evaluations of \cref{tab:zs-convergence}. Dotted horizontal lines mark 80\% of reproduced TD-JEPA's full-budget score on each suite.}
\label{fig:supp-convergence}
\end{figure}

\begin{table}[!htbp]
\centering
\caption{\textbf{\method{} reaches 80\% of reproduced TD-JEPA's full-budget score with $12\times$ fewer pretraining samples on DMC and $2\times$ fewer on OGBench under BFN16.} Pretraining sample scaling ($\text{updates}\times\text{batch}$, $\times 10^6$, summing backbone and OGBench density heads for \method{}) across all 13 state benchmarks on the test split ($H_{\mathrm{exec}}{=}1$, reusing full-budget per-task $\alpha$ for DMC QGF+BFN4), reported as mean\,$\pm$\,s.e.\ across 3 training seeds. Bold marks \method{}'s full budget and matched-sample baseline rows.}
\label{tab:zs-convergence}
\footnotesize
\setlength{\tabcolsep}{5pt}
\begin{tabular}{@{}l l r r c l@{}}
\toprule
Suite & Method & Samples (M) & Updates & Suite score & Per seed (3917, 3502, 8948) \\
\midrule
DMC & \method{} (BFN16, $H_{\mathrm{exec}}{=}1$) & 16.0 & 7{,}813 & 451.3{\scriptsize$\,\pm\,$9.1} & 469.2, 445.0, 439.7 \\
(return) &  & 32.0 & 15{,}625 & 523.5{\scriptsize$\,\pm\,$5.9} & 535.1, 516.0, 519.3 \\
 &  & 64.0 & 31{,}250 & 558.8{\scriptsize$\,\pm\,$1.9} & 556.0, 558.0, 562.5 \\
 &  & \textbf{128.0} & 62{,}500 & \textbf{572.5}{\scriptsize$\,\pm\,$0.3} & 572.5, 573.0, 572.1 \\
\cmidrule{2-6}
 & \method{} (QGF+BFN4, $H_{\mathrm{exec}}{=}1$) & 16.0 & 7{,}813 & 531.6{\scriptsize$\,\pm\,$7.5} & 546.6, 523.2, 525.0 \\
 &  & 32.0 & 15{,}625 & 591.0{\scriptsize$\,\pm\,$3.3} & 596.8, 585.3, 590.9 \\
 &  & 64.0 & 31{,}250 & 643.7{\scriptsize$\,\pm\,$1.7} & 647.1, 642.3, 641.7 \\
 &  & \textbf{128.0} & 62{,}500 & \textbf{655.2}{\scriptsize$\,\pm\,$4.2} & 662.8, 648.5, 654.4 \\
\cmidrule{2-6}
 & TD-JEPA (reproduced, actor) & \textbf{128.0} & 125{,}000 & 480.1{\scriptsize$\,\pm\,$15.8} & 498.9, 448.7, 492.7 \\
 &  & 384.0 & 375{,}000 & 557.8{\scriptsize$\,\pm\,$15.7} & 527.9, 580.9, 564.6 \\
 &  & 768.0 & 750{,}000 & 567.8{\scriptsize$\,\pm\,$2.1} & 565.4, 572.0, 566.0 \\
 &  & 3072.0 & 3{,}000{,}000 & 638.9{\scriptsize$\,\pm\,$9.3} & 648.9, 620.4, 647.4 \\
\cmidrule{2-6}
 & One-step FB (actor) & 102.4 & 100{,}000 & 236.3{\scriptsize$\,\pm\,$13.9} & 214.1, 232.9, 261.9 \\
 &  & 204.8 & 200{,}000 & 302.6{\scriptsize$\,\pm\,$1.8} & 305.2, 303.4, 299.2 \\
 &  & 409.6 & 400{,}000 & 329.2{\scriptsize$\,\pm\,$6.8} & 340.3, 330.6, 316.7 \\
 &  & 819.2 & 800{,}000 & 257.7{\scriptsize$\,\pm\,$14.7} & 239.4, 246.7, 286.8 \\
 &  & 1024.0 & 1{,}000{,}000 & 216.1{\scriptsize$\,\pm\,$17.9} & 181.1, 227.3, 239.8 \\
\midrule
OGBench & \method{} (BFN16, $H_{\mathrm{exec}}{=}1$) & 32.0 & 7{,}813 & 24.7{\scriptsize$\,\pm\,$0.8} & 23.2, 25.2, 25.8 \\
(success, \%) &  & 64.0 & 15{,}625 & 28.7{\scriptsize$\,\pm\,$0.2} & 28.2, 28.9, 29.0 \\
 &  & 128.0 & 31{,}250 & 33.5{\scriptsize$\,\pm\,$1.3} & 30.9, 34.3, 35.3 \\
 &  & 192.0 & 46{,}875 & 38.0{\scriptsize$\,\pm\,$0.7} & 37.3, 37.3, 39.3 \\
 &  & \textbf{256.0} & 62{,}500 & \textbf{38.4}{\scriptsize$\,\pm\,$0.5} & 38.8, 37.4, 38.9 \\
\cmidrule{2-6}
 & TD-JEPA (reproduced, actor) & 64.0 & 250{,}000 & 22.2{\scriptsize$\,\pm\,$0.8} & 21.0, 21.8, 23.8 \\
 &  & 128.0 & 500{,}000 & 28.5{\scriptsize$\,\pm\,$0.8} & 28.3, 27.2, 29.9 \\
 &  & \textbf{256.0} & 1{,}000{,}000 & 32.0{\scriptsize$\,\pm\,$1.9} & 32.7, 28.3, 34.9 \\
\cmidrule{2-6}
 & One-step FB (actor) & 128.0 & 125{,}000 & 7.8{\scriptsize$\,\pm\,$1.0} & 7.0, 6.6, 9.9 \\
 &  & \textbf{256.0} & 250{,}000 & 10.0{\scriptsize$\,\pm\,$1.1} & 7.9, 10.3, 11.7 \\
\bottomrule
\end{tabular}
\end{table}

\FloatBarrier
\subsection{Full per-dataset benchmark and ablation results}
\label{app:zs-all}

\Cref{tab:zeroshot-full} reports per-dataset standard errors for the rolling-execution ($H_{\mathrm{exec}}{=}1$) methods of \cref{tab:zeroshot}, and \cref{tab:zeroshot-full-h8} reports the corresponding open-loop whole-chunk ($H_{\mathrm{exec}}{=}8$) results for Flow BC and \method{} (using 2-layer Flow BC on DMC and 4-layer Flow BC on OGBench).

\begin{table}[!htbp]
\centering
\caption{\textbf{Under rolling execution ($H_{\mathrm{exec}}{=}1$), \method{} with BFN16 alone outperforms the one-step FB actor on 12 of 13 state datasets and ties on \texttt{cube-double}.} Per-dataset mean\,$\pm$\,s.e.\ across 3 training seeds on the test split of \cref{tab:zeroshot} (2-layer Flow BC on DMC and 4-layer Flow BC on OGBench, with suite-average s.e.\ across the 3 per-seed suite means).}
\label{tab:zeroshot-full}
\scriptsize
\setlength{\tabcolsep}{4.5pt}
\setlength{\aboverulesep}{0pt}\setlength{\belowrulesep}{0pt}\renewcommand{\arraystretch}{1.15}%
\begin{tabular}{@{}l c >{\columncolor{baseblock}}c *{2}{>{\columncolor{abiblock}}c}>{\columncolor{abiblock}[\tabcolsep][0pt]}c@{}}
\toprule
 & One-step FB & \multicolumn{1}{>{\columncolor{baseblock}}c}{\emph{Unguided}} & \multicolumn{3}{>{\columncolor{abiblock}[\tabcolsep][0pt]}c@{}}{\textbf{$+$ \method{} critic (ours)}} \\
\cmidrule(lr){2-2}\cmidrule{3-6}
Dataset & Actor & Flow BC & BFN16 & QGF & QGF+BFN4 \\
\midrule
walker & 128.9{\scriptsize$\,\pm\,$78.8} & 158.2{\scriptsize$\,\pm\,$0.3} & 628.8{\scriptsize$\,\pm\,$1.4} & 717.6{\scriptsize$\,\pm\,$26.4} & 781.3{\scriptsize$\,\pm\,$17.6} \\
cheetah & 364.2{\scriptsize$\,\pm\,$49.2} & 103.5{\scriptsize$\,\pm\,$1.7} & 566.0{\scriptsize$\,\pm\,$3.8} & 671.5{\scriptsize$\,\pm\,$11.9} & 670.7{\scriptsize$\,\pm\,$15.6} \\
quadruped & 313.6{\scriptsize$\,\pm\,$57.9} & 50.6{\scriptsize$\,\pm\,$2.9} & 690.5{\scriptsize$\,\pm\,$7.3} & 728.6{\scriptsize$\,\pm\,$4.2} & 742.4{\scriptsize$\,\pm\,$9.6} \\
pointmass & 57.7{\scriptsize$\,\pm\,$10.8} & $-$78.6{\scriptsize$\,\pm\,$10.4} & 404.8{\scriptsize$\,\pm\,$5.5} & 400.5{\scriptsize$\,\pm\,$6.5} & 426.5{\scriptsize$\,\pm\,$1.3} \\
\midrule[0.4pt]
\emph{DMC (avg)} & 216.1{\scriptsize$\,\pm\,$17.9} & 58.4{\scriptsize$\,\pm\,$3.5} & 572.5{\scriptsize$\,\pm\,$0.3} & 629.5{\scriptsize$\,\pm\,$2.8} & 655.2{\scriptsize$\,\pm\,$4.2} \\
\midrule
antmaze-mn & 47.0{\scriptsize$\,\pm\,$2.0} & 5.0{\scriptsize$\,\pm\,$1.5} & 72.7{\scriptsize$\,\pm\,$3.0} & 65.7{\scriptsize$\,\pm\,$4.1} & 68.7{\scriptsize$\,\pm\,$2.4} \\
antmaze-ln & 19.0{\scriptsize$\,\pm\,$2.1} & 2.3{\scriptsize$\,\pm\,$0.9} & 40.3{\scriptsize$\,\pm\,$3.2} & 0.7{\scriptsize$\,\pm\,$0.7} & 33.0{\scriptsize$\,\pm\,$3.0} \\
antmaze-ms & 31.0{\scriptsize$\,\pm\,$4.0} & 0.3{\scriptsize$\,\pm\,$0.3} & 72.7{\scriptsize$\,\pm\,$3.0} & 45.3{\scriptsize$\,\pm\,$3.2} & 67.3{\scriptsize$\,\pm\,$2.9} \\
antmaze-ls & 4.7{\scriptsize$\,\pm\,$2.0} & 0.0{\scriptsize$\,\pm\,$0.0} & 9.0{\scriptsize$\,\pm\,$2.6} & 0.0{\scriptsize$\,\pm\,$0.0} & 5.7{\scriptsize$\,\pm\,$2.2} \\
antmaze-me & 5.0{\scriptsize$\,\pm\,$2.1} & 0.3{\scriptsize$\,\pm\,$0.3} & 46.3{\scriptsize$\,\pm\,$1.2} & 33.0{\scriptsize$\,\pm\,$1.5} & 38.3{\scriptsize$\,\pm\,$3.4} \\
cube-single & 4.0{\scriptsize$\,\pm\,$2.0} & 11.0{\scriptsize$\,\pm\,$1.2} & 77.3{\scriptsize$\,\pm\,$2.3} & 49.0{\scriptsize$\,\pm\,$2.5} & 56.3{\scriptsize$\,\pm\,$3.5} \\
cube-double & 0.0{\scriptsize$\,\pm\,$0.0} & 0.0{\scriptsize$\,\pm\,$0.0} & 0.0{\scriptsize$\,\pm\,$0.0} & 0.0{\scriptsize$\,\pm\,$0.0} & 0.0{\scriptsize$\,\pm\,$0.0} \\
scene & 8.7{\scriptsize$\,\pm\,$2.0} & 7.0{\scriptsize$\,\pm\,$2.0} & 16.3{\scriptsize$\,\pm\,$2.0} & 6.3{\scriptsize$\,\pm\,$1.8} & 5.0{\scriptsize$\,\pm\,$1.5} \\
puzzle-3x3 & 8.0{\scriptsize$\,\pm\,$3.6} & 0.7{\scriptsize$\,\pm\,$0.3} & 13.0{\scriptsize$\,\pm\,$3.1} & 9.3{\scriptsize$\,\pm\,$2.9} & 10.3{\scriptsize$\,\pm\,$4.1} \\
\midrule[0.4pt]
\emph{OGBench (avg)} & 14.1{\scriptsize$\,\pm\,$0.8} & 3.0{\scriptsize$\,\pm\,$0.4} & 38.6{\scriptsize$\,\pm\,$0.6} & 23.3{\scriptsize$\,\pm\,$0.3} & 31.6{\scriptsize$\,\pm\,$0.9} \\
\bottomrule
\end{tabular}

\end{table}

\begin{table}[!htbp]
\centering
\caption{\textbf{Under open-loop whole-chunk execution ($H_{\mathrm{exec}}{=}8$), all three \method{} guidance modes substantially improve over unguided Flow BC on both suites.} Per-dataset mean\,$\pm$\,s.e.\ across 3 training seeds on the test split of \cref{tab:zeroshot}, propagating per-dataset s.e.\ to suite means as specified in \cref{app:zs-protocol}.}
\label{tab:zeroshot-full-h8}
\scriptsize
\setlength{\tabcolsep}{5pt}
\setlength{\aboverulesep}{0pt}\setlength{\belowrulesep}{0pt}\renewcommand{\arraystretch}{1.15}%
\begin{tabular}{@{}l >{\columncolor{baseblock}}c *{2}{>{\columncolor{abiblock}}c}>{\columncolor{abiblock}[\tabcolsep][0pt]}c@{}}
\toprule
 & \multicolumn{1}{>{\columncolor{baseblock}}c}{\emph{Unguided}} & \multicolumn{3}{>{\columncolor{abiblock}[\tabcolsep][0pt]}c@{}}{\textbf{$+$ \method{} critic (ours)}} \\
\cmidrule{2-5}
Dataset & Flow BC & BFN16 & QGF & QGF+BFN4 \\
\midrule
walker & 164.7{\scriptsize$\,\pm\,$0.5} & 506.5{\scriptsize$\,\pm\,$4.6} & 682.7{\scriptsize$\,\pm\,$26.1} & 728.6{\scriptsize$\,\pm\,$22.1} \\
cheetah & 107.7{\scriptsize$\,\pm\,$3.0} & 574.3{\scriptsize$\,\pm\,$4.5} & 662.4{\scriptsize$\,\pm\,$7.5} & 671.9{\scriptsize$\,\pm\,$6.6} \\
quadruped & 55.3{\scriptsize$\,\pm\,$4.0} & 672.8{\scriptsize$\,\pm\,$2.9} & 702.1{\scriptsize$\,\pm\,$10.5} & 730.1{\scriptsize$\,\pm\,$3.0} \\
pointmass & $-$74.4{\scriptsize$\,\pm\,$14.0} & 413.9{\scriptsize$\,\pm\,$16.6} & 415.0{\scriptsize$\,\pm\,$10.1} & 446.7{\scriptsize$\,\pm\,$8.5} \\
\midrule[0.4pt]
\emph{DMC (avg)} & 63.3{\scriptsize$\,\pm\,$3.7} & 541.9{\scriptsize$\,\pm\,$4.5} & 615.6{\scriptsize$\,\pm\,$7.7} & 644.3{\scriptsize$\,\pm\,$6.2} \\
\midrule
antmaze-mn & 0.3{\scriptsize$\,\pm\,$0.3} & 59.7{\scriptsize$\,\pm\,$1.9} & 46.7{\scriptsize$\,\pm\,$1.5} & 58.0{\scriptsize$\,\pm\,$1.5} \\
antmaze-ln & 0.0{\scriptsize$\,\pm\,$0.0} & 10.0{\scriptsize$\,\pm\,$0.0} & 0.0{\scriptsize$\,\pm\,$0.0} & 8.7{\scriptsize$\,\pm\,$0.9} \\
antmaze-ms & 0.3{\scriptsize$\,\pm\,$0.3} & 46.3{\scriptsize$\,\pm\,$2.4} & 13.7{\scriptsize$\,\pm\,$3.2} & 36.7{\scriptsize$\,\pm\,$6.1} \\
antmaze-ls & 0.0{\scriptsize$\,\pm\,$0.0} & 0.7{\scriptsize$\,\pm\,$0.3} & 0.0{\scriptsize$\,\pm\,$0.0} & 0.3{\scriptsize$\,\pm\,$0.3} \\
antmaze-me & 0.0{\scriptsize$\,\pm\,$0.0} & 37.7{\scriptsize$\,\pm\,$4.9} & 30.7{\scriptsize$\,\pm\,$3.0} & 32.3{\scriptsize$\,\pm\,$2.7} \\
cube-single & 8.0{\scriptsize$\,\pm\,$1.5} & 80.7{\scriptsize$\,\pm\,$0.9} & 87.3{\scriptsize$\,\pm\,$1.7} & 72.3{\scriptsize$\,\pm\,$2.9} \\
cube-double & 1.0{\scriptsize$\,\pm\,$0.6} & 0.0{\scriptsize$\,\pm\,$0.0} & 1.0{\scriptsize$\,\pm\,$0.6} & 0.3{\scriptsize$\,\pm\,$0.3} \\
scene & 6.7{\scriptsize$\,\pm\,$1.8} & 14.7{\scriptsize$\,\pm\,$0.7} & 6.7{\scriptsize$\,\pm\,$1.5} & 5.7{\scriptsize$\,\pm\,$1.2} \\
puzzle-3x3 & 2.3{\scriptsize$\,\pm\,$1.5} & 13.3{\scriptsize$\,\pm\,$2.7} & 13.7{\scriptsize$\,\pm\,$0.7} & 15.0{\scriptsize$\,\pm\,$3.2} \\
\midrule[0.4pt]
\emph{OGBench (avg)} & 2.1{\scriptsize$\,\pm\,$0.3} & 29.2{\scriptsize$\,\pm\,$0.7} & 22.2{\scriptsize$\,\pm\,$0.6} & 25.5{\scriptsize$\,\pm\,$0.9} \\
\bottomrule
\end{tabular}

\end{table}

\FloatBarrier
\Cref{tab:zeroshot-variants} reports the \method{} component ablation matrix across all 13 state datasets at $H_{\mathrm{exec}}{=}8$ using the shared 2-layer Flow BC proposer on both suites: \emph{Default} combines 16-step state history, multi-horizon $\mathcal H=\{4,8,16,32\}$ successor bootstrapping, and the density readout $\hJ_D$, \emph{$\psi$ readout} switches to linear $\hJ_\psi$, \emph{No history} uses single-state inputs, \emph{Within-chunk target} removes backbone bootstrapping while retaining density-head bootstrapping, \emph{Both} combines single-state inputs with the within-chunk target, and \emph{8-step chunks only} trains solely at $H{=}8$.

\begin{table}[!htbp]
\centering
\caption{\textbf{Across controlled 2-layer Flow BC ablations ($H_{\mathrm{exec}}{=}8$), the feature-moment readout $\hJ_\psi$ raises DMC return across all modes while the within-chunk target with $\hJ_D$ raises OGBench success across all modes.} Per-dataset mean\,$\pm$\,s.e.\ across 3 training seeds on the test split. \underline{Underlined} entries mark the suite configuration of \cref{sec:exp-zeroshot}, QGF+BFN4$^\circ$ uses four guided draws, and ``--'' marks combinations not run.}
\label{tab:zeroshot-variants}
\scriptsize
\setlength{\tabcolsep}{4pt}
\begin{tabular}{@{}l c c c c c c@{}}
\toprule
\multicolumn{7}{@{}l}{\emph{Selection (BFN16)}} \\
Dataset & Default & $\psi$ readout & No history & \shortstack{Within-chunk\\target} & Both & 8-step chunks only \\
\midrule
walker & 509.5{\scriptsize$\,\pm\,$5.6} & \underline{506.5{\scriptsize$\,\pm\,$4.6}} & 523.1{\scriptsize$\,\pm\,$3.3} & 515.4{\scriptsize$\,\pm\,$1.8} & 526.7{\scriptsize$\,\pm\,$3.3} & 516.7{\scriptsize$\,\pm\,$4.9} \\
cheetah & 529.6{\scriptsize$\,\pm\,$20.0} & \underline{574.3{\scriptsize$\,\pm\,$4.5}} & 457.8{\scriptsize$\,\pm\,$9.6} & 499.2{\scriptsize$\,\pm\,$20.5} & 504.5{\scriptsize$\,\pm\,$5.6} & 513.7{\scriptsize$\,\pm\,$11.8} \\
quadruped & 590.1{\scriptsize$\,\pm\,$9.4} & \underline{672.8{\scriptsize$\,\pm\,$2.9}} & 460.4{\scriptsize$\,\pm\,$23.7} & 644.3{\scriptsize$\,\pm\,$2.4} & 614.1{\scriptsize$\,\pm\,$5.6} & 480.8{\scriptsize$\,\pm\,$39.1} \\
pointmass & 434.2{\scriptsize$\,\pm\,$16.8} & \underline{413.9{\scriptsize$\,\pm\,$16.6}} & 469.9{\scriptsize$\,\pm\,$14.8} & 456.3{\scriptsize$\,\pm\,$28.7} & 473.1{\scriptsize$\,\pm\,$20.7} & 391.7{\scriptsize$\,\pm\,$9.9} \\
\midrule[0.4pt]
\emph{DMC (avg)} & 515.8{\scriptsize$\,\pm\,$7.1} & 541.9{\scriptsize$\,\pm\,$4.5} & 477.8{\scriptsize$\,\pm\,$7.4} & 528.8{\scriptsize$\,\pm\,$8.8} & 529.6{\scriptsize$\,\pm\,$5.6} & 475.7{\scriptsize$\,\pm\,$10.6} \\
\midrule
antmaze-mn & 41.0{\scriptsize$\,\pm\,$6.1} & 3.0{\scriptsize$\,\pm\,$2.1} & 54.3{\scriptsize$\,\pm\,$3.3} & \underline{54.0{\scriptsize$\,\pm\,$5.0}} & 61.7{\scriptsize$\,\pm\,$2.4} & 17.7{\scriptsize$\,\pm\,$2.0} \\
antmaze-ln & 0.3{\scriptsize$\,\pm\,$0.3} & 0.0{\scriptsize$\,\pm\,$0.0} & 4.0{\scriptsize$\,\pm\,$2.0} & \underline{4.7{\scriptsize$\,\pm\,$1.5}} & 5.7{\scriptsize$\,\pm\,$1.2} & -- \\
antmaze-ms & 13.7{\scriptsize$\,\pm\,$4.8} & 0.0{\scriptsize$\,\pm\,$0.0} & 37.7{\scriptsize$\,\pm\,$0.9} & \underline{43.3{\scriptsize$\,\pm\,$4.4}} & 55.3{\scriptsize$\,\pm\,$6.5} & 1.7{\scriptsize$\,\pm\,$0.3} \\
antmaze-ls & 1.0{\scriptsize$\,\pm\,$0.6} & 0.0{\scriptsize$\,\pm\,$0.0} & 3.3{\scriptsize$\,\pm\,$2.3} & \underline{1.3{\scriptsize$\,\pm\,$0.3}} & 3.3{\scriptsize$\,\pm\,$0.7} & -- \\
antmaze-me & 13.3{\scriptsize$\,\pm\,$3.0} & 0.0{\scriptsize$\,\pm\,$0.0} & 3.3{\scriptsize$\,\pm\,$0.3} & \underline{30.0{\scriptsize$\,\pm\,$1.5}} & 14.0{\scriptsize$\,\pm\,$1.5} & -- \\
cube-single & 61.3{\scriptsize$\,\pm\,$0.9} & 66.0{\scriptsize$\,\pm\,$3.5} & 65.3{\scriptsize$\,\pm\,$2.3} & \underline{71.0{\scriptsize$\,\pm\,$0.6}} & 63.7{\scriptsize$\,\pm\,$3.2} & 54.7{\scriptsize$\,\pm\,$0.9} \\
cube-double & 0.0{\scriptsize$\,\pm\,$0.0} & 0.0{\scriptsize$\,\pm\,$0.0} & 0.3{\scriptsize$\,\pm\,$0.3} & \underline{0.3{\scriptsize$\,\pm\,$0.3}} & 0.0{\scriptsize$\,\pm\,$0.0} & 0.0{\scriptsize$\,\pm\,$0.0} \\
scene & 2.3{\scriptsize$\,\pm\,$1.2} & 8.3{\scriptsize$\,\pm\,$2.2} & 6.0{\scriptsize$\,\pm\,$0.6} & \underline{2.7{\scriptsize$\,\pm\,$1.2}} & 5.3{\scriptsize$\,\pm\,$0.9} & 1.7{\scriptsize$\,\pm\,$0.3} \\
puzzle-3x3 & 11.7{\scriptsize$\,\pm\,$3.3} & 16.7{\scriptsize$\,\pm\,$1.3} & 17.3{\scriptsize$\,\pm\,$1.7} & \underline{16.0{\scriptsize$\,\pm\,$1.5}} & 15.3{\scriptsize$\,\pm\,$0.9} & 8.0{\scriptsize$\,\pm\,$2.5} \\
\midrule[0.4pt]
\emph{OGBench (avg)} & 16.1{\scriptsize$\,\pm\,$1.0} & 10.4{\scriptsize$\,\pm\,$0.5} & 21.3{\scriptsize$\,\pm\,$0.6} & 24.8{\scriptsize$\,\pm\,$0.8} & 24.9{\scriptsize$\,\pm\,$0.9} & -- \\
\midrule
\multicolumn{7}{@{}l}{\emph{Guidance (QGF)}} \\
\midrule
walker & 627.9{\scriptsize$\,\pm\,$7.8} & \underline{682.7{\scriptsize$\,\pm\,$26.1}} & 665.1{\scriptsize$\,\pm\,$4.9} & 622.8{\scriptsize$\,\pm\,$19.7} & 638.7{\scriptsize$\,\pm\,$11.5} & 623.0{\scriptsize$\,\pm\,$19.4} \\
cheetah & 578.4{\scriptsize$\,\pm\,$8.2} & \underline{662.4{\scriptsize$\,\pm\,$7.5}} & 534.2{\scriptsize$\,\pm\,$11.6} & 510.4{\scriptsize$\,\pm\,$38.1} & 511.8{\scriptsize$\,\pm\,$29.0} & 563.1{\scriptsize$\,\pm\,$29.8} \\
quadruped & 662.3{\scriptsize$\,\pm\,$4.1} & \underline{702.1{\scriptsize$\,\pm\,$10.5}} & 644.9{\scriptsize$\,\pm\,$6.0} & 682.2{\scriptsize$\,\pm\,$5.4} & 678.6{\scriptsize$\,\pm\,$8.3} & 667.4{\scriptsize$\,\pm\,$9.6} \\
pointmass & 415.4{\scriptsize$\,\pm\,$28.3} & \underline{415.0{\scriptsize$\,\pm\,$10.1}} & 446.5{\scriptsize$\,\pm\,$14.9} & 447.4{\scriptsize$\,\pm\,$16.4} & 467.2{\scriptsize$\,\pm\,$15.5} & 382.7{\scriptsize$\,\pm\,$5.3} \\
\midrule[0.4pt]
\emph{DMC (avg)} & 571.0{\scriptsize$\,\pm\,$7.7} & 615.6{\scriptsize$\,\pm\,$7.7} & 572.7{\scriptsize$\,\pm\,$5.1} & 565.7{\scriptsize$\,\pm\,$11.6} & 574.1{\scriptsize$\,\pm\,$9.0} & 559.1{\scriptsize$\,\pm\,$9.3} \\
\midrule
antmaze-mn & 11.3{\scriptsize$\,\pm\,$4.9} & 0.0{\scriptsize$\,\pm\,$0.0} & 25.0{\scriptsize$\,\pm\,$2.1} & \underline{32.0{\scriptsize$\,\pm\,$4.5}} & 45.0{\scriptsize$\,\pm\,$4.6} & 0.3{\scriptsize$\,\pm\,$0.3} \\
antmaze-ln & 0.0{\scriptsize$\,\pm\,$0.0} & 0.0{\scriptsize$\,\pm\,$0.0} & 0.0{\scriptsize$\,\pm\,$0.0} & \underline{0.0{\scriptsize$\,\pm\,$0.0}} & 0.0{\scriptsize$\,\pm\,$0.0} & -- \\
antmaze-ms & 0.0{\scriptsize$\,\pm\,$0.0} & 0.0{\scriptsize$\,\pm\,$0.0} & 4.7{\scriptsize$\,\pm\,$3.7} & \underline{11.7{\scriptsize$\,\pm\,$2.6}} & 19.0{\scriptsize$\,\pm\,$0.6} & 0.0{\scriptsize$\,\pm\,$0.0} \\
antmaze-ls & 0.0{\scriptsize$\,\pm\,$0.0} & 0.0{\scriptsize$\,\pm\,$0.0} & 0.0{\scriptsize$\,\pm\,$0.0} & \underline{0.0{\scriptsize$\,\pm\,$0.0}} & 0.0{\scriptsize$\,\pm\,$0.0} & -- \\
antmaze-me & 9.3{\scriptsize$\,\pm\,$1.7} & 0.0{\scriptsize$\,\pm\,$0.0} & 0.7{\scriptsize$\,\pm\,$0.3} & \underline{24.0{\scriptsize$\,\pm\,$3.0}} & 10.7{\scriptsize$\,\pm\,$2.2} & -- \\
cube-single & 48.3{\scriptsize$\,\pm\,$4.3} & 41.0{\scriptsize$\,\pm\,$2.3} & 50.7{\scriptsize$\,\pm\,$1.2} & \underline{54.3{\scriptsize$\,\pm\,$5.5}} & 51.0{\scriptsize$\,\pm\,$1.0} & 38.3{\scriptsize$\,\pm\,$1.9} \\
cube-double & 0.0{\scriptsize$\,\pm\,$0.0} & 0.0{\scriptsize$\,\pm\,$0.0} & 0.0{\scriptsize$\,\pm\,$0.0} & \underline{0.0{\scriptsize$\,\pm\,$0.0}} & 0.0{\scriptsize$\,\pm\,$0.0} & 0.0{\scriptsize$\,\pm\,$0.0} \\
scene & 0.0{\scriptsize$\,\pm\,$0.0} & 3.0{\scriptsize$\,\pm\,$1.5} & 0.0{\scriptsize$\,\pm\,$0.0} & \underline{1.3{\scriptsize$\,\pm\,$0.9}} & 2.7{\scriptsize$\,\pm\,$0.7} & 0.0{\scriptsize$\,\pm\,$0.0} \\
puzzle-3x3 & 3.0{\scriptsize$\,\pm\,$1.5} & 6.0{\scriptsize$\,\pm\,$2.1} & 5.0{\scriptsize$\,\pm\,$2.3} & \underline{11.7{\scriptsize$\,\pm\,$1.9}} & 10.3{\scriptsize$\,\pm\,$1.2} & 2.0{\scriptsize$\,\pm\,$1.5} \\
\midrule[0.4pt]
\emph{OGBench (avg)} & 8.0{\scriptsize$\,\pm\,$0.8} & 5.6{\scriptsize$\,\pm\,$0.4} & 9.6{\scriptsize$\,\pm\,$0.6} & 15.0{\scriptsize$\,\pm\,$0.9} & 15.4{\scriptsize$\,\pm\,$0.6} & -- \\
\midrule
\multicolumn{7}{@{}l}{\emph{Guided selection (QGF+BFN4$^\circ$)}} \\
\midrule
walker & 695.9{\scriptsize$\,\pm\,$4.6} & \underline{724.2{\scriptsize$\,\pm\,$26.6}} & 721.1{\scriptsize$\,\pm\,$8.3} & 685.5{\scriptsize$\,\pm\,$7.1} & 697.7{\scriptsize$\,\pm\,$18.3} & 675.8{\scriptsize$\,\pm\,$9.9} \\
cheetah & 569.6{\scriptsize$\,\pm\,$10.1} & \underline{671.8{\scriptsize$\,\pm\,$7.1}} & 531.0{\scriptsize$\,\pm\,$10.8} & 516.1{\scriptsize$\,\pm\,$37.3} & 532.9{\scriptsize$\,\pm\,$10.3} & 554.9{\scriptsize$\,\pm\,$27.4} \\
quadruped & 652.1{\scriptsize$\,\pm\,$7.3} & \underline{730.4{\scriptsize$\,\pm\,$3.2}} & 655.6{\scriptsize$\,\pm\,$8.3} & 684.9{\scriptsize$\,\pm\,$2.5} & 687.1{\scriptsize$\,\pm\,$3.1} & 658.7{\scriptsize$\,\pm\,$11.7} \\
pointmass & 434.2{\scriptsize$\,\pm\,$25.2} & \underline{432.6{\scriptsize$\,\pm\,$9.4}} & 489.6{\scriptsize$\,\pm\,$21.2} & 467.1{\scriptsize$\,\pm\,$17.5} & 496.8{\scriptsize$\,\pm\,$13.3} & 398.7{\scriptsize$\,\pm\,$16.9} \\
\midrule[0.4pt]
\emph{DMC (avg)} & 587.9{\scriptsize$\,\pm\,$7.1} & 639.8{\scriptsize$\,\pm\,$7.3} & 599.3{\scriptsize$\,\pm\,$6.6} & 588.4{\scriptsize$\,\pm\,$10.5} & 603.6{\scriptsize$\,\pm\,$6.3} & 572.0{\scriptsize$\,\pm\,$8.9} \\
\midrule
antmaze-mn & 30.7{\scriptsize$\,\pm\,$1.2} & 0.7{\scriptsize$\,\pm\,$0.7} & 49.0{\scriptsize$\,\pm\,$2.6} & \underline{48.3{\scriptsize$\,\pm\,$4.1}} & 55.7{\scriptsize$\,\pm\,$3.0} & 5.7{\scriptsize$\,\pm\,$0.3} \\
antmaze-ln & 0.0{\scriptsize$\,\pm\,$0.0} & 0.0{\scriptsize$\,\pm\,$0.0} & 2.3{\scriptsize$\,\pm\,$0.7} & \underline{4.3{\scriptsize$\,\pm\,$1.2}} & 2.0{\scriptsize$\,\pm\,$0.6} & -- \\
antmaze-ms & 5.7{\scriptsize$\,\pm\,$0.9} & 0.3{\scriptsize$\,\pm\,$0.3} & 31.0{\scriptsize$\,\pm\,$0.6} & \underline{33.3{\scriptsize$\,\pm\,$2.2}} & 41.3{\scriptsize$\,\pm\,$4.1} & 1.0{\scriptsize$\,\pm\,$1.0} \\
antmaze-ls & 0.0{\scriptsize$\,\pm\,$0.0} & 0.0{\scriptsize$\,\pm\,$0.0} & 0.0{\scriptsize$\,\pm\,$0.0} & \underline{0.3{\scriptsize$\,\pm\,$0.3}} & 4.3{\scriptsize$\,\pm\,$0.9} & -- \\
antmaze-me & 7.0{\scriptsize$\,\pm\,$2.6} & 0.0{\scriptsize$\,\pm\,$0.0} & 1.7{\scriptsize$\,\pm\,$1.2} & \underline{21.7{\scriptsize$\,\pm\,$2.3}} & 13.0{\scriptsize$\,\pm\,$2.9} & -- \\
cube-single & 45.7{\scriptsize$\,\pm\,$5.9} & 40.0{\scriptsize$\,\pm\,$2.5} & 50.7{\scriptsize$\,\pm\,$3.4} & \underline{44.3{\scriptsize$\,\pm\,$3.4}} & 39.7{\scriptsize$\,\pm\,$4.2} & 40.3{\scriptsize$\,\pm\,$4.6} \\
cube-double & 0.3{\scriptsize$\,\pm\,$0.3} & 0.7{\scriptsize$\,\pm\,$0.3} & 0.3{\scriptsize$\,\pm\,$0.3} & \underline{0.0{\scriptsize$\,\pm\,$0.0}} & 0.0{\scriptsize$\,\pm\,$0.0} & 0.3{\scriptsize$\,\pm\,$0.3} \\
scene & 0.7{\scriptsize$\,\pm\,$0.3} & 7.3{\scriptsize$\,\pm\,$1.2} & 3.7{\scriptsize$\,\pm\,$0.9} & \underline{1.0{\scriptsize$\,\pm\,$1.0}} & 1.0{\scriptsize$\,\pm\,$1.0} & 2.0{\scriptsize$\,\pm\,$0.6} \\
puzzle-3x3 & 4.7{\scriptsize$\,\pm\,$2.6} & 11.0{\scriptsize$\,\pm\,$3.5} & 11.3{\scriptsize$\,\pm\,$1.5} & \underline{12.7{\scriptsize$\,\pm\,$3.8}} & 13.7{\scriptsize$\,\pm\,$2.0} & 5.7{\scriptsize$\,\pm\,$1.2} \\
\midrule[0.4pt]
\emph{OGBench (avg)} & 10.5{\scriptsize$\,\pm\,$0.8} & 6.7{\scriptsize$\,\pm\,$0.5} & 16.7{\scriptsize$\,\pm\,$0.5} & 18.4{\scriptsize$\,\pm\,$0.8} & 19.0{\scriptsize$\,\pm\,$0.8} & -- \\
\bottomrule
\end{tabular}

\end{table}

\FloatBarrier
\subsection{Density readouts on frozen baseline representations}
\label{app:zs-baseline-density}

To isolate multi-chunk representation learning from the task readout on OGBench, we train the categorical Bellman density readout of \cite{anonymous2026density} on the frozen 1M-update representations of reproduced TD-JEPA from \cref{app:zs-swap} and one-step FB from \cref{app:zs-baselines} across all nine state OGBench datasets and all three training seeds. Every density head uses the identical architecture, optimizer, schedule, batch size, $62.5\text{k}$-update budget, 2048-state negative support, episode split, and reference-transition set of \cref{app:zs-abi} with $\gamma{=}0.99$ and one-step recorded continuations matching the single-action interface of both baselines. For reproduced TD-JEPA, the two forward heads read the concatenation of $\phi(s)$ and both task-free state--action predictor embeddings while the backward head reads $\psi(s')$. For one-step FB, the two forward heads read the concatenation of its two forward representations $F(s,a)$ while the backward head reads $B(s',a')$ at the recorded departing action $a'$. All backbone inputs are standardized with training statistics while the backbones remain frozen. At test time, every critic ranks 16 candidate chunks from the shared noise stream via BFN16 with no tuning parameter on the test episodes of \cref{tab:zeroshot}, where the baseline critics score each chunk by its first action and \method{} scores the 8-step chunk.

\Cref{fig:supp-baseline-density,tab:zs-baseline-density} compare native linear task inference against the shared density readout across execution modes and proposer capacities. In \cref{fig:supp-baseline-density}(a), under rolling execution ($H_{\mathrm{exec}}{=}1$) with 4-layer Flow BC, both the default cross-chunk \method{} representation and the within-chunk target representation outperform frozen reproduced TD-JEPA and one-step FB representations under the identical density readout, with the largest per-dataset margins on \texttt{antmaze-me}, \texttt{antmaze-ln}, \texttt{antmaze-mn}, and \texttt{cube-single}. In \cref{fig:supp-baseline-density}(b), replacing native linear task inference (open diamonds) with the density readout (bars) raises reproduced TD-JEPA on every training seed and raises one-step FB on average at $H_{\mathrm{exec}}{=}1$, yet both \method{} representations still lead on every seed in the 9-dataset suite mean. Under whole-chunk execution ($H_{\mathrm{exec}}{=}8$), single-action baseline representations paired with the density readout do not improve over native linear task inference, whereas both chunk-conditioned \method{} representations lead both baselines on every training seed with either 4-layer or 2-layer Flow BC.

\begin{table}[!htbp]
\centering
\caption{\textbf{Under the identical density readout, \method{} representations outperform frozen reproduced TD-JEPA and one-step FB representations across all tested execution modes and proposer capacities.} OGBench success rate (\%) under BFN16 on the test split of \cref{tab:zeroshot} across 3 training seeds (dataset means and 9-dataset suite means\,$\pm$\,s.e.\ across per-seed suite means). \emph{Linear}: native baseline task inference from \cref{app:zs-protocol}. \emph{Density}: categorical density heads \cite{anonymous2026density} trained for $62.5\text{k}$ updates on each frozen representation. Bold marks each row's highest entry, and ``--'' marks combinations not rerun.}
\label{tab:zs-baseline-density}
\begingroup
\footnotesize
\setlength{\tabcolsep}{4.5pt}
\begin{tabular}{@{}l c cc cc cc@{}}
\toprule
 &  & \multicolumn{2}{c}{One-step FB} & \multicolumn{2}{c}{Reproduced TD-JEPA} & \multicolumn{2}{c}{\method{} (density)} \\
\cmidrule(lr){3-4}\cmidrule(lr){5-6}\cmidrule(l){7-8}
Dataset & Unguided & linear & density & linear & density & default & within-chunk \\
\midrule
\multicolumn{8}{@{}l}{\emph{4-layer Flow BC, rolling execution ($H_{\mathrm{exec}}{=}1$)}} \\
antmaze-mn & 5.0 & 21.3 & 34.3 & 27.3 & 52.7 & 69.7 & \textbf{72.7} \\
antmaze-ln & 2.3 & 4.3 & 3.3 & 4.7 & 13.0 & 31.3 & \textbf{40.3} \\
antmaze-ms & 0.3 & 30.7 & 43.7 & 26.7 & 55.0 & 54.0 & \textbf{72.7} \\
antmaze-ls & 0.0 & 0.0 & 1.3 & \textbf{15.0} & 13.3 & 11.0 & 9.0 \\
antmaze-me & 0.3 & 0.0 & 0.0 & 7.3 & 0.0 & 30.7 & \textbf{46.3} \\
cube-single & 11.0 & 7.3 & 19.0 & 16.7 & 61.0 & 69.7 & \textbf{77.3} \\
cube-double & 0.0 & 0.0 & \textbf{0.7} & 0.0 & 0.0 & 0.0 & 0.0 \\
scene & 7.0 & 21.3 & 17.3 & \textbf{32.7} & 22.0 & 16.7 & 16.3 \\
puzzle-3x3 & 0.7 & \textbf{30.0} & 17.7 & 20.3 & 12.3 & 14.7 & 13.0 \\
\emph{OGBench (avg)} & 3.0\,$\pm$\,0.4 & 12.8\,$\pm$\,0.9 & 15.3\,$\pm$\,0.9 & 16.7\,$\pm$\,0.3 & 25.5\,$\pm$\,1.7 & 33.1\,$\pm$\,1.5 & \textbf{38.6}\,$\pm$\,0.6 \\
{\scriptsize Per seed (3917, 3502, 8948)} & {\scriptsize 2.6, 3.7, 2.7} & {\scriptsize 13.2, 14.0, 11.1} & {\scriptsize 14.8, 14.0, 17.0} & {\scriptsize 17.1, 17.0, 16.1} & {\scriptsize 25.1, 22.7, 28.7} & {\scriptsize 35.6, 33.3, 30.3} & {\scriptsize 37.6, 38.7, 39.7} \\
\midrule
\multicolumn{8}{@{}l}{\emph{4-layer Flow BC, whole-chunk execution ($H_{\mathrm{exec}}{=}8$)}} \\
\emph{OGBench (avg)} & 2.1\,$\pm$\,0.5 & 7.6\,$\pm$\,0.4 & 7.2\,$\pm$\,0.1 & 11.7\,$\pm$\,0.1 & 8.3\,$\pm$\,0.5 & 22.9\,$\pm$\,1.1 & \textbf{29.2}\,$\pm$\,1.0 \\
{\scriptsize Per seed (3917, 3502, 8948)} & {\scriptsize 1.9, 3.0, 1.3} & {\scriptsize 7.3, 8.4, 7.0} & {\scriptsize 7.3, 7.3, 7.0} & {\scriptsize 11.9, 11.7, 11.4} & {\scriptsize 9.0, 8.6, 7.2} & {\scriptsize 24.6, 23.3, 20.7} & {\scriptsize 29.2, 31.0, 27.4} \\
\midrule
\multicolumn{8}{@{}l}{\emph{2-layer Flow BC, whole-chunk execution ($H_{\mathrm{exec}}{=}8$)}} \\
\emph{OGBench (avg)} & -- & -- & 5.4\,$\pm$\,0.6 & 7.3\,$\pm$\,0.6 & 4.6\,$\pm$\,0.2 & 16.1\,$\pm$\,0.9 & \textbf{24.8}\,$\pm$\,0.9 \\
{\scriptsize Per seed (3917, 3502, 8948)} & -- & -- & {\scriptsize 6.4, 4.4, 5.4} & {\scriptsize 8.2, 7.4, 6.2} & {\scriptsize 4.6, 4.3, 4.9} & {\scriptsize 16.4, 17.4, 14.3} & {\scriptsize 26.1, 23.1, 25.2} \\
\bottomrule
\end{tabular}
\endgroup
\end{table}

\begin{figure}[!htbp]
\centering
\includegraphics[width=\textwidth]{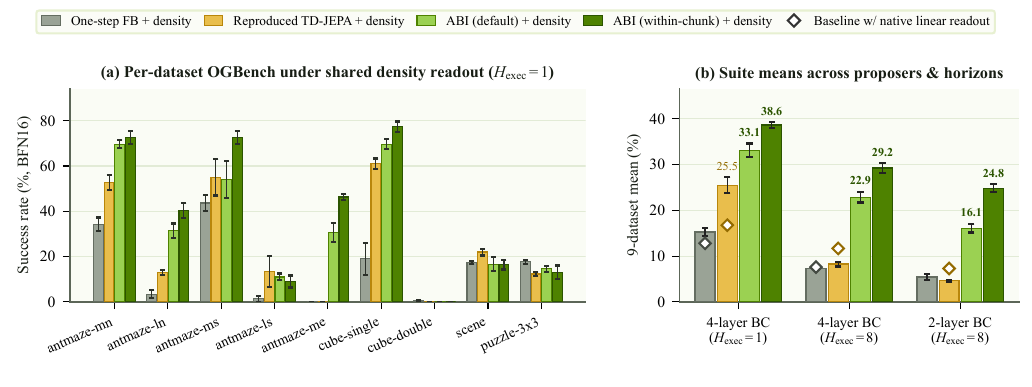}
\caption{\textbf{Under the identical categorical Bellman density readout, frozen \method{} representations outperform reproduced TD-JEPA and one-step FB representations on the nine-dataset mean across proposers and execution horizons.} OGBench test-split success rate (\%, BFN16, mean\,$\pm$\,1 s.e.\ across 3 training seeds) from \cref{tab:zs-baseline-density}. \emph{(a)} Per-dataset comparison at $H_{\mathrm{exec}}{=}1$ with 4-layer Flow BC. \emph{(b)} Nine-dataset suite means across proposer capacities and execution horizons, with open diamonds marking native linear task inference on the same frozen baseline backbones.}
\label{fig:supp-baseline-density}
\end{figure}

\FloatBarrier
\subsection{Multi-task closed-loop rollout progressions}
\label{app:zs-qualitative}

Complementing the three representative rollouts in \cref{fig:valuemap}, \cref{fig:supp-rollouts} visualizes step-by-step closed-loop trajectories from identical initial states across eight distinct locomotion, navigation, and manipulation tasks on seed-3917 development episodes under whole-chunk selection (BFN16, $H_{\mathrm{exec}}{=}8$, shared proposal noise streams). On dense-reward DMC (\cref{tab:zeroshot-full-h8}), guiding the 2-layer Flow BC proposer with $\hJ_\psi$ sustains coordinated high-speed running and dynamic jumping gaits on \texttt{quadruped}, whereas unguided Flow BC loses balance and flips over. On sparse-goal \texttt{antmaze-mn}, $\hJ_D$ guides Flow BC across four rooms from southwest to northeast (task 1) and around an interior wall from southeast to southwest (task 5), whereas unguided Flow BC stalls in local pockets until the 1{,}000-step horizon. Across OGBench manipulation with the 4-layer Flow BC proposer of \cref{app:zs-flowbcplus}, \method{} completes horizontal and diagonal pick-and-place on \texttt{cube-single}, sequentially opens both the drawer and the window on \texttt{scene}, and toggles the $3\times 3$ button grid to its target pattern in 47 steps on \texttt{puzzle-3x3}, whereas unguided Flow BC fails to reach the goal configuration on every task.

\begin{figure}[!htbp]
\centering
\includegraphics[width=0.94\textwidth]{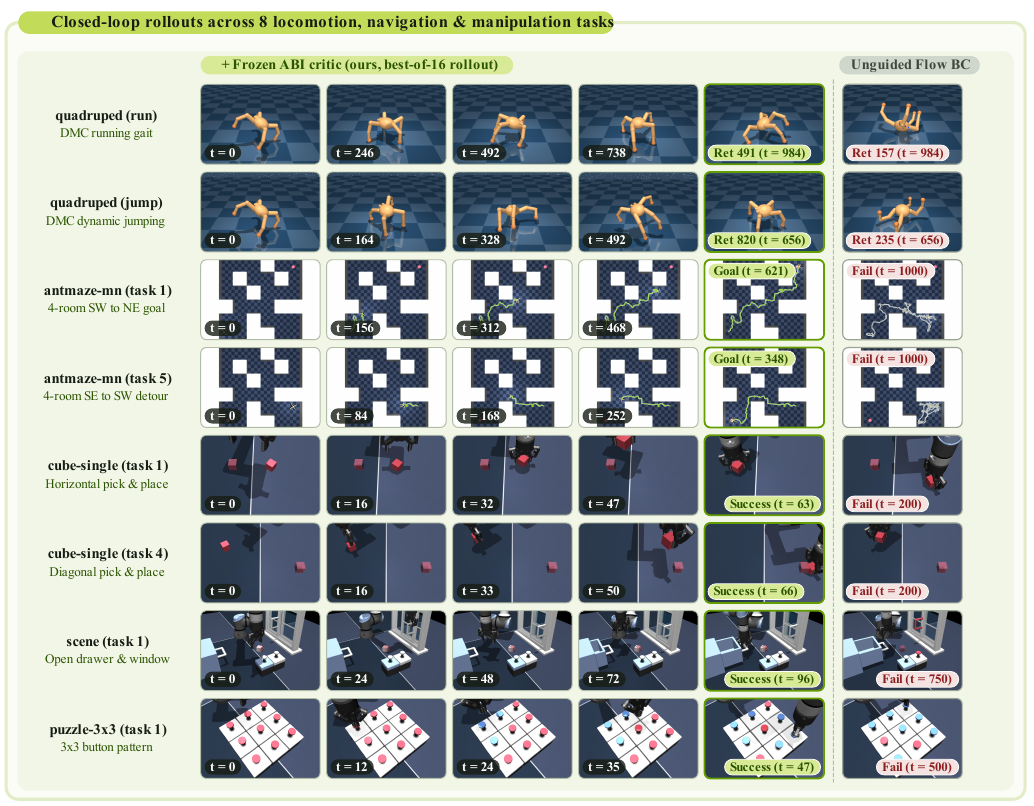}
\caption{\textbf{Across eight locomotion, navigation, and manipulation tasks, frozen \method{} critic selection guides Flow BC to completion from the same initial states where unguided Flow BC fails.} Development-episode closed-loop rollouts (seed 3917, BFN16, $H_{\mathrm{exec}}{=}8$, shared proposal noise) showing five frames from $t{=}0$ to task completion at the printed time steps under \method{} guidance (left five columns, green trails on \texttt{antmaze-mn}) alongside the terminal state of unguided Flow BC from the identical reset state (right column, gray trails on \texttt{antmaze-mn}), extending \cref{fig:valuemap}.}
\label{fig:supp-rollouts}
\end{figure}

\newpage

\subsection{OGBench data and evaluation tasks}
\label{app:zs-ogbench}

Offline OGBench trajectories are task-agnostic, and many are short or noisy, so a guided proposer must compose recorded behaviors to solve the evaluation tasks of \cref{app:zs-protocol}, which no dataset demonstrates. In \cref{fig:ogb-data}, the AntMaze datasets contain 1{,}000-step random-goal \texttt{navigate}, 200-step \texttt{stitch}, and noisy 500-step \texttt{explore} trajectories, drawn with the evaluation starts and goals, while \texttt{play} episodes record 1{,}000-step undirected object interaction. Reaching the 20 single-stage and multi-stage goal configurations from the start states in \cref{fig:ogb-tasks} likewise requires assembling manipulation play into directed behavior.

\begin{figure}[!t]
\centering
\includegraphics[width=0.96\linewidth]{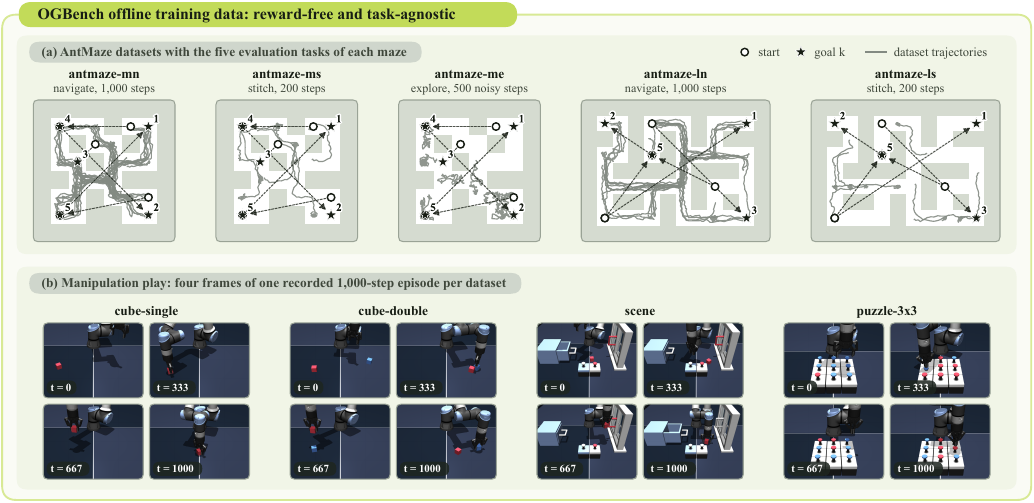}
\caption{\textbf{OGBench offline training datasets cover multi-room navigation fragments and task-agnostic manipulation play.} \emph{(a)}~The five AntMaze datasets with eight trajectories sampled from each, alongside the start (circle) and goal (star) of the five evaluation tasks of each maze. \emph{(b)}~Four frames of one recorded 1{,}000-step episode of each manipulation play dataset, rendered from stored simulator states.}
\label{fig:ogb-data}

\vspace{0.25em}
\includegraphics[width=0.96\linewidth]{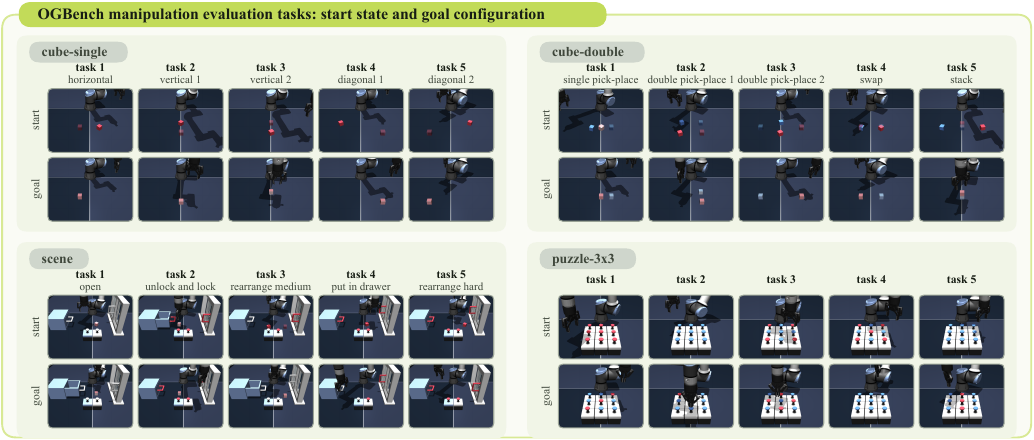}
\caption{\textbf{Each OGBench manipulation environment specifies five single-stage and multi-stage goal-reaching evaluation tasks.} Start state (top row) and goal configuration (bottom row) of each task across \texttt{cube-single}, \texttt{cube-double}, \texttt{scene}, and \texttt{puzzle-3x3}. In the cube environments, the goal positions also appear as translucent cubes in the start state.}
\label{fig:ogb-tasks}
\end{figure}

\section{Limitations and scope}
\label{app:limitations}

This section expands the limitations discussed in \cref{sec:conclusion}, complementing the metric comparison in \cref{app:not-claimed}.

\paragraph{Offline support and boundary exits.} Offline trajectories constrain the critic on covered state--chunk pairs, whereas \cref{prop:nonid} shows that off-support action preferences and gradients are not identified without local smoothness or broader action coverage. \method{} therefore anchors test-time selection and subspace gradient steps to the proposer's distribution, and \cref{lem:exit} bounds value error inside a residual-controlled region by its in-region residual and a discounted exit factor.

\paragraph{Single-step policy improvement versus $Q^\star$ iteration.} Because \method{} evaluates fixed continuation $\beta$ rather than iterating policy improvement to $Q^\star$, downstream gains follow single-step improvement over the proposer under the candidate-spread and continuation conditions in \cref{prop:headroom} and the rolling tail condition in \cref{prop:r7-rolling}.

\paragraph{Task composition.} Each evaluation episode uses one task readout, including multi-stage OGBench tasks specified by a single goal. Feature-moment and fixed-support density scores compose linearly in reward space as shown in \cref{app:composition}, whereas objectives that switch or logically couple readouts within an episode would require progress-augmented state representations and remain future work.

\fi

\end{document}